%% file: main.tex
\documentclass{elsarticle}
\usepackage{lineno,hyperref}
\modulolinenumbers[5]
\usepackage{xcolor}

\usepackage{tikz}
\usetikzlibrary{matrix,chains,positioning,decorations.pathreplacing,arrows,fit,calc}
\usepackage{amsfonts}
\usepackage{amssymb}
\usepackage{amsthm}
\usepackage{mathtools}
\usepackage[linesnumbered,vlined,ruled]{algorithm2e}

\usepackage{url}

\newcommand{\conditionR}{\emph{R}}
\newcommand{\conditionH}{\emph{H}}
\newcommand{\conditionRS}{\emph{R+S}}
\newcommand{\conditionHS}{\emph{H+S}}

\newcommand{\agentReg}{\emph{Regular agent}}
\newcommand{\agentPower}{\emph{Power pill agent}}
\newcommand{\agentGhost}{\emph{Fear-ghosts agent}}

\newcommand{\taskone}{retrospection task}
\newcommand{\tasktwo}{agent comparison task}

\newcommand{\fc}{\operatorname{\textit{fc}}}
\newcommand{\conv}{\operatorname{\textit{conv}}}

\newcommand{\RankBiserialCorrelation}{$r_{rb}$}

\usepackage{multirow}
\usepackage{array}
\newcolumntype{C}{>{$}c<{$}}

\begin{document}

\begin{frontmatter}

\title{Local and Global Explanations of Agent Behavior: Integrating Strategy Summaries with Saliency Maps}

\author[augsburg]{Tobias Huber}
\author[augsburg]{Katharina Weitz}
\author[augsburg]{Elisabeth Andr{\'e}}
\author[technion]{Ofra Amir}

\address[augsburg]{Universit{\"a}t Augsburg, Universit{\"a}tsstra{\ss}e 6a, Augsburg, Germany}
\address[technion]{Technion - Israel Institute of Technology}

\begin{abstract}
With advances in reinforcement learning (RL), agents are now being developed in high-stakes application domains such as healthcare and transportation. Explaining the behavior of these agents is challenging, as the environments in which they act have large state spaces, and their decision-making can be affected by delayed rewards, making it difficult to analyze their behavior. To address this problem, several approaches have been developed. Some approaches attempt to convey the \emph{global} behavior of the agent, describing the actions it takes in different states. Other approaches devised \emph{local} explanations which provide information regarding the agent's decision-making in a particular state. 
In this paper, we combine global and local explanation methods, and evaluate their joint and separate contributions, providing (to the best of our knowledge) the first user study of combined local and global explanations for RL agents.
Specifically, we augment strategy summaries that extract important trajectories of states from simulations of the agent with saliency maps which show what information the agent attends to.
Our results show that the choice of what states to include in the summary (global information) strongly affects people's understanding of agents: participants shown summaries that included important states significantly outperformed participants who were presented with agent behavior in a randomly set of chosen world-states.
We find mixed results with respect to augmenting demonstrations with saliency maps (local information), as the addition of saliency maps did not significantly improve performance in most cases.
However, we do find some evidence that saliency maps can help users better understand what information the agent relies on in its decision making, suggesting avenues for future work that can further improve explanations of RL agents.
\end{abstract}


\end{frontmatter}


\input{intro.tex}

\input{related.tex}
\input{argmax.tex}
\input{highlights.tex}
\input{implementation.tex}
\input{study_design.tex}
\input{results.tex}
\input{discussion.tex}
\input{conclusion.tex}

\bibliographystyle{elsarticle-harv}
\bibliography{refs_ofra,refs_tobi, refs_katha}

\newpage

\input{appendix.tex}

\end{document}

%% file: intro.tex
\section{Introduction}
\label{sec:introduction}

The maturing of artificial intelligence (AI) methods has led to the introduction of intelligent systems in  areas such as healthcare and transportation~\cite{stone2016artificial}. Since these systems are used by people in such high-stakes domains, it is crucial for users to be able to understand and anticipate their behavior. For instance, a driver of an autonomous vehicle will need to anticipate situations in which the car fails and hands over control to her, while a clinician will need to understand the treatment regime recommended by an agent to determine whether it aligns with the patient's preferences.

The recognition of the importance of human understanding of agents' behavior, together with the complexity of current AI systems, have led to a growing interest in developing ``explainable AI'' methods~\cite{doshi2017roadmap,gunning2017explainable,aha2017ijcai}. The idea of making AI systems explainable is itself not new, and was already discussed since the early days of expert systems~\cite{swartout1983xplain,chandrasekaran1989explaining}. However, state-of-the-art AI algorithms use more complex representations and algorithms (e.g., deep neural networks), making them harder to interpret. For example, in contrast to classical agent planning approaches such as the belief-desire-intention (BDI) framework~\cite{rao1995bdi} in which the goals of the agent are explicitly defined, current agents often use policies trained using complex reward functions and feature representations that are difficult for people to understand.

In this paper, we focus on the problem of describing and explaining the behavior of agents operating in sequential decision-making settings, which are  trained in a deep reinforcement learning framework. In particular, we explore the usefulness of \emph{global} and \emph{local} post-hoc explanations~\cite{molnar2019interpretable} of agent behavior. Global explanations describe the overall policy of the agent, that is, the actions it takes in different regions of the state space. An example of such global explanations are strategy summaries~\cite{amir2019summarizing}, which show demonstrations of the agent's behavior in a carefully selected set of world states. Local explanations, in contrast, aim to explain specific decisions made by the agent. For instance, saliency maps are used to show users what information the agent is attending to~\cite{greydanus2018}. 

We explore the combination of global and local information describing agent policies.
The motivation for integrating the two approaches is their complementary nature: while local explanations can help users understand what information the agent attends to in specific situations, they do not provide any information about its behavior in different contexts.
This is reinforced by a previous study conducted by Alqaraawi et al.~\cite{alqaraawi2020evaluating} who evaluated local explanations and came to the conclusion that sole instance-level explanations are not sufficient and should be augmented with global information.
Similarly, while demonstrating what actions the agent takes in a wide range of scenarios can provide users with a sense of the overall strategy of the agent, it does not provide any explanations as to what information the agent was considering when choosing how to act in a certain situation.

To examine the benefits of these two complementary approaches and their relative usefulness, we integrate strategy summaries with saliency maps. Specifically, we adapt the HIGHLIGHTS-DIV algorithm 
for generating strategy summaries from our previous work~\cite{amir18highlights} such that it can be applied to deep learning settings, and integrate it with saliency maps that are generated based on Layer-Wise Relevance Propagation (LRP) (using a method we previously published in ~\cite{huber2019enhancing}).  We combine these two approaches by adding to the summary generated by HIGHLIGHTS-DIV saliency maps showing what the agent attends to.

We evaluate this combination of global and local explanations in a user study in which we explore both the benefits of HIGHLIGHTS-DIV summaries and the benefits of adding saliency maps to strategy summaries. 
Specifically, we compare random summaries and HIGHLIGHTS-DIV summaries, both with and without the addition of saliency maps. 
Study participants complete two types of tasks requiring the analysis of different agents trained to play the game of Pacman: an \tasktwo{} in which they compare the performance of two agents, and a retrospection task, in which they  reflect on an agent's strategy.
We chose those tasks to investigate whether the users trusted the right agent and to evaluate their mental models of the agents, respectively.  

Our results show that participants who were shown HIGHLIGHTS-DIV summaries performed better on both tasks compared to participants who were shown random summaries, and were also more satisfied with HIGHLIGHTS-DIV summaries. We find mixed results with respect to the benefits of adding saliency maps to summaries, which improved participants' ability to identify some aspects of agents' strategies, but in most cases did not lead to improved performance. 

The paper makes the following contributions:
\begin{itemize}
    \item It demonstrates that the HIGHLIGHTS-DIV algorithm, which was so far only used on classic reinforcement learning, can be applied to deep reinforcement learning agents with slight adjustments. 
    \item It proposes a joint local and global explanation approach for RL agents by integrating LRP saliency maps and HIGHLIGHTS-DIV summaries.
    \item It evaluates the combination of global and local summaries in a user study, demonstrating the benefits of HIGHLIGHTS-DIV summaries and the potential benefits and limitations of local explanations based on saliency maps.
\end{itemize}

The remainder of this article is structured as follows: Section \ref{sec:related_work} reviews prior work on explainable intelligent agents, Sections \ref{sec:argmax} and \ref{sec:highlights} describe our previous works on local and global explanations, respectively. Section \ref{sec:implementation} details our combined implementation of those two methods, including the adaptation of HIGHLIGHTS-DIV to deep reinforcement learning.
We describe the empirical evaluation we conducted in Section \ref{sec:study_design}, and its results are summarized in Section \ref{sec:results}. 
Finally, we discuss the results of the study and future directions in Section \ref{sec:discussion}, and conclude in Section \ref{sec:conclusion}.

%% file: related.tex
\section{Related Work}
\label{sec:related_work}
In this section, we review related works on explainable AI. We begin with a short review of global and local explanation methods of machine learning models, elaborating on the use of saliency maps, which we also make use of. We then discuss in more depth prior works on global and local explanations of policies of agents operating in sequential decision-making settings such as RL agents.

\paragraph{Global and local methods for interpretable machine learning}
Broadly, our work relates to the problem of interpretable machine learning, that is, explanations for the decisions of prediction models~\cite{doshi2017roadmap}. 
Few interpretable machine learning approaches provide global explanations, e.g., by showing examples of a set of instances and specifying how they were classified~\cite{ribeiro2016should,kim2016examples} or by generating prototypical images that maximize the activation of specific neurons \cite{simonyan13dicn}. 

The majority of methods focus on local explanations that explain single decisions of the model.
To this end, various methods to measure the relevance of a part of the input for the model's decision have been proposed.
For visual input, this information is often displayed as saliency maps that highlight how relevant each pixel is for a particular decision of the agent. 
Since the input for the Atari agents we use in this study is visual, we will use the word saliency map method even if the very same algorithm can be used on non-visual input data.

Gradient-based saliency map generation methods \cite{simonyan13dicn,springenberg14guided-backprop,sundararajan2017axiomatic,selvaraju2016grad-cam} utilize the derivative with respect to the input to estimate how much a small change in this input's value would change the prediction. 

Occlusion-based methods (\cite{zeiler14deconv,ribeiro2016should,sixt2020restricting}), occlude areas inside the input and measure how much this changes the model's prediction.
The idea behind this is to introduce uncertainty to the occluded area and to see how much the model is influenced by the loss of information in that area.
Occlusion-based methods often come with the advantage of being independent of the model's structure but with the drawback of not being as precise as some model-specific methods. 

In contrast to the aforementioned methods for generating saliency maps, Bach et al. \cite{bach2015lrp} proposed Layer-wise Relevance Propagation (LRP), which uses the intermediate activations of the neurons during the forward pass to estimate the contribution of each input pixels to prediction.  

Common to all interpretable ML approaches is that they focus on one-shot decisions.
Thus, they do not fully address the problem of explaining behavior in sequential decision making settings, where the agent takes actions, earns rewards and affects the state of the world.

\paragraph{Local explanations of agent behavior} 
Several approaches have been introduced for explaining specific decisions in the context of Markov Decision Processes (MDP).
Some works attempt to provide justifications for a policy~\cite{khan2009minimal,khan2011automatically,dodson2011natural} by making statements about particular actions choices (e.g. an action was chosen because it will lead to a state that has higher value with higher probability). Others provide causal explanations by integrating a causal structure of the domain~\cite{seegebarth2012making,vanderwaa2018contrastive}.
Krarup et al.~\cite{krarup2019model} propose methods for generating contrastive explanations to explain action choices.

In this paper, we focus on the use of saliency maps for local explanations.
Several works have implemented saliency maps in the context of Deep Reinforcement Learning (DRL).
Because many DRL algorithms utilize CNNs it is possible to directly use the methods we covered in the previous paragraph on those algorithms.
Zahavy et al. \cite{zahavy2016graying} and Wang et al. \cite{wang2015dueling} for example used gradient-based saliency maps on traditional and Dueling Deep Q-Network (DQN) algorithms.

Greydanus et al.~\cite{greydanus2018} and Iyer et al.~\cite{iyer2018transparency} propose novel occlusion-based algorithms, where Greydanus et al. use Gaussian blur instead of complete occlusion and Iyer et al. utilize template matching to identify objects in the input.
This allows them to train a new agent on this additional information and then selectively occlude those objects. 

Lapuschkin et al.~\cite{lapuschkin2019} used LRP to visualize the classical DQN architecture. 
In this paper, we use a more selective LRP variant which we tested on RL agents in our previous work \cite{huber2019enhancing} (see section \ref{sec:argmax}).

\paragraph{Global explanations of agent behavior}
Several  global explanation methods describing what actions an agent takes in different states have been proposed.
Hayes et al.~\cite{hayes2017improving} developed a system that allows users to ``debug'' an agent's strategy by querying its decisions in situations specified by the user. In contrast to this approach, strategy summarization methods select a set of important states to share with the user, such that the user does not need to query the agent with respect to specific states. We note that the two approaches are complementary.
Booth et al.~\cite{booth2019evaluating} compile logical formulas that specify when certain behaviors occur, e.g., by stating for which regime in the state space an agent will perform a particular action. However, this approach requires a state representation that is understandable to the user, which may not be the case in many complex domains, especially when DRL is used. 

Our work takes the approach of summarizing agent policies (which we refer to as ``strategy summaries'') by demonstrating the behavior of an agent in a subset of world states which are considered important by the agent~\cite{amir2019summarizing,amir2018agent}.
Several methods have been proposed for selecting the subset of demonstrations to present in a summary. Some methods choose states that best enable the reconstruction of the original policy, using computational models such as inverse reinforcement learning (inferring the agent's reward function) or imitation learning (constructing a mapping from states to agents' actions)~\cite{huang2017enabling,lage2019exploring}.
An alternative approach uses heuristics for identifying ``interesting'' situations. The HIGHLIGHTS-DIV algorithm we utilize falls into this category, as it selects states based on the distribution of Q-values of different actions. We chose to use this approach since it does not make any assumptions about people's reasoning, is simpler computationally and was shown to improve users' understanding of agent behavior. Similar approaches have been developed in parallel~\cite{huang2018establishing,sequeira2019interestingness}, varying in the specific formulation of the interestingness criteria used to determine which states to include in the summary. 

Another recent line of work explored the problem of generating plans that are more understandable to people~\cite{kulkarni2019explicable,chakraborti2019plan,cashmore2019towards}.
The idea underlying this approach is that by having a model of human plans in a domain, it is possible to generate plans that achieve the desired goal while being as consistent as possible with people's mental models.
However, in contrast to the strategy summarization approach, these approaches have only considered goal-based plans for short-term tasks.
Furthermore, they require  a model of how people plan in the domain, which might not always be feasible to obtain.

\paragraph{Evaluation of explanation methods for RL agents}

Some recent user studies examined the use of saliency maps and strategy summaries to explain the behavior of RL agents to people.

Alqaraawi et al.~\cite{alqaraawi2020evaluating} and Selvaraju et al.~\cite{selvaraju2016grad-cam} found that participants, who saw saliency maps, were able to predict the decision of an image classification model better then participants who did not see them.
However, the participants were still only correct in about 60\% of the cases and Alqaraawi et al. proposed to look beyond instance-level explanations in the future.
For actual RL agents, Iyer et al.~\cite{iyer2018transparency} and Anderson et al.~\cite{anderson2019mere-mortals} also used an action prediction task to evaluate saliency maps but found no clear advantage of saliency maps. 
In addition to the prediciton task, Anderson et al. used a \taskone{} to get an even better understanding of participants' mental models and, in  addition to saliency maps, investigated reward decomposition \cite{erwig2018explaining} and a combination of both methods.
Here, they found significant positive effects for reward decomposition and the combined approach and a marginally significant (p = 0.086) effect in favor of saliency maps.

Strategy summaries have been evaluated using several different tasks. Huang et al.~\cite{huang2017enabling} and  Lage et al.~\cite{lage2019exploring} asked participants to predict what actions an agent would take based on summaries optimized for policy reconstructions. Their results show that summary methods that better match with people's computational models lead to improved action prediction, but that people may use different models in different contexts.
Summaries generated by a variety of interestingness criteria were shown to improve people's ability to identify regions of the state space in which an agent spends more time and regions of the state space in which an agent requires additional training~\cite{sequeira2019interestingness}. 
Importance-based summaries (e.g. HIGHLIGHTS-DIV) were shown to improve people's ability to identify the better performing agent in an \tasktwo{}~\cite{amir18highlights} and their ability to decide  whether to trust an agent in specific world states~\cite{huang2018establishing}. 

In sum, this work extends the existing state-of-the-art in explanations of RL agents, by proposing an integrated global and local explanation method, which enhances HIGHLIGHTS-DIV summaries (global) with LRP saliency maps (local), and conducting a user study to examine the joint and separate contributions of the local and global information to people's understanding of the behavior of RL agents.

%% file: argmax.tex
\section{Saliency Maps}
\label{sec:argmax}

In this section, we describe the local explanation method which we use in our combined local and global explanation approach.
While the development of the local explanation method is not the focus of this paper, we include the details of the approach for completeness.
We revisit the foundations of Layer-wise Relevance Propagation (LRP) and show how to use it on the original DQN.
Then we describe our previously published $argmax$-rule, an adjustment to this algorithm, which generates more selective saliency maps and which we use in this work.
In addition to some previously
published illustrations of the selectivity of the $argmax$-rule, we implemented new sanity checks for our saliency maps and report their results.

\subsection{Foundations}
\label{sec:argmax_foundations}

LRP does not describe a specific algorithm but a concept which can be applied to any classifier $f$ that fulfills the following two requirements.
First, $f$ has to be decomposable into several layers of computation where each layer can be modeled as a vector of real-valued functions.
Secondly, the first layer has to be the input $x$ of the classifier containing, for example, the input pixels of an image, and the last layer has to be the real-valued prediction of the classifier $f(x)$.
Any DRL agent fulfills those requirements if we only consider the output value that corresponds to the action we want to analyze.

For a given input $x$, the goal of any method following the LRP concept is to assign relevance values $R_{j}^{l}$ to each computational unit $j$ of each layer of computation $l$, in such a way that $R_{j}^{l}$ measures the
local contribution of the unit $j$ to the prediction $f(x)$.
A method of calculating those relevance values $R_{j}^{l}$ is said to follow the LRP concept if it sets the relevance value of the output unit to be the prediction $f(x)$ and calculates all other relevance values by defining 
\begin{align}\label{ErsteLRPGleichung}
R_{j}^{l} := \sum_{k \in \{j \text{ is input for neuron } k\}} R_{j \leftarrow k}^{l,l+1},
\end{align}
for \textbf{messages} $R_{j \leftarrow k}^{l,l+1}$, such that 
\begin{align}\label{ZweiteLRPGleichung}
R_{k}^{l+1} = \sum_{j \in \{j \text{ is input for neuron } k\}} R_{j \leftarrow k}^{l,l+1}.
\end{align}
In this way a LRP variant is determined by choosing messages $R_{j \leftarrow k}^{l,l+1}$.
Through definition \ref{ErsteLRPGleichung} it is then possible to calculate all relevance values $R_{j}^{l}$ in a backward pass, starting from the prediction $f(x)$ and going towards the input layer.
Furthermore, equation \ref{ZweiteLRPGleichung} gives rise to
\begin{align*}
\sum_{k} R_{k}^{l+1} & = \sum_{k} \sum_{j \in \{j \text{ is input for neuron } k\}} R_{j \leftarrow k}^{l,l+1} \\
& = \sum_{j} \sum_{k \in \{j \text{ is input for neuron } k\}} R_{j \leftarrow k}^{l,l+1} = \sum_{j} R_{j}^{l}.
\end{align*} 
This ensures that the relevance values of each layer $l$ are a linear decomposition of the prediction
\begin{align*}
f(x)= \dots = \sum_{j = 1}^{dim(l)} R_{j}^{l} = \dots = \sum_{j = 1}^{dim(input)} R_{j}^{input}.
\end{align*}
Such a linear decomposition is easier to interpret than the original classifier because we can think of positive values $R_{j}^{l}$ to contribute evidence in favor of the decision of the classifier and of negative relevance values to contribute evidence against the decision.

To use LRP on a DQN agent we first have to look at its network architecture. 
The DQN $f$, as introduced by Mnih et al. \cite{Mnih15}, consists of three convolutional layers $\conv_{1},...,\conv_{3}$ followed by two fully connected layers $\fc_{1}$ and $\fc_{2}$.
For an input $x$ we write $\fc_{i}(x)$ and $\conv_{i}(x)$ for the output of the layers $\fc_{i}$ and $\conv_{i}$, respectively, during the forward pass that calculates $f(x)$.
In this notation, the Q-Values (i. e. the output of the whole DQN) are $\fc_{2}(x)$.

Following the LRP notation, we denote the relevance value of the $j$-th neuron in the layer $l$ with $R_{j}^{l}$. 
As described above, we have to define messages $R_{j \leftarrow k}^{l,l+1}$ for any two consecutive Layers $l,l+1$ to determine a LRP variant.
For now we assume that $l+1$ is one of the fully connected layers $\fc_{i}$.
The convolutional case works analogously and will be covered in more detail in the next section.
$R_{j \leftarrow k}^{l,l+1}$ should measure the contribution of the $j$-th neuron of $\fc_{i-1}$ to the $k$-th neuron of $\fc_{i}$, therefore we have to look at the calculation of $\fc_{i}(x)_{k}$.
The fully connected layer $\fc_{i}$ uses a weight matrix $W_{i}$, a bias vector $b_{i}$ and an activation function $\sigma_{i}$ as parameters for its output.
Let $W_{i}^{k}$ be the $k$-th row of $W_{i}$ and $b_{i}^{k}$ the $k$-th entry of $b_{i}$. 
Then the activation of the $k$-th neuron in $\fc_{i}(x)$ is
\begin{align*}
	\sigma_{i}(W_{i}^{k} \cdot \fc_{i-1}(x)  + b_{i}^{k} ),
\end{align*} 
where $\cdot$ denotes the dot product and $\fc_{0}$ is the flattened output of $\conv_{3}$.

Usually the ReLU function $\sigma(x)=max(0,x)$ is used as activation function $\sigma_{i}$ in the DQN architecture.
Bach et al. \cite{bach2015lrp} argue that any monotonous increasing function $\sigma$ with $\sigma(0)=0$, like the ReLU function, conserves the relevance of the dot product $W_{i}^{k} \cdot \fc_{i-1}(x)$.
Newer LRP variants, like the one used by Montavon et al. \cite{montavon18}, also omit the bias when defining $R_{j \leftarrow k}^{l,l+1}$ .
With those two assumptions the relevance of each neuron of $\fc_{i-1}$ to $\fc_{i}(x)_{k}$ is the same as their contribution to the dot product $W_{i}^{k} \cdot \fc_{i-1}(x) = \sum_{j} w_{jk}\fc_{i-1}(x)_{j}$.
This is a linear decomposition, so we can use $ w_{jk}\fc_{i-1}(x)_{j} $ to measure the contribution of the $j$-th neuron of $\fc_{i-1}$.

Since we want to find the parts of the input that contributed evidence in favor of the decision of the DQN agent, we restrict ourself to the positive parts of that sum.
That is, we set 
\begin{align*}
z_{jk}^{+} \coloneqq 
\begin{cases}
	w_{jk}\fc_{i-1}(x)_{j}               & \text{if }  w_{jk}\fc_{i-1}(x)_{j} > 0\\
	0               & \text{if } w_{jk}\fc_{i-1}(x)_{j} \leq 0\\
\end{cases}.
\end{align*}
With this, we define the messages as $ R_{j \leftarrow k}^{l,l+1} \coloneqq \frac{z_{jk}^{+}}{\sum_{j} z_{jk}^{+} } R_{k}^{l+1} $. 
This method is called $z^{+}$-rule (without bias) and satisfies the LRP equation \ref{ZweiteLRPGleichung}.

\subsection{An argmax approach to LRP}
\label{chap:argmax}

\input{argmax_tikz.tex}

In this subsection, we introduce our adjustment to the LRP variant called $z^{+}$-rule which we revisited in the previous subsection.
Recent work \cite{iyer2018transparency,goel2018}  indicates that DRL agents focus on certain objects within the visual input. 
With our approach, we aim to generate saliency maps that reflect this property by focusing on the most relevant parts of the input instead of giving too many details.
For this purpose, we propose to use an $argmax$ function to find the most contributing neurons in each convolutional layer. 

This idea is inspired by Mopuri et al.~\cite{CNNFixations}, who generated visualizations for neural networks solely based on the positions of neurons that provide evidence in favor of the prediction.
During this process, they follow only the most contributing neurons in each convolutional layer.
Our method adds relevance values to the positions of those neurons and therefore expands the approach of Mopuri et al. by an additional dimension of information.
Since those relevance values follow the LRP concept, they also possess the advantageous properties of the LRP concept like conservation of the prediction value.

As we have seen in the foundations section \ref{sec:argmax_foundations}, a LRP method is defined by its messages $R_{j\leftarrow k}^{l,l+1}$ which propagate the relevance from a layer $l+1$ to the preceding layer $l$.
If $l+1$ is a fully connected layer $\fc_{i}$ of the DQN (see section \ref{sec:argmax_foundations} for our notation of the DQN architecture), we use the same messages that are used in the $z^{+}$-rule.
In the case that $l$ and $l+1$ are convolutional layers $\conv_{i-1}$ and $\conv_{i}$, we propose new messages based on the $argmax$ function.
To define those messages we analyze how the activation of a neuron $\conv_{i}(x)_{k}$ was calculated during the forward pass.
Let $W$ and $A$ denote the weight kernel and part of $\conv_{i-1}(x)$ respectively that were used to calculate $\conv_{i}(x)_{k}$ during the forward pass.
If we write $W$ and $A$ in appropriate vector form, we get
\begin{align*}
	\conv_{i}(x)_{k} = \sigma(\sum_{j} w_{j}a_{j} + b ),
\end{align*}
where $\sigma$ denotes the activation function of $\conv_{i}$ and $b$ the bias corresponding to $W$.
Analogously to the $z^{+}$-rule we assume that the activation function and the bias can be neglected when determining the relevance values of the inputs $a_{i}$.
We propose to use an $argmax$ function to find the most relevant input neurons by defining the messages in the following way
\begin{align*}
R_{j\leftarrow k}^{l,l+1} \coloneqq 
\begin{cases}
R_{k}^{l+1}               & \text{if }  j = argmax\{ w_{j}a_{j} \}\\
0               & \text{if not}.\\
\end{cases}
\end{align*}
This definition satisfies the LRP condition given by equation \ref{ZweiteLRPGleichung} because the only non vanishing summand of the sum
\begin{align*}
	\sum_{j \in \{j \text{ is input for neuron } k\}} R_{j \leftarrow k}^{l,l+1}
\end{align*}
is $ R_{k}^{l+1}$.

If we use the same $argmax$ approach to propagate relevance values from $\conv_{1}$ to the input $\conv_{0}$, then we get very sparse saliency maps where only a few neurons are highlighted.
If we highlight the entire areas of the input $\conv_{0}$ that were used to calculate relevant neurons of $\conv_{1}$, then we lose information about the relevance values inside those areas.
Therefore, we draw inspiration from the guided Grad-CAM approach introduced in \cite{selvaraju2016grad-cam}.
Guided Grad-CAM uses one thorough relevance analysis for the neurons of the last convolutional layer to get relevant areas for the specific prediction and another thorough relevance calculation for the input pixels to get fine granular relevance values inside those areas.
We already did a thorough analysis of the neurons of the last convolutional layer by using the $z^{+}$-rule on the fully connected layers.
By following the most relevant neurons through the convolutional layers we keep track of the input areas that contributed the most to those values.
Mimicking the second thorough analysis of the Guided Grad-CAM approach we propose to use the $z^{+}$-rule to propagate relevance values from $\conv_{1}$ to $\conv_{0}$.
This generates fine granular relevance values inside the areas identified by following the most contributing neurons and ascertains that those relevance values follow the LRP concept.

Figure \ref{fig:arg_z} visualizes the differences between our $argmax$ approach and the $z^{+}$-rule.
An implementation of our algorithm that builds up on the iNNvestigate framework \cite{alber2018innvestigate} can be found here: \url{https://github.com/HuTobias/LRP_argmax}.

\subsection{Illustration of the Selectivity of the argmax-rule}
\label{chap:results}

In order to verify that our $argmax$ approach, described in section \ref{sec:argmax}, creates more selective saliency maps than the $z^{+}$-rule (see section \ref{sec:argmax_foundations}), we tested our approach on three different Atari  2600 games. 
For all games, we trained an agent using the DQN implementation of the OpenAI baselines framework \cite{baselines2017}.
The results of all experiments are shown in our previous work \cite{huber2019enhancing}.
We review the Pacman results here, since we use this game in the user study evaluating our combined explanation approach.

\begin{figure}
	\centering	
	\includegraphics[width=0.3\linewidth]{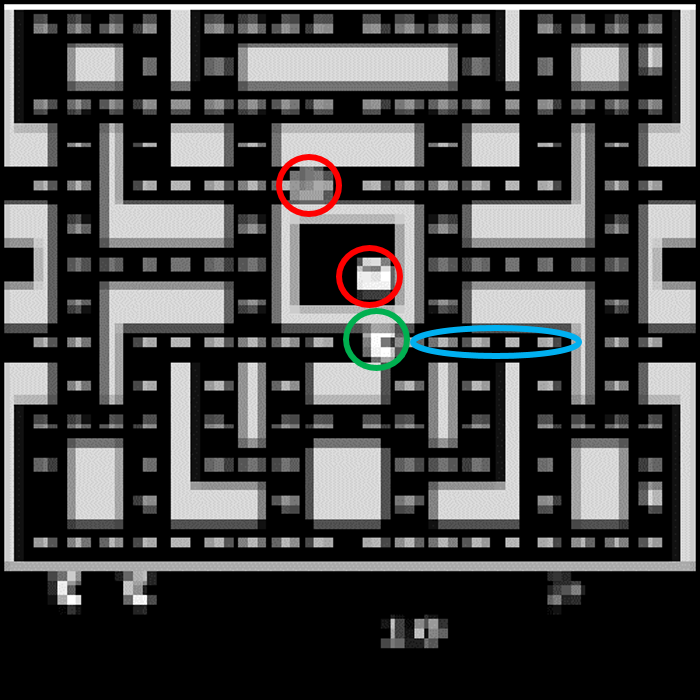}
	\includegraphics[width=0.3\linewidth]{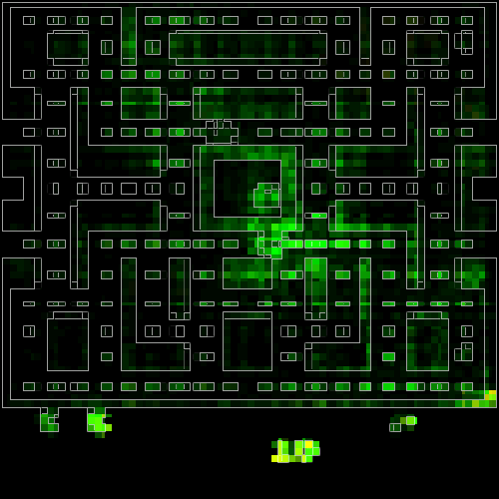}
	\includegraphics[width=0.3\linewidth]{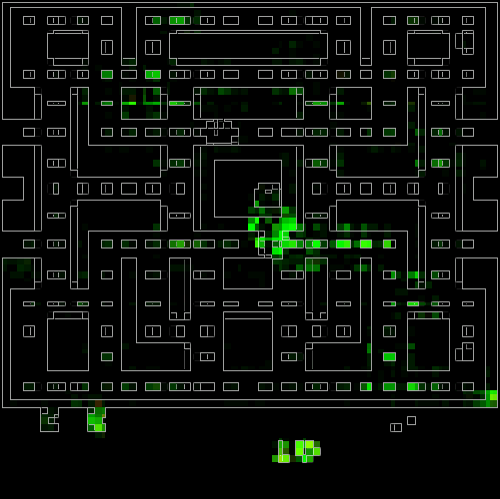}
	\caption{The left image shows a screen of Pacman. The player (green circle) has to collect pellets (blue area) while avoiding ghosts (red circles). The saliency map created for this game-state by the $z^{+}$-rule (middle)  highlights a huge area as relevant while our $argmax$ approach (right) focuses on the vicinity of the player.}
	\label{fig:focus_grob}
\end{figure}

In the game Pacman the player has to navigate through a maze and collect pellets while avoiding enemy ghosts.
Because this game contains many important objects and gives the agent a huge variety of possible strategies, DQN agents struggle in this environment and perform worse than the average human player (see \cite{Mnih15}).
Explainable AI methods are especially desirable in environments like this, where the agent is struggling, because they help us to understand where the agent had difficulties.  
The saliency maps created with the $z^{+}$-rule (figure \ref{fig:focus_grob}) reflect the complexity of Pacman by showing that the agent tries to look at nearly all of the objects in the game.
This information might be helpful to optimize the DRL agent, but it also distracts from the areas which influenced the agents' decision the most.
Figure \ref{fig:focus_grob} shows that the saliency map created by the $argmax$ approach is more focused on the vicinity of the agent and makes it clearer what the agent is focusing on the most.  
Figure \ref{fig:focus_grob} also illustrates that a fine-granular saliency map in the vicinity of the agent is necessary to see that the agent will most likely decide on moving to the right as his next action.

\subsection{Sanity Checks}
\label{sec:sanity_checks}

\begin{figure}[ht]
    \centering
    \begin{minipage}{0.15\linewidth}
    \centering
    \includegraphics[width=\linewidth]{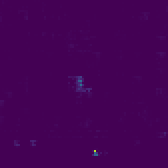}
    original
    \end{minipage}
     \begin{minipage}{0.15\linewidth}
    \centering
    \includegraphics[width=\linewidth]{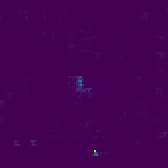}
    $\fc_2$
    \end{minipage}
     \begin{minipage}{0.15\linewidth}
    \centering
    \includegraphics[width=\linewidth]{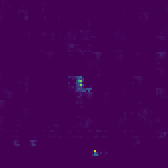}
    $\fc_1$
    \end{minipage}
     \begin{minipage}{0.15\linewidth}
    \centering
    \includegraphics[width=\linewidth]{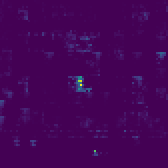}
    $\conv_3$
    \end{minipage}
     \begin{minipage}{0.15\linewidth}
    \centering
    \includegraphics[width=\linewidth]{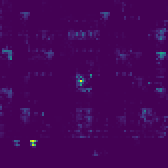}
    $\conv_2$
    \end{minipage}
    \begin{minipage}{0.15\linewidth}
    \centering
    \includegraphics[width=\linewidth]{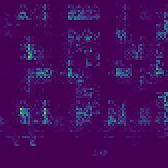}
    $\conv_1$
    \end{minipage}
    
    \caption{Example for how the LRP-argmax saliency maps change when the network's layers are randomized cascadingly, beginning with output layer $\fc_{2}$.} 
    \label{fig:sanity_vis}
\end{figure}

It is not yet possible to verify whether a saliency map algorithm perfectly reflects what a model learned.
However, a basic prerequisite for this is that the saliency maps depend on the weights learned by the model.
To verify this, Adebayo et al.~\cite{adebayo2018sanity} proposed sanity checks that cascadingly randomize each layer of the network, starting with the output layer.
If the saliency maps depend on the learned weights, then this will lead to increasingly different visualisations.
Sixt et al.~\cite{sixt2020} applied the sanity checks to several LRP variants but they have never been used on our $argmax$-rule.
Therefore, we implemented the sanity checks\footnote{The code we used for the sanity checks can be found here: \url{https://github.com/HuTobias/HIGHLIGHTS-LRP/tree/master/sanity_checks}} for our $argmax$-rule and test it on the regular Pacman agents described in section \ref{sec:study_design}.
An example of these tests for a single state is shown in Fig.~\ref{fig:sanity_vis}.

\begin{figure}[ht]
    \centering
    \includegraphics[width=0.3\linewidth]{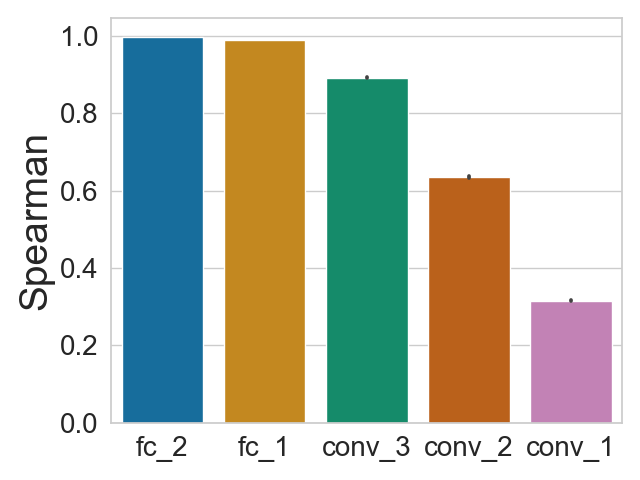}
    \includegraphics[width=0.3\linewidth]{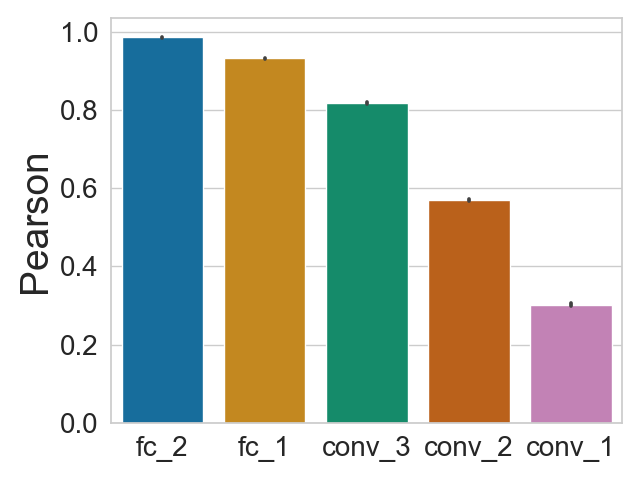}
    \includegraphics[width=0.3\linewidth]{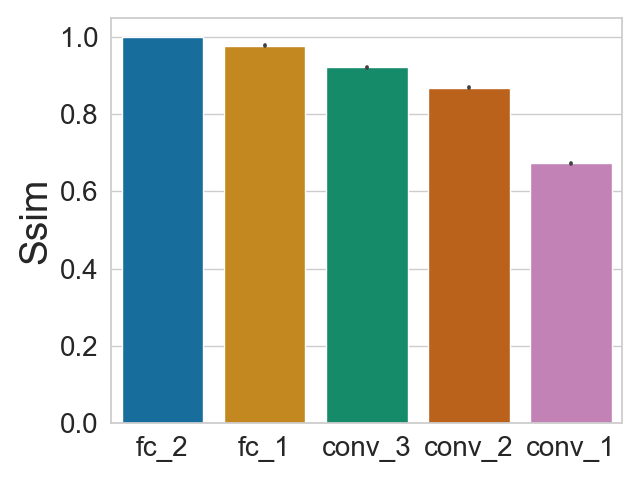}
    \caption{The average similarities between saliency maps for the fully trained agent and agents where the layers have been randomized cascadingly, starting with the last layer $fc_{2}$. The values are based on a stream of 1000 actions in the Atari 2600 Pacman game.
    }
    \label{fig:sanity_graphs}
\end{figure}

To measure how similar two saliency maps are we use three different metrics proposed by Adebayo et al.~\cite{adebayo2018sanity}: Spearman rank correlation, structural similarity (ssim) and Pearson correlation of the histogram of gradients. 
To account for a possible change of sign in the saliency maps, we adopt an approach by Sixt et al~\cite{sixt2020} and use the maximum similarity of the original and the inverted saliency map.
\newcommand{\myR}{\mathbb{R}^{m \times n \times c}}
That means that for two saliency maps $S,S^{'} \in \myR$ and a similarity measurement $sim: \myR \times \myR  \rightarrow \mathbb{R}$ we calculate the actual similarity with
\begin{equation}
    \max (sim(S,S^{'}),sim(\mathbf{1}-S,S^{'}))
\end{equation}
where $\mathbf{1} \in \myR$ is filled with $1$s.
Fig.~\ref{fig:sanity_graphs} shows the average similarities per randomized layer for a gameplay stream of 1000 states.

The relatively high values for the structural similarity (ssim) can be explained by the high amount of intersecting zeros in all saliency maps.
Apart from that, we see the same trends already observed by Sixt et al.~\cite{sixt2020} and Adebayo et al.\cite{adebayo2018sanity}: 
the sanity maps do analyze the learned weights but the fully connected layers are not sufficiently analyzed.
As a consequence, the saliency maps are not class discriminatory.
However, class discriminatory saliency maps often come with other drawbacks like being noise \cite{sixt2020} or not analyzing all layers \cite{selvaraju2016grad-cam}. 

%% file: argmax_tikz.tex
	\begin{figure}[t]
		
		\begin{center}
			\begin{tikzpicture}[
			plain/.style={
				draw=none,
				fill=none,
			},
			net/.style={
				matrix of nodes,
				nodes={
					draw,
					circle,
					inner sep=2pt
				},
				nodes in empty cells,
				column sep=0.7cm,
				row sep=3pt
			},
			>=latex
			]	
			
			\matrix[net,row sep= 0.01\linewidth,column sep=0.15\linewidth] at (0,0) (mat)
			{
				|| & || & ||	  \\
				|| & || & ||	& || &  |[plain]|   \\
				|| & || & ||	& || &  || \\
				|| & || & ||	& || &  |[plain]|    \\
				|| & |[inner sep = 0.5pt]| $a_{j}$ & |[inner sep = 0.5pt]| $a_{k}$	& |[plain]|   \\
			};
			\foreach \ai in {1,2,3,4,5}
			{\foreach \aii in {1,2,3,4,5}
				\draw[-latex]  (mat-\aii-1) -- (mat-\ai-2);
			}
			\foreach \ai in {1,2,3,4,5}
			{\foreach \aii in {1,2,3,4,5}
				\draw[-latex]  (mat-\aii-2) -- (mat-\ai-3);
			}
			\foreach \ai in {1,2,3,4,5}
			{\foreach \aii in {2,3,4}
				\draw[-latex]  (mat-\ai-3) -- (mat-\aii-4);
			}
			\foreach \ai in {2,3,4}
			{\foreach \aii in {3}
				\draw[-latex]  (mat-\ai-4) -- (mat-\aii-5);
			}
			
			\draw[-latex] (mat-5-2) -- (mat-5-3)
			node[midway,sloped,below] {$w_{jk}$};
			
			\node[inner sep =0pt] (input) at ($(mat-3-1) + (-2cm,-0.12cm)$)
			{\includegraphics[width=.07\textwidth]{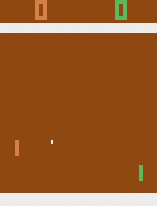}};
			
			\foreach \ai in {1,2,3,4,5}
			{
				\draw[-latex]	(input) -- (mat-\ai-1);
				\draw[-latex]	(input.north east) -- (mat-\ai-1);
				\draw[-latex]	(input.south east) -- (mat-\ai-1);
			}
			
			\end{tikzpicture}
			
			\small{Forward pass.}
		\end{center}
		
		\begin{minipage}{.45\linewidth}

			\begin{tikzpicture}[
			plain/.style={
				draw=none,
				fill=none,
			},
			net/.style={
				matrix of nodes,
				nodes={
					draw,
					circle,
					inner sep=2.5pt
				},
				nodes in empty cells,
				column sep=0.7cm,
				row sep=2pt
			},
			>=latex
			]	
			
			\matrix[net,column sep=0.1\linewidth] at (0,0) (mat)
			{
				|[ fill = red!25]| & |[ fill = red!20]|	& |[ fill = red!20]|   \\
				|[ fill = red!70]| & |[ fill = red!60]|	& |[ fill = red!45]| & |[ fill = red!75]|  \\
				|[ fill = red!45]| & |[ fill = red!20]|	& |[ fill = red!45]| & |[ fill = red!50]| & |[ fill = red!100]| \\
				|[ fill = red!50]| & |[ fill = red!40]|	& |[ fill = red!20]| & |[ fill = red!75]|  \\
				|[ fill = red!10]| & |[ fill = red!60,inner sep=0.1pt]|  $R_{j}$	& |[ fill = red!70,inner sep=0.1pt]|  $R_{k}$   \\
			};
			
			\node [below =0.3pt of mat-5-2] {
				\begin{small}	 	
				$
				R_{j} = \sum_{k} \frac{(a_{j} w_{jk})^{+}}{\sum_{j} (a_{j} w_{jk})^{+} } R_{k}
				$
				\end{small}
			};
			
			\foreach \ai in {2,3,4}
			{\foreach \aii in {3}
				\draw[-latex]  (mat-\aii-5) -- (mat-\ai-4);
			}
			\foreach \ai in {1,2,3,4,5}
			{\foreach \aii in {2,3,4}
				\draw[-latex]  (mat-\aii-4) -- (mat-\ai-3);
			}

			\foreach \ai in {1,2,3,4,5}
			{\foreach \aii in {1,2,3,4,5}
				\draw[-latex]  (mat-\aii-3) -- (mat-\ai-2);
			}
			\foreach \ai in {1,2,3,4,5}
			{\foreach \aii in {1,2,3,4,5}
				\draw[-latex]  (mat-\aii-2) -- (mat-\ai-1);
			}
			
			\node[inner sep =0pt] (heatmap) at ($(mat-3-1) + (-1.6cm,-0.12cm)$)
			{\includegraphics[width=0.2\linewidth]{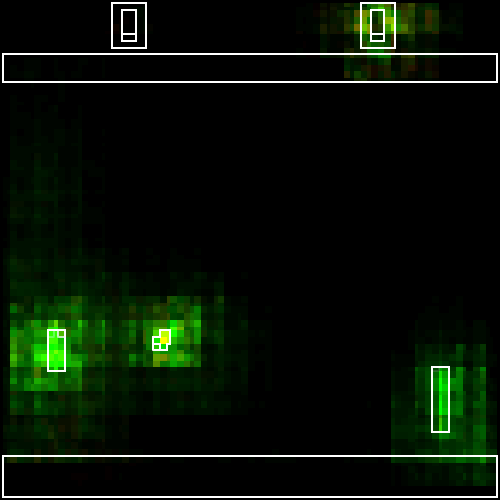}};
			
			\foreach \ai in {1,2,3,4,5}
			{
				\draw[-latex]	(mat-\ai-1) -- (heatmap.north east);
				\draw[-latex]	(mat-\ai-1) -- (heatmap);
				\draw[-latex]	(mat-\ai-1) -- (heatmap.south east);
			}
			
			\end{tikzpicture}		
			\small{Relevance propagation using the $z^{+}$-Rule.}		
		\end{minipage}
		\hspace{0.05\linewidth}		
		\begin{minipage}{.45\linewidth}	
			
			\begin{tikzpicture}[
			plain/.style={
				draw=none,
				fill=none,
			},
			net/.style={
				matrix of nodes,
				nodes={
					draw,
					circle,
					inner sep=2.5pt
				},
				nodes in empty cells,
				column sep=0.7cm,
				row sep=2pt
			},
			>=latex
			]	
			
			
			\matrix[net,column sep=0.1\linewidth,right = 0.5cm of heatmap]  (mat)
			{
				|[ fill = red!40]| & |[ fill = red!40]|	& |[ fill = red!20]|   \\
				|[ fill = red!20]| & |[ fill = red!0]|	& |[ fill = red!40]| & |[ fill = red!40]|  \\
				|| & |[fill = red!20]|	& |[ fill = red!40]| & |[ fill = red!50]| & |[ fill = red!100]| \\
				|[ fill = red!70]| & ||	&  |[ fill = red!40]|  & |[ fill = red!40]| \\
				|| & |[fill = red!70,inner sep=0.1pt]| $R_{j}$	& |[ fill = red!70,inner sep=0.1pt]|  $R_{k}$  \\
			};
			
			\node [below =0.5pt of mat-5-2] {
				\begin{small}	 	
				$
				R_{j} = \sum\limits_{j = argmax\{a_{j}w_{jk}\}}^{} R_{k}   
				$
				\end{small}
			};
			
			\foreach \ai in {2,3,4}
			{\foreach \aii in {3}
				\draw[-latex]  (mat-\aii-5) -- (mat-\ai-4);
			}
			\foreach \ai in {1,2,3,4,5}
			{\foreach \aii in {2,3,4}
				\draw[-latex]  (mat-\aii-4) -- (mat-\ai-3);
			}
			
			\draw[-latex]  (mat-2-3) -- (mat-5-2);
			\draw[-latex]  (mat-3-3) -- (mat-1-2);
			\draw[-latex]  (mat-1-3) -- (mat-3-2);
			\draw[-latex]  (mat-4-3) -- (mat-1-2);
			\draw[-latex]  (mat-5-3) -- (mat-5-2);
			
			\draw[-latex]  (mat-3-2) -- (mat-2-1);
			\draw[-latex]  (mat-1-2) -- (mat-1-1);
			\draw[-latex]  (mat-5-2) -- (mat-4-1);
			
			\node[inner sep =0pt] (heatmap) at ($(mat-3-1) + (-1.6cm,-0.12cm)$)
			{\includegraphics[width=0.2\linewidth]{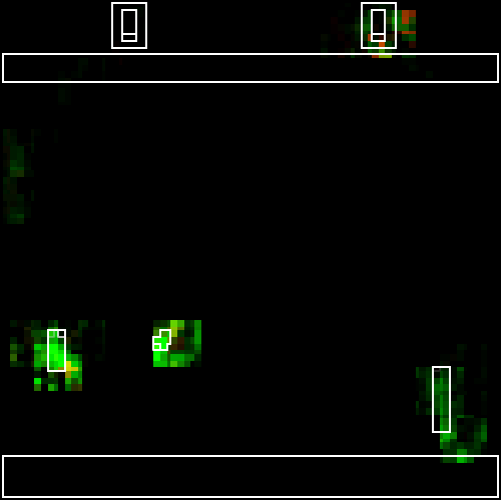}};
			
			\foreach \ai in {1,2,3,4,5}
			{
				\draw[-latex]	(mat-\ai-1) -- (heatmap.north east);
				\draw[-latex]	(mat-\ai-1) -- (heatmap);
				\draw[-latex]	(mat-\ai-1) -- (heatmap.south east);
			}

			\end{tikzpicture}
			
			\small{Relevance propagation using the $argmax$ approach.}
		\end{minipage}	
		\caption{A visualization of how our $argmax$ approach differs from the $z^{+}$ Rule. }
		\label{fig:arg_z}
	\end{figure}

%% file: highlights.tex
\section{Strategy Summarization}
\label{sec:highlights}

This section describes the strategy summarization approach to global explanations, and the HIGHLIGHTS algorithm and its extension HIGHLIGHTS-DIV, which we developed and evaluated in prior work~\cite{amir18highlights}.

Our formalization of the summarization problem assumes that the agent uses a Markov Decision Process (MDP), where $A$ is the set of actions available to the agent, $S$ is the set of states, $R$: $S \times A \rightarrow \mathbb{R}$ is the reward function, which maps each state and action to a reward, and $Tr$ is the transition probability function, i.e., $Tr(s', a, s)$ defines the probability of reaching state $s'$, when taking action $a$ in state $s$. The agent has a policy $\pi$ which specifies which action to take in each of the states. 

We formalize the problem of summarizing an agent's behavior as follows: from execution traces of an agent, choose a set $T = \langle t_{1},...,t_{k} \rangle$ of trajectories to include in the summary, where each trajectory is composed of a sequence of $l$ consecutive states and the actions taken in those states $\langle(s_{i},a_{i}),...,(s_{i+l-1},a_{i+l-1})\rangle$. We consider trajectories rather than single states because seeing what action was taken by the agent in a specific state might not be meaningful without a broader context (e.g., watching a self-driving car for one second will not reveal much useful information). Because it is infeasible that people will be able to review the behavior of an agent in all possible states, we assume a limited budget $k$ for the size of the summary, such that $|T| = k$. This budget limits the amount of time and cognitive effort that a person needs to invest in reviewing the agent's behavior.

There are several factors that could be considered when deciding which states to include in a summary, such as the effect of taking a different action in that state, the diversity of the states that are included in the summary and the frequency at which states are likely to be encountered by the agent. The approach we describe here focuses on the first factor, which we refer to as the ``importance'' of a state. Intuitively, a good summary should provide a person reviewing the summary with a sense of the agent's behavior in states that the person considers important (e.g., when making a mistake would be very costly). The importance of states included in the summary could substantially affect the ability of a person to assess an agent's capabilities. For example, imagine a summary of a self-driving car that only shows the car driving on a highway with no interruptions. This summary would provide people with very little understanding of how the car might act in other, more important, scenarios (e.g., when another car drives into its lane, when there is road construction). In contrast, a summary showing the self-driving car in a range on more interesting situations (e.g., overtaking another car, breaking when a person enters the road) would convey more useful information to people reviewing it. 

\subsection{The ``Highlights'' Algorithm}
\label{sec:alg}
The HIGHLIGHTS algorithm generates a summary of an agent's behavior from simulations of the agent in an online manner. It uses the notion of state \emph{importance}~\cite{torrey2013teaching} to decide which states to include in the summary.  Intuitively, a state is considered important if taking a wrong action in that state can lead to a significant decrease in future rewards, as determined by the agent's Q-values. Formally, the importance of a state, denoted $I(s)$, is defined as: 
\begin{equation}
\label{eq:importance}
I(s)=\max\limits_{a}Q^{\pi}_{(s,a)}-\min\limits_{a}Q^{\pi}_{(s,a)}
\vspace{-0.1cm}
\end{equation}
 This measure has been shown to be useful for choosing teaching opportunities in the context of student-teacher reinforcement learning~\cite{torrey2013teaching,amir2016interactive}. 

Before providing a detailed pseudo-code of the algorithm, we describe its operation at a high-level. HIGHLIGHTS generates a summary that includes trajectories that capture the most important states that an agent encountered in a given number of simulations. To do so, at each step it evaluates the importance of the state and adds it to the summary if its importance value is greater than the minimal value currently represented in the summary (replacing the minimal importance state). To provide more context to the user,  for each such state HIGHLIGHTS also extracts a trajectory of states neighboring it and the actions taken in those states.

A pseudo-code of the HIGHLIGHTS algorithm is  given in Algorithm~\ref{alg:highlights}. 
Table~\ref{tb:parameters} summarizes the parameters of the algorithm. 
HIGHLIGHTS takes as input the policy of the agent $\pi$ which is used to determine the agent's actions in the simulation and state importance values, the budget for the number of trajectories to include in the summary ($k$) and the length of each trajectory surrounding a state ($l$). Each such trajectory includes both states preceding  the important state and states that were encountered immediately after it. The number of subsequent states to include is determined by the $statesAfter$ parameter (the number of preceding states can be derived from this parameter and $l$). We also specify the number of simulations that can be run ($numSimulations$), and the minimal ``break'' interval between trajectories ($intervalSize$) which is used to prevent overlaps between trajectories. HIGHLIGHTS outputs a summary of the agent's behavior, which is a set of trajectories ($T$).

\begin{table}[ht]
\centering
\small
\resizebox{0.85\columnwidth}{!}{%
\begin{tabular}{|p{2.5cm}|p{7cm}|}
\hline
\textbf{Parameter} & \textbf{Description (value used in experiments)}                      \\ \hline
$k$                & Summary budget, i.e., number of trajectories (5)                                  \\ \hline
$l$                & Length of each trajectory (40)                                        \\ \hline
$numSimulations$   & The number of simulations run by HIGHLIGHTS (50)                      \\ \hline
$intervalSize$     & Minimal number of states between two trajectories in the summary (50) \\ \hline
$statesAfter$      & Number of states following $s$ to include in the trajectory (10)      \\ \hline
\end{tabular}}
\caption{Parameters of the HIGHLIGHTS algorithm and the values assigned to them in the experiments reported in ~\cite{amir18highlights} (in parentheses).}
\label{tb:parameters}
\end{table}

The algorithm maintains two data structures: $T$ is a priority queue (line 2), which will eventually hold the trajectories chosen for the summary; $t$ is a list of state-action pairs (line 3), which holds the current trajectory the agent encounters. The procedure runs simulations of the agent acting in the domain. At each step of the simulation, the agent takes an action based on its policy and advances to a new state (line 8). That state-action pair is added to the current trajectory (line 11). If the current trajectory reached its maximal length, the oldest state in the trajectory is removed (lines 9-10).  HIGHLIGHTS computes the importance of $s$ based on the Q-values of the agent itself, as defined in Equation~\ref{eq:importance} (line 14). 

If a sufficient number of states were encountered since the last trajectory was added to the summary, state $s$ will be considered for the summary (the $c==0$ condition in line 17). State $s$ will be added to the summary if one of two conditions hold: either the size of the current summary is smaller than the summary size budget, or the importance of $s$ is greater than the minimal importance value of a state currently represented in the summary (line 17). If one of these conditions holds, a trajectory corresponding to $s$ will be added to the summary. The representation of a trajectory in the summary (a $summaryTrajectory$ object) consists of the set of state-action pairs in the trajectory (which will be presented in the summary), and the importance value $I_{s}$ based on which the trajectory was added (such that it could be compared with the importance of states encountered later). This object ($st$) is initialized with the importance value (line 20) and is added to the summary (line 21), replacing the trajectory with minimal importance if the summary reached the budget limit (lines 18-19). Because the trajectory will also include states that follow $s$, the final set of state-action pairs in the trajectory is updated later (lines 15-16). Last, we set the state counter $c$ to the interval size, such that the immediate states following $s$ will not be considered for the summary. At the end of each simulation, the number of runs is incremented (line 24). The algorithm terminates when it reaches the specified number of simulations. 

\begin{algorithm}
\SetAlFnt{\small\sf} 
\DontPrintSemicolon 
\KwIn{$\pi, k, l, numSimulations, intervalSize, statesAfter$}
\KwOut{$T$}
$runs = 0$ \\
$T \leftarrow PriorityQueue(k, importanceComparator)$ \\
$t \leftarrow$ empty list \\
$c = 0$ \\
\While {$(runs < numSimulations)$} {
$sim = InitializeSimulation()$ \\
\While {$(!sim.ended())$} {
$(s,a) \leftarrow sim.advanceState(\pi)$ \\
\If{$(|t| == l)$} {
$t.remove()$ 
}
$t.add((s,a))$ \\
\If {$(c>0)$} {
$c = c-1$
}
$I_{s} \leftarrow computeImportance(\pi,s)$ \\
\If{$(IntervalSize - c == statesAfter)$} {
lastSummaryTrajectory.setTrajectory(t) \\
}
\If{($(|T|<k)$ or $(I_{s} > minImportance(T)))$ and $(c==0))$ } {
\If{$|T|==k$} {
T.pop()
}
$st\leftarrow$ new $summaryTrajectory(I_{s})$ \\
$T.add(st)$ \\
$lastSummaryTrajectory \leftarrow st$ \\
$c = intervalSize$ \\
}
}
runs = runs+1
}
\caption{The HIGHLIGHTS algorithm. }
\label{alg:highlights}
\end{algorithm}

Originally, HIGHLIGHTS was implemented in an online algorithm  because it is less costly, both in terms of runtime and in terms of memory usage. 
 In addition, such an algorithm can be incorporated into the agent's own learning process without additional cost. In this paper, we adapt the algorithm to work offline, as described in Section~\ref{sec:implementation}.
 
\subsection{Considering State Diversity: the HIGHLIGHTS-DIV algorithm}
\label{sec:algDiv}
Because HIGHLIGHTS  considers the importance of states in isolation when deciding whether to add them to the summary, the produced summary might include trajectories that are similar to each other. This could happen in domains in which the most important scenarios tend to be similar to each other. To mitigate this problem, we developed a simple extension to the HIGHLIGHTS algorithm, which we call HIGHLIGHTS-DIV. Similarly to HIGHLIGHTS, this algorithm also determines which states to include in the summary based on their importance. However, it also attempts to avoid including a very similar set of states in the summary, thus potentially utilizing the summary budget more effectively. 

HIGHLIGHTS-DIV takes into consideration the diversity of states in the following way: when evaluating a state $s$, it first identifies the state most similar to $s$ that is currently included in the summary\footnote{We assume that distance metric to compare states can be defined. This can be done in many domains,  e.g., by computing Euclidean distance if states are represented by feature vectors.}, denoted $s'$. Then, instead of comparing the importance of a state to the minimal importance value that is currently included in the summary, HIGHLIGHTS-DIV compares $I_{s}$ to $I_{s'}$. If $I_{s}$ is greater than $I_{s'}$, the trajectory which includes $s'$ in the summary will be replaced with the current trajectory (which includes $s$). This approach allows less important states to remain represented in the summary (because they will not be compared to some of the more important states that differ from them), potentially increasing the diversity of trajectories in the summary and thus conveying more information to users. 

\subsection{Empirical evaluation of HIGHLIGHTS and HIGHLIGHTS-DIV}
We summarize the main results of the study conducted in our previous work, which demonstrated the usefulness of HIGHLIGHTS and HIGHLIGHTS-DIV summaries. For complete details of the study design and its results see Amir \& Amir~\cite{amir18highlights}. The performance of the basic HIGHLIGHTS algorithm was compared with that of two baselines: (1) random summaries generated by sampling $k$ trajectories uniformly from the agent's execution trace, and (2) summaries generated from the first $k$ trajectories the agent encounters. The task used in the study was identifying the agent that performs better in pairwise comparisons, based on the summaries. Three Ms. Pacman agents were trained varying in their quality: a high-quality agent, medium-quality agent and low-quality agent. This was achieved by varying the number of training episodes. 

In the first experiment, 40 participants recruited from Amazon Mechanical Turk (23 female, mean age = 35.35, STD = 10.4), were asked to make the pairwise agent comparisons based on summaries generated by either the basic HIGHLIGHTS algorithm or one of the two baselines (Random or First). The study used a within-subject design, such that each participant completed nine comparison tasks showing all combinations of pairs of agents and the summary method (e.g., comparing the high-quality agent to the low quality agent based on the HIGHLIGHTS summary). In the second experiment 48 additional participants (25 female, mean age=36, STD=11.6), performed the same task, but this time summaries were generated either by HIGHLIGHTS-DIV, basic HIGHLIGHTS or the random baseline (since the ``first'' baseline led to the worst performance in the first experiment). In both experiments, participants were incentivized to answer correctly as they received a bonus payment depending on their performance. 

\begin{figure}
    \centering
    \includegraphics[width=0.9\linewidth]{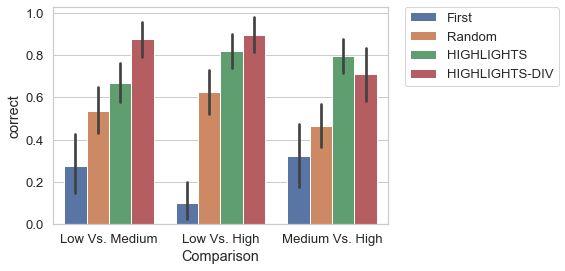}
    \caption{Correctness rates of participants (aggregated from both experiments) in choosing the better performing agent. The x-axis shows the three different pairwise agent comparison tasks (low quality agent vs. medium quality agent, etc.). In all cases, HIGHLIGHTS and HIGHLIGHTS-DIV outperformed the two baselines. HIGHLIGHTS-DIV led to significant improvement over HIGHLIGHTS only in the Low Vs. Medium agent comparison, which was the most difficult comparison as the two agents were most similar to each other in performance.}
    \label{fig:highlights_study}
\end{figure}

Results aggregated from both experiments are shown in Figure~\ref{fig:highlights_study}. Both HIGHLIGHTS and HIGHLIGHTS-DIV summaries led to significantly improved performance of participants compared to the baselines. HIGHLIGHTS-DIV further led to improved performance compared to HIGHLIGHTS, especially when comparing the medium quality agent with the high quality agent, which was the hardest comparison to make as their actual performance did not differ by much. Participants also expressed a subjective preference to HIGHLIGHTS summaries compared to baselines. 

%% file: implementation.tex
\section{Integrating Local and Global Information}
\label{sec:implementation}

In this section, we describe our integration of the local LRP-argmax saliency maps described in section \ref{sec:argmax} into the global HIGHLIGHTS-DIV summaries described in section \ref{sec:algDiv}.
To this end, we describe how the agents were trained, what adjustments we made to  HIGHLIGHTS-DIV for the deep reinforcement learning algorithm we used, how we generated saliency maps and how the combined information is displayed.

\paragraph{Agent training}
To evaluate our combined explanation approach, we trained several Pacman agents using the OpenAI baselines \cite{baselines2017} implementation of the DQN-algorithm \cite{mnih2015human}.
The network architecture used in this implementation is described in \ref{sec:argmax_foundations}.
The environment we use is the Atari 2600 game MsPacman included in the Arcade Learning Environment (ALE) \cite{bellemare13arcade}, which we refer to in this work as Pacman for simplicity.

For each step in a game, the input state consists of the four last frames (the raw pixel values of a single screen of the game) $f_{1}$ to $f_{4}$.
Each frame $f_{i}$ is converted to grey scale and scaled down to $84\times84$ pixels. 
The frames are then stacked to enable the agent to see temporal differences (i.e. movement).
The agent  chooses an action only every four frames and this action is repeated for the next four frames.
In Pacman, the agent has nine different actions to choose from, which correspond to the meaningful actions that can be achieved with an Atari 2600 controller (do nothing, up, down, left, right, up-left, up-right, down-left, down-right).

The reward is based on the ALE \cite{bellemare13arcade} reward function that uses the increase of the in-game score at the beginning of the four frames of a state compared to the score after those frames.
The final reward functions we used are detailed in section \ref{sec:study_design}, when we describe the agents used in the empirical evaluation.

\paragraph{Generating gameplay streams and saliency maps}
Since deep neural networks increase the time that the agent needs for each prediction and the LRP-analysis of this decision requires additional computation time, we recorded a stream of $10,000$ steps for each agent and used them to create our summaries. 
These streams also increase the reproducibility of our experiments\footnote{Since the streams are fairly big we did not upload them. They are available upon request from the authors.}.

We computed the average in-game score of each trained agent over the entire stream.
This allows us to objectively say which agent achieved the most points during the simulations used for our summaries and therefore gives us a ground-truth for the agent comparison task (see section \ref{sec:study_design}).

Since the Atari 2600 version of Pacman does not respond to input for the first 250 frames (empirically tested) after the game starts, we exclude those frames from the streams.
Furthermore, we force the agent to repeat the `do nothing' action for a random amount of steps between $0$ and $30$, until it is allowed to choose actions based on its policy.
This method introduces randomness into the deterministic Pacman game and is also used during training by the DQN algorithm \cite{mnih2015human,baselines2017}.
Saliency maps are created using the LRP-argmax algorithm described in Section~\ref{sec:argmax}.

\paragraph{Adjustments to HIGHLIGHTS-DIV}
For the summaries, we make several adjustments to the HIGHLIGHTS-DIV algorithm described in Section~\ref{sec:algDiv}, to adapt it to the DQN settings.
First, we change the way importance is calculated.
Instead of using equation \ref{eq:importance} which calculates the importance by comparing the highest with the lowest Q-value, we use the difference between the highest and second highest Q-values. 
Let $\operatorname{second highest}$ be the operation that finds the second highest value in a set, then this can be written as:
\begin{equation}
\label{eq:second_importance}
I(s)=\max\limits_{a}Q^{\pi}_{(s,a)}-\operatornamewithlimits{second highest}\limits_{a}Q^{\pi}_{(s,a)}
\end{equation}

While examining the gap between the best and worst actions worked well in a simpler Pacman environment in which there were only four possible actions, it did not generalize well to the Atari environment where there is a larger number of actions.
One possible explanation for this is that some of the 9 actions of the Pacman environment overlap.
For example, ``left" and ``top left" can be used interchangeably in many states.
Therefore the agent might ignore some of the actions completely. To verify this, we examined the frequency of choosing each action, and found that two of the three agents we trained were clearly biased against certain actions.\footnote{The results can be seen in \url{https://github.com/HuTobias/HIGHLIGHTS-LRP/tree/master/action_checks}} Therefore, some Q-values are largely uninformed by exploration and might have arbitrarily low values, making the worst Q-value non-informative.
For the diversity computation in HIGHLIGHTS-DIV, we use Euclidean distance over the raw $84\times84\times4$ input states.

Since we pre-generated a stream of $10,000$ states, we implement an offline version of HIGHLIGHTS-DIV that selects the states for the summary retrospectively from the generated stream.
The procedure begins by sorting the states based on their Q-values, and adding them to the summary according to this ordering.
To reduce the number of overlapping trajectories, we compare each new state with all states in the current summary and corresponding context states (this is equivalent to the HIGHLIGHTS-DIV variant).
To find a suitable threshold that determines when a state is too similar to the states that were already selected for the summary, we randomly pick a subset of $1,000$ random states from the recorded stream and calculate the similarity between each pair of states in this set. 
Then, we set the threshold to be a percentile of the distribution of those similarity values.
We empirically found (by manually examining a sample of states) that using a threshold of $3\%$ led to no obvious duplicate trajectories for any of the agents.

\paragraph{Video generation}
The videos we generate from the states chosen by the summary show $30$ frames per second.
To emphasize that demonstrations show different trajectories, they are separated by a black screen that appears for $1$  second (inspired by the fade-out effect used in  ~\cite{sequeira2019interestingness}).
To prevent the users from using the in-game score to gauge how good an agent is, we mask the bottom half of the screen with black pixels.
In pilot studies, participants complained that the videos were flickering too much.
One of the reasons for this is that the Atari 2600 implementation of Pacman does not show every object in every frame to save computing power.
Since we showed all frame after each other these objects appeared to be blinking and distracted the viewers.
To combat this problem, we do not display the current frame $f_{i}$. 
Instead we display $\max(f_{i},f_{i-1})$, the maximum of each pixel over the current frame $f_{i}$ and the preceding frame $f_{i-1}$.
While this introduces some artifacts (e.g. red pellets showing through blue ghosts) it considerably reduces the flickering. 

Another measure we take against this flickering is to interpolate between the different saliency maps instead of showing a completely different saliency map for each frame.
Let $f_{1}$ to $f_{4}$ be the four frames of an input state and let $s_{1}$ to $s_{4}$ be the saliency maps for each of these frames that analyze the agent's decision in this state.
For $i<4$ the action that Pacman will take after frame $f_{i}$ is not related to the saliency map $s_{i}$, since the agent only decides on a new action every four frames and is still repeating the action that he decided on based on the last state (composed of the 4 frames before $f_1$).
Therefore we show the saliency map $s_{4}$ over the frame $f_{4}$ and for the other frames ($i<4$) we interpolate between the last shown saliency map and $s_{4}$.

Before this interpolation we normalize the saliency maps to have a maximum of $1$ and a minimum of $0$. 
We do this over all 4 frames of the states $s_{1},...,s_{4}$ to avoid losing information that might be transported in the magnitude of relevance values between the frames.

Finally, we add the interpolated saliency maps to the green channel of the original screen frame.
Our complete implementation can be found here: \url{https://github.com/HuTobias/HIGHLIGHTS-LRP}

%% file: study_design.tex
\section{Empirical Evaluation}
\label{sec:study_design}
To evaluate our hypothesis that there is benefit to combining global and local explanations of RL agents, we conducted a user study. In this study, participants were asked to compare different agents and to reflect on the strategies of agents based on the information they were shown. We next describe in detail the study design, the specific hypotheses we tested, and the metrics we used to evaluate the results.

\subsection{Study Design}

\paragraph{Empirical domain}

We used the Atari game Pacman for our experiments (see section \ref{sec:implementation} for the specific implementation).
Atari games are a common benchmark for state of the art reinforcement learning algorithms \cite{bellemare13arcade,baselines2017,Mnih15,wang2015dueling} and to test explanation methods for those algorithms \cite{amir18highlights,greydanus2018,huber2019enhancing,lapuschkin2019,weitkamp2019}.
We chose Pacman since it is not as reaction-based as some other Atari games (e.g. Breakout or Enduro) and allows the RL agents to develop different strategies.
Furthermore, no additional domain knowledge is necessary to understand Pacman and the rules are not too complicated.
This enables us to conduct a study with a wide range of participants by simply explaining the rules at the beginning of the study.

In the game, Pacman obtains points by eating food pellets while navigating through a maze  and escaping ghosts.
There are two types of pellets: regular pills for which Pacman receives 10 points, and power pills that are worth 50 points and also turn the ghosts blue, which makes them edible by Pacman. Pacman receives 200, 400, 800, 1600 points for each ghost it eats successively. 
At random intervals cherries spawn and move through the labyrinth.
Eating a cherry gives 100 points.

To evaluate participants' ability to differentiate between alternative agents and analyze their strategies, we trained agents that behave qualitatively different. To this end, we modified the reward function used for training (similar approach to that used by Sequeira et al.~\cite{sequeira2019interestingness}), resulting in three types of agents.
As mentioned in section \ref{sec:implementation}, we based all of those reward functions on the default ALE~\cite{bellemare13arcade} reward function, which measures the increase in in-game score (as described above) between the first and last frame of a state.

\begin{itemize}
    \item \agentReg: This agent was trained using the default reward function of the ALE
    \footnote{To remove unnecessary magnitude we divided the rewards by the factor 10, such that a regular pill gives a reward of 1.}.
     \item \agentPower: This agent was trained using a reward function that only assigned positive rewards to eating power pills\footnote{We achieved this by only giving the agent a reward if the increase in score was between 50 and 99. The range is necessary since Pacman is forced to eat at least one regular pill directly before it eats a power pill.}.
    \item \agentGhost: This agent used the default ALE reward function but was given an additional negative reward of $-100$ when being eaten by ghosts, causing it to more strongly fear ghosts (which is implicitly learned by other agent due to the lack of future rewards caused by being eaten).
\end{itemize}

Each agent was trained for 5 Million steps with with the algorithm described in section \ref{sec:implementation}.
At the end of this training period the best performing policy is restored.

\paragraph{Experimental conditions}
To evaluate the potential benefits of integrating global and local explanations, and their relative importance, we assigned participants to four different conditions (summarized in Table~\ref{tb:conditions}). The first two conditions included only global information, while the remaining two conditions integrated local explanations as well:
\begin{itemize}
    \item \textbf{Random Summaries (\conditionR)}: In this condition, participants were shown summaries that were generated by randomly selecting state-action pairs from the streams of the Pacman agents playing the game. We note that since each state had the same probability of being chosen, in practice states that are encountered more frequently will be more likely included. Hence, this is equivalent to selecting states based on the likelihood of encountering them. To ensure that the randomly generated summary was not, by chance, particularly good or particularly bad, we generated 10 different random summaries and randomly assigned them to participants in this condition. 
    \item \textbf{HIGHLIGHTS-DIV summaries (\conditionH)}: In this condition, participants were shown summaries generated by the HIGHLIGHTS-DIV algorithm. The specific implementation of this algorithm and the parameters we used for diversity are described in section \ref{sec:implementation}. 
    \item \textbf{Random Summaries+Saliency (\conditionRS)}: These summaries included the same states as those shown in the \conditionR{} summaries, but each image was overlayed with a saliency map generated by the LRP-argmax algorithm described in section \ref{sec:argmax}.
    \item \textbf{HIGHLIGHTS-DIV summaries+Saliency (\conditionHS)}: These summaries included the same states as those shown in the \conditionH{} summaries, where each image was overlayed with a saliency map generated by the LRP-argmax algorithm described in section \ref{sec:argmax}.
\end{itemize}

\begin{table}
\begin{tabular}{| l |c | c |}
\hline
  & `Random' summaries & HIGHLIGHTS-DIV \\ \hline
No saliency maps & \conditionR{} & \conditionH{} \\ \hline
LRP saliency maps & \conditionRS{} & \conditionHS{} \\ \hline
\end{tabular}
\caption{The four study conditions.}
\label{tb:conditions}
\end{table}

We used a budget of $k=5$ for the summaries. That is, each summary included 5 base states chosen either randomly or by HIGHLIGHTS-DIV, where for each state we included a surrounding context window of 10 states that occurred right before and after the chosen state and an interval size of 10 states to prevent directly successive states in the summary.

The video creation and saliency map overlay process is described in detail in section \ref{sec:implementation}.
All video summaries used in the study are available online.\footnote{\url{https://github.com/HuTobias/HIGHLIGHTS-LRP/tree/master/Survey_videos}}.

We note that we did not include a condition that shows only local explanations, since by definition a local explanation is given for a specific state, forcing us to make some choice about which states to show (which means making a global decision). However, the \conditionRS{} condition simulates a scenario where local explanations are shown for randomly selected states.

\paragraph{Participants}
We recruited participants through Amazon Mechnical Turk ($N=134$, the majority of participants were between the ages of 25 and 44,
47 females). 
Participation was limited to people from the US, UK, or Canada (to ensure sufficient English level) with task approval rate greater than 97\%. Since saliency maps are not designed for color blind people, the participants were also asked if they were color blind and stopped from participating if they are.

\paragraph{Procedure}
Participants were first asked to answer demographic questions (age, gender) and questions regarding their experience with Pacman and their views on AI.
Then, they were shown a tutorial explaining the rules of the game Pacman and were asked to play the game to familiarize themselves with it.
To verify that participants understood the rules, they were asked to complete a quiz, and were only allowed to proceed with the survey after answering all questions correctly.
After completing the quiz, they were given information and another quiz regarding the Pacman agent video summaries.
In conditions \conditionRS{} and \conditionHS{}, this also included an explanation and a quiz about saliency maps. Then, they proceeded to the main experimental tasks. See \ref{appendix:questionnaire} for the complete questionnaire.
Participants were compensated as follows: they received \$4 base payment, and an additional bonus of 10 cents for each correct answer. The study protocol was approved by the Institutional Review Board at the Technion. 

\paragraph{Main tasks}
\label{sec:main_tasks}

We aimed to investigate three aspects related to the participants in the study: (1) the mental model of the participant about the agent, (2) participants' ability to assess agents' performance (appropriate trust), and (3) participants' satisfaction with respect to the explanations presented. 

\begin{figure}[ht]
    \centering
    \includegraphics[width=0.8\linewidth]{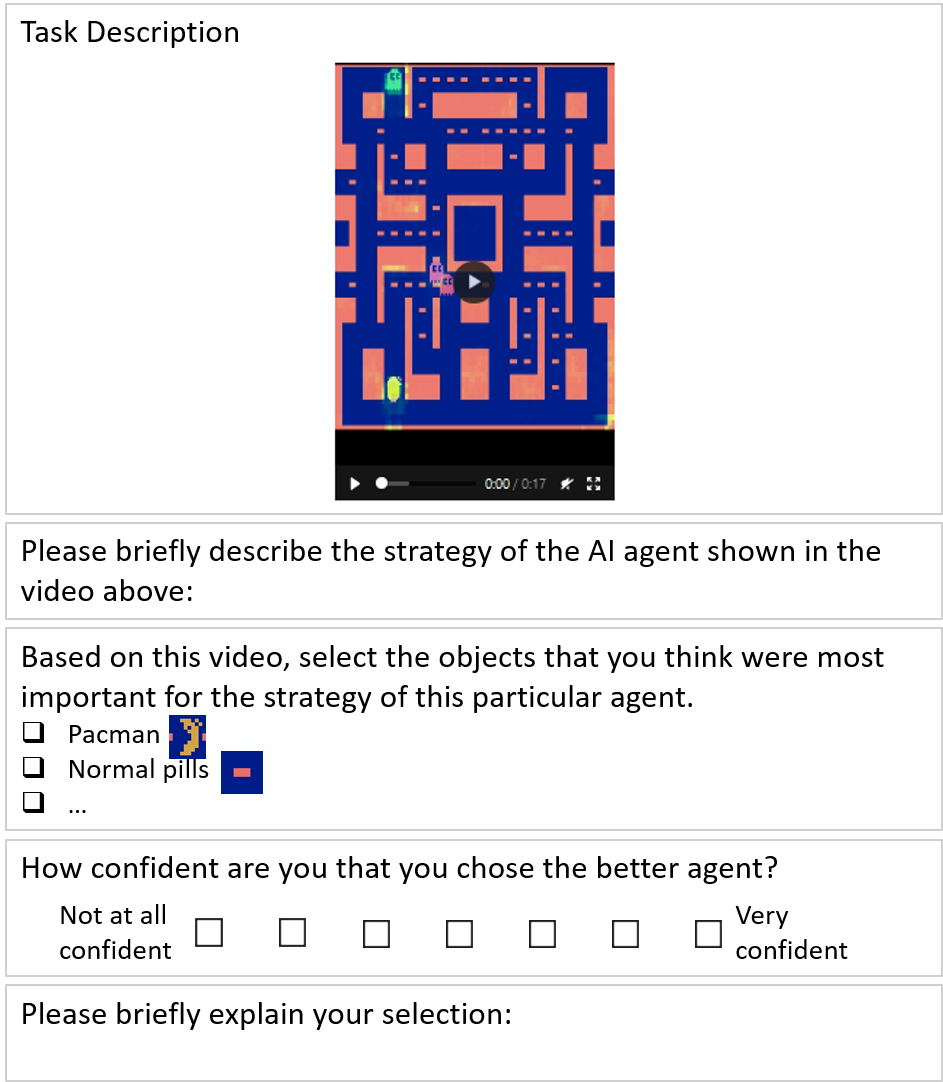}
    \caption{A sketch of the \taskone{}: participants were asked to analyze the behavior of each agent by providing a textual description of its strategy and identifying the objects that are most important to its decision-making.
    The full task can be seen in \ref{appendix:questionnaire}.}
    \label{fig:retro_task}
\end{figure}

\textbf{Task 1: Eliciting Mental Models through Retrospection.} 
By mental model, we understand the cognitive representation that the participant has about a complex model~\cite{halasz1983mental, norman2014some}, in our case, the agent. 
Humans automatically form mental models of agents based on their behavior~\cite{anjomshoae2019explainable}.
These mental models help users understand and explain an agent's behavior.
The examination of participants' mental models and their correctness helps to verify if explainable AI has been successfully applied \cite{Rutjes2019AIHCI,arrieta2020XAIconcepts}.
To evaluate which mental models participants have formed about the agent's behavior, we designed a \textbf{\taskone{}}. 
Here we used a task reflection method inspired by prior studies~\cite{anderson2019mere-mortals,sequeira2019interestingness}, which is recommended by Hoffman et al.~\cite{hoffman2018metrics}.
This task asked the participants to analyze the behavior of the three different AI agents, \agentReg, \agentPower{} and \agentGhost. 
The ordering of the agents was randomized. Specifically, participants were shown the video summary (according to the condition they were assigned to), and were asked to  briefly describe the strategy of the AI agent (textual), and to select up to 3 objects that they think were most important to the strategy of the agent (the possible objects were Pacman, power pills, normal pills, ghosts, blue ghosts and cherries). They were also asked how confident they were in their responses, and to justify their reasoning. Figure~\ref{fig:retro_task} shows a sketch of a retrospection task. 

\begin{figure}[ht]
    \centering
    \includegraphics[width=0.8\linewidth]{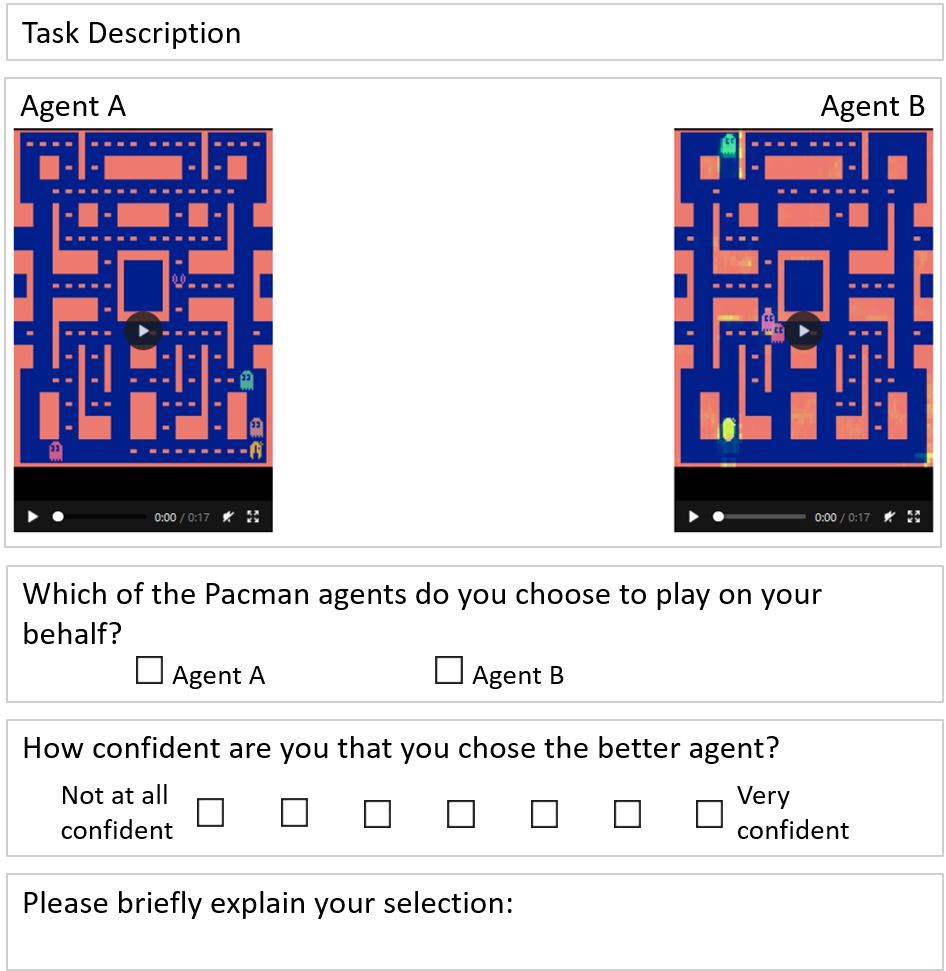}
    \caption{A sketch of the \tasktwo{}: participants were asked to choose which agent they would like to play on their behalf (i.e, identify the better performing agent) according to the two summary videos.
    The full task can be seen in \ref{appendix:questionnaire}.}
    \label{fig:trust_task}
\end{figure}

\textbf{Task 2: Measuring Appropriate Trust through Agent Comparison.} We use the term appropriate trust, based on the work of Lee and See~\cite{lee2004trust} who present a conceptual `trust in automation' framework. They define appropriate trust as a well-calibrated trust that matches the true capabilities of a technical system.
We measure the appropriate trust using an \textbf{\tasktwo{}}. Here, the participants were shown summaries of two of the three agents at a time, and were asked to indicate which agent performs better in the Pacman game (similar to tasks used in ~\cite{amir18highlights,selvaraju2016grad-cam}).
They thus made three comparisons (\agentReg{} Vs. \agentPower, \agentReg{} Vs. \agentGhost and \agentPower{} Vs. \agentGhost).
We do not ask the participant directly about their trust in the two agents shown. 
Instead, the participants have to choose one of the two agents that they would like to to play on their behalf (see Figure \ref{fig:trust_task}).
This implicit question reveals which agent participants consider more reliable and qualified for the task.
As in the retrospection task, they were asked to indicate their level of confidence and to provide a textual justification for their decision.
The ordering of the three agent comparisons was randomized.

\textbf{Explanation satisfaction questions.} 
Miller~\cite{miller2018explanation,miller2017explainable} argues that the end users' impressions about the agent should be queried and included into the evaluations of the explainable AI methods.
This would ensure that the developed explanation methods are comprehensible not only to ML-experts but also to end-users.
We address this concern in our study by measuring participants' subjective satisfaction.
To this end, we used \textbf{explanation satisfaction questions} adapted from the questionnaire proposed by Hoffman et al.~\cite{hoffman2018metrics}. 
We did this separately for the \taskone{} (immediately after completing the three retrospection tasks) and for the \tasktwo{} (after completing the three comparisons), as we hypothesized there may be differences in the usefulness of the summaries for these two different types of tasks. 
Specifically, participants were asked the following questions using a 5-point Likert scale:
\begin{enumerate}
    \item From watching the videos of the AI agents, I got an idea of the agents' strategies. 
    \item The videos showing the AI agents play contain sufficient detail about the agents' behavior.
    \item The videos showing the AI agents play contain irrelevant details.
    \item The videos showing the AI agents play were useful for \emph{the task}. (only shown in groups \conditionR{} and \conditionH)
    \item The gameplay scenarios shown in the videos were useful for \emph{the task}.
    (only shown in groups \conditionRS{} and \conditionHS)
    \item The green highlighting in the videos was useful for for \emph{the task}.
    (only shown in groups \conditionRS{} and \conditionHS)
\end{enumerate}
We substituted \emph{the task} with either \emph{anlayzing the agents' behavior} or \emph{choosing the agent that performs better}, depending on the  task they had just completed. 

\subsection{Hypotheses}
\label{sec:hypotheses}
Overall, we hypothesized that HIGHILIGHTS-DIV summaries will be more useful than random summaries in both the retrospection and agent comparison tasks, and that adding saliency maps will further improve participants' performance. More specifically, we state the following hypotheses:
\begin{itemize}
    \item H1: For both tasks, participants shown summaries generated by HIGHLIGHTS-DIV will perform better than participants shown randomly generated summaries. That is, performance in \conditionH{} will be better than performance in \conditionR{} and performance in \conditionHS{} will be better than performance in \conditionRS. We expect HIGLIGHTS-DIV summaries to be more useful as they demonstrate the agent's behavior in more meaningful states, which should help both in identifying which agent performs better (in line with prior findings~\cite{amir18highlights,huang2018establishing}), as well as in determining whether an agent is capable of performing well in certain scenarios~\cite{huang2018establishing}. 
    We expect similar effects in terms of  participants' explanation satisfaction in each task.
    
    \item H2: For both tasks, adding saliency maps will improve participant's performance and satisfaction. That is, we expect the performance in \conditionRS{} will be better than in \conditionR{} and similarly that performance in \conditionHS{} will be better than in \conditionH.
    Here, too, we expect similar effects in terms of participants' explanation satisfaction in each task.
    We expect this to be the case as the saliency maps allow people to see not only what actions the agent chooses, but also what information it attends to.
    Previous studies also found positive effects of saliency maps on participants' mental models \cite{anderson2019mere-mortals,alqaraawi2020evaluating} and on their ability to choose the better performing prediction model \cite{selvaraju2016grad-cam}.
    
    \item H3: The effect of the summary generation method on satisfaction and performance will be greater than that of the inclusion of saliency maps in the agent comparison task. That is, we expect that global information will be more crucial for identifying the better performing agent, as it explicitly demonstrates how the agents act.
    \item H4: The effect of adding saliency maps on satisfaction and performance will be stronger than that of the summary generation method in the retrospection task. Since saliency maps explicitly show what information the agent attends to, we hypothesize it will contribute more to identifying the agent's strategy. However, this is complicated by the fact that random summaries might not include interesting scenarios, making saliency maps less helpful in this case.  Therefore, our more specific hypothesis are:
    \begin{itemize}
        \item H4.1: Participants in the saliency conditions will be more likely to identify Pacman, the main source of information for our agents, 
        as an important object.
        \item H4.2: Participants in the HIGHLIGHTS conditions will be more likely to identify objects that relate to agent goals, such as power pills and blue ghosts. Therefore, they will also more accurately describe the agents' strategies.
    \end{itemize}
\end{itemize}

\subsection{Analysis} 
\label{sec:analysis}
We analyze the main hypotheses using the the non-parametric Mann-Whitney test~\cite{mcknight2010mann}, as our dependent variables are not normally distributed.
We report effect sizes using rank biserial correlation~\cite{tomczak2014need}.
Additionally, we report the mean values and the 95\% confidence interval (CI) computed using the bootstrap method. In all plots the error bars correspond to the 95\% confidence intervals.

To make sure that the participants involved in our analysis did in fact watch the videos of the agents, we recorded whether they clicked play on each video in addition to how often each video was paused.
We did not force them to watch the videos to filter out participants that would have just pressed play to avoid the forcing mechanism.
Since we saw from the raw data that some participants only stopped watching videos after the \taskone{}, we checked each task separately.
As a heuristic to measure how attentively a user watched the videos of a task, we took the sum of pauses of the videos in this task, where watching a video until the end was recorded as a pause and not clicking play was counted as $-1$ pause.
Based on this heuristic we removed all participants from the \taskone{} who did not have at least three pauses (5 participants) and all participants from \tasktwo{} who did not have at least six pauses (11 participants).
The number of necessary pauses in each task is equal to the number of videos in this task.

For evaluating the retrospection task we use a scoring system, where two of the authors involved in the training of the agents assigned a score to each item for each agent before the study started (see \ref{appendix:scoring_functions} for details).
For example for the \agentPower{}, which was only rewarded when it ate a Power pill, selecting the Power pill or Pacman  increased the score by $1$ point and including any other item reduced the score by $1$ point.
Furthermore, selecting more than three items resulted in a score of zero, since the participants were told to select a maximum of three items.

Inspired by Anderson et al.~\cite{anderson2019mere-mortals} we use summative content analysis \cite{hsieh2005content_analysis} to evaluate  participants' textual responses.
An independent coder (not one of the authors) classified responses to the questions ``Please briefly describe the strategy of the AI agent shown in the video above'' in the \taskone{}, and the question ``Please briefly explain how you came to your selection'' in both the \taskone{} and the \tasktwo{}.
Each question was asked three times (once for each agent description or agent comparison) resulting in $402$ answers per question.
For the first question, the coder identified $67$ different concepts in the answers.
For example, the answer ``The strategy of this Pacman agents seems to be to mainly avoid the ghosts as it eats the normal pills on the screen. Although it can be seen eating a power pill, the clip still does not show Pacman seeking out and eating the ghosts'' was coded to ``prioritizing normal pills'', ``avoiding ghosts'' and ``do not care about blue ghosts''.
We aggregated those concepts to $16$ groups by combining similar concepts like ``eating normal pills'' and ``prioritizing normal pills''. 

To evaluate the correctness of  participants' answers we implemented a simple scoring system. For each agent and for each answer group, we decided whether it is correct, irrelevant or wrong, based on predefined `ground-truth' answers that two of the authors, who were involved in the training of the agents, wrote for each agent before the study started.
The exact groups and their assigned scores can be found in \ref{appendix:scoring_functions} and the open-sourced code.

The answers to the second question regarding participants' justifications of their responses were classified into six categories (the answer could be based on the game rules, the saliency maps, the gameplay, participants' interpretation and two categories for unjustified or unrelated justifications which we grouped into one ``unjustified'' category) and an additional seventh category for the \tasktwo{}, that encoded that the user could not decide between the two agent and guessed.

We note that the classifications assigned by the coder are not mutually exclusive.

%% file: results.tex
\section{Results}
\label{sec:results}

In this section, we report the results of our study. 
We first describe the characteristics of the participant population with respect to their AI experience, attitude towards AI and Pacman experience. 
Then we assess the main hypotheses (H1--H4) (results summarized in Table~\ref{tb:p_values}) and further provide a descriptive analysis of additional variables such as participants' confidence and analysis of mistakes. 

\paragraph{AI and Pacman experience} We verify that participants in different conditions did not differ much in their AI experience and views and in their experience with the game Pacman.
To this end we asked them when they played Pacman for the last time and
across all four conditions the majority of participants answered: `I played Pacman more than 5 years ago'.
After receiving a short description of what AI is (using a formulation based on Russel~\cite{russell2016artificial}), 104 participants stated that they had experience with AI.
The exact kind of experience ranged from `I know AI from the media' (78 participants) to `I do research on AI related topics' (14 participants). 
On average the users had a positive attitude towards AI (mean of $3.95$ on a 5-point Likert scale).
There are no meaningful differences between the conditions (see \ref{appendix:demographics} for more details).

\begin{figure}
    \small
    \centering         
    \begin{minipage}{0.48\linewidth}
    \centering
    \includegraphics[width=\linewidth]{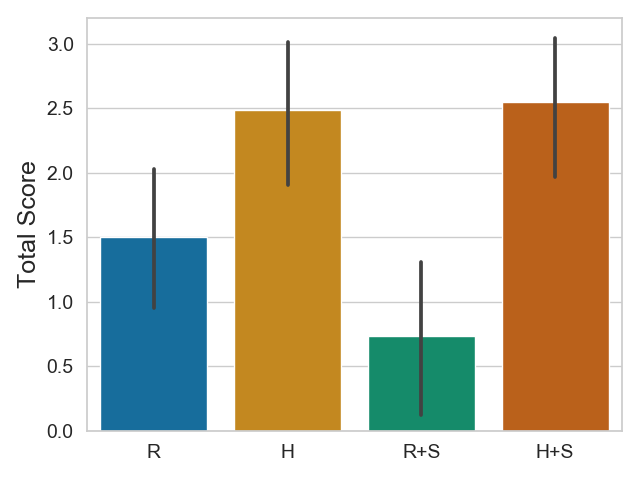}
    (a) Total score (summed over all three agents) for the object selection in the \taskone{}. The scoring system is described in \ref{sec:analysis}.
    \end{minipage}
    \begin{minipage}{0.48\linewidth}
    \centering
    \includegraphics[width=\linewidth]{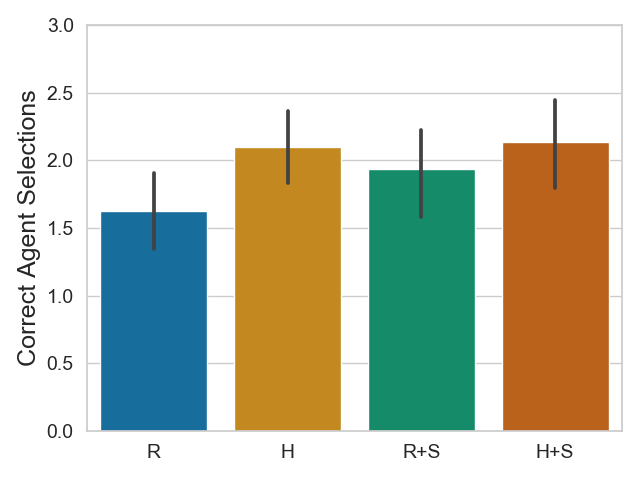}
    (b) Number of correct agent selections in the \tasktwo{} (Out of three selections).
    \end{minipage}
    \caption{Comparison of  participants' average performance in each task, by condition. Participants in the HIGHLIGHTS conditions \conditionH{} and \conditionHS{} outperformed the random conditions \conditionR{} and \conditionRS{}. 
    Saliency maps only had a slight positive effect when added to random summaries in the \tasktwo{} 
    }
    \label{fig:total_score}
\end{figure}

\begin{figure}
    \centering
    \small
    \begin{minipage}{0.48\linewidth}
    \centering
    \includegraphics[width=\linewidth]{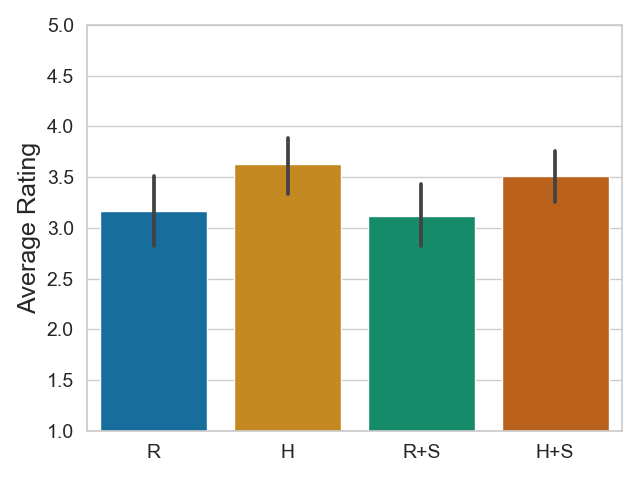}
    (a) Participants' satisfaction in the \taskone{} averaged over all explanations satisfaction questions. 
    \end{minipage}
        \begin{minipage}{0.48\linewidth}
    \centering
    \includegraphics[width=\linewidth]{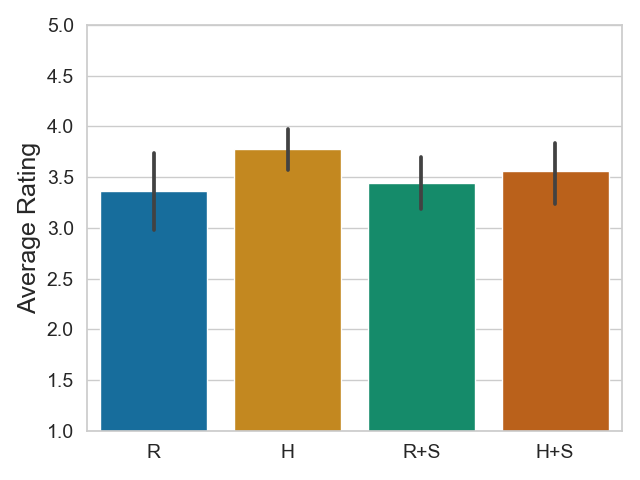}
    (b) Participants' satisfaction in the \tasktwo{} averaged over all explanation satisfaction questions.
    \end{minipage}
    \caption{Comparison of  participants' average explanation satisfaction in each task, by condition.
    Each participant rated their agreement with several statements adapted from the explanation satisfaction questions proposed by Hoffman et al.~\cite{hoffman2018metrics} on a 5-point Likert scale (see Section \ref{sec:main_tasks}).
    Participant's final rating was averaged over all those ratings, reversing the rating of the negative statements. 
    Overall,  participants in the HIGHLIGHTS conditions \conditionH{} and \conditionHS{} rated the explanations highest.}
    \label{fig:total_satisfaction}
\end{figure}

\paragraph{(H1) Participants shown HIGHLIGHT-DIV summaries performed better than participants shown random summaries} 
Participants' correctness rates for the \tasktwo{} are shown in Figure~\ref{fig:total_score}(b). These results support H1, which states that HIGHLIGHTS-DIV summaries will lead to improved performance in both the \tasktwo{} and the \taskone{}.
The exact definition of performance per task is described in more detail in section  \ref{sec:analysis}.
Specifically, in the \tasktwo{} we find that participants in condition \conditionH{} significantly outperformed participants in condition \conditionR{} (\conditionH{}: mean=2.1, 95\% CI=[1.83, 2.33], \conditionR: mean= 1.63, 95\% CI=[1.34, 1.91],  Mann-Whitney test U=334.5, $p=0.014$, \RankBiserialCorrelation{}=0.3)\footnote{Here 95\% CI is the 95\% confidence interval and \RankBiserialCorrelation{} is Rank biserial correlation.}.
While participants in the \conditionHS{} condition achieved higher mean correctness rates than participants in the \conditionRS{} condition, this difference is not statistically significant 
(\conditionHS: mean=0.71, 95\% CI=[0.6, 0.82], \conditionRS: mean=0.65, 95\% CI=[0.54, 0.75],  Mann-Whitney test U=391, $p=0.180$, \RankBiserialCorrelation{}=0.13).
Similarly,  participants' average explanation satisfaction ratings, shown in Fig.~\ref{fig:total_satisfaction}(b), indicate that participants in condition \conditionH{} were more satisfied with the videos they received than the other participants. 
However, this difference is not significant (see Table \ref{tb:p_values}).

\begin{table}{}
\begin{tabular}{l | l | C |C |C |C}

      \textbf{Task} & \textbf{Variable} & \multicolumn{2}{l|}{ \parbox{0.25\linewidth}{\textbf{Effect of strategy summarization:}}} & \multicolumn{2}{l}{\parbox{0.25\linewidth}{\textbf{Effect of saliency maps:}}} \\ 

  &   & \conditionH{} > \conditionR{} & \conditionHS{} > \conditionRS{} & \conditionRS{} > \conditionR{} & \conditionHS{} > \conditionH{} \\
  \hline

\multirow{2}{*}{\parbox{0.17\linewidth}{\taskone{}}} &
    score & 0.008^{*} & 3.3e-05^{*} & 0.965 & 0.514 \\
  & satisfaction & 0.021^{*} & 0.035^{*} & 0.677 & 0.710 \\
  & text score & & & & 0.088^{\dagger} \\
    \hline

\multirow{2}{*}{\parbox{0.17\linewidth}{\tasktwo{}}} &
score & 0.014^{*} & 0.180 & 0.062^{\dagger} & 0.307\\
& satisfaction & 0.147 & 0.235 & 0.627 & 0.833 \\

\end{tabular}{}
\caption{Summary of all significance tests (calculated with Mann-Whitney tests).
The $^{*}$ denotes statistically significant differences and $^{\dagger}$ denotes a p-value $<0.1$.}
\label{tb:p_values}
\end{table}

With respect to participants' performance during the \taskone{}, we find even stronger results (Fig.~\ref{fig:total_score}(a)) then in the \tasktwo{}, further supporting H1.
Here too,  participants in condition \conditionH{} obtained a higher score in the object selection sub-task than participants in condition \conditionR{} (\conditionH: mean=2.5, 95\% CI=[1.89, 3.03], \conditionR: mean=1.5, 95\% CI=[0.92, 2.06],  Mann-Whitney test U=346.5, $p=0.008$, \RankBiserialCorrelation{}=0.34) and participants in the \conditionHS{} condition received a higer score then participants in the \conditionRS{} condition (\conditionHS: mean=2.55, 95\% CI=[2.02, 3.06], \conditionRS: mean=0.73, 95\% CI=[0.13, 1.31],  Mann-Whitney test U=206.5, $p=0.00003$, \RankBiserialCorrelation{}=0.58).
We found analogous significant differences in participants' explanation satisfaction during the \taskone{} (Fig.~\ref{fig:total_satisfaction}(a)). 
Here, participants in condition \conditionH{} were more satisfied than participants in condition \conditionR{} (\conditionH: mean=3.63, 95\% CI=[3.35, 3.88], \conditionR: mean=3.17, 95\% CI=[2.82, 3.5],  Mann-Whitney test U=373.0, $p=0.021$, \RankBiserialCorrelation{}=0.29) and participants in the \conditionHS{} condition were more satisfied than participants in the \conditionRS{} condition (\conditionHS: mean=3.52, 95\% CI=[3.25, 3.78], \conditionRS: mean=3.12, 95\% CI=[2.81, 3.43],  Mann-Whitney test U=364.5, $p=0.035$, \RankBiserialCorrelation{} 0.27).

\begin{figure}
    \centering
    \begin{minipage}{0.6\linewidth}
    \centering
    \includegraphics[width=\linewidth]{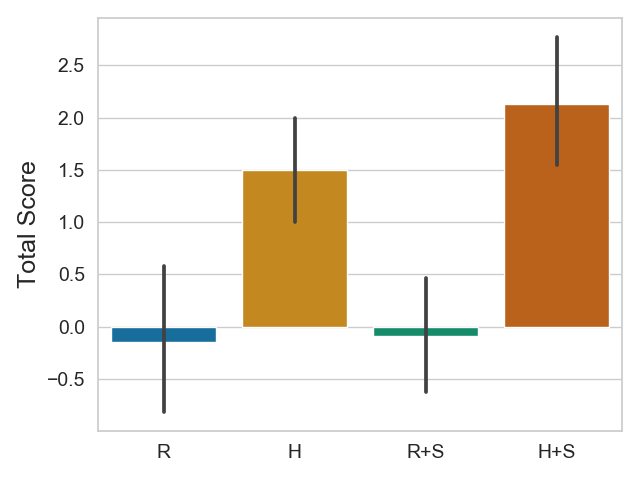}
   \end{minipage}
   \caption{Participants' total  score for their textual descriptions  of the agents strategy during the \tasktwo{} (summed over all three agents). The scoring function is described in \ref{sec:analysis}.
    The descriptions of participants in the HIGHLIGHTS-DIV conditions \conditionH{} and \conditionHS{} received a higher score than those of participants in the random conditions. 
    The addition of saliency maps (\conditionHS{}) slightly improved this effect further.}
   \label{fig:text_score}
\end{figure}

\paragraph{(H2) Adding saliency maps improved performance in some areas depending on the task}
There were no significant differences
supporting our second hypothesis H2 which predicted that adding saliency maps will improve participants' performance in both tasks.
Nevertheless,  we report two positive effects of saliency maps that are only marginally\footnote{In accordance with convention (Vogt et al.~\cite{vogt2005dictionary}), we use \emph{marginally significant} to describe $0.05 \leq p < 0.1$} significant and which might guide future research in this area.
For the \tasktwo{}, we find that the saliency maps only improved performance when added to random summaries (\conditionR: mean=0.54, 95\% CI=[0.45, 0.64], \conditionRS: mean=0.65, 95\% CI=[0.54, 0.75], Mann-Whitney test U=390.5, $p=0.062$, \RankBiserialCorrelation{}=0.21).
Fig.~\ref{fig:total_score}(a) shows that the saliency maps did not help participants identify the most important objects in the \taskone{}.
However, the summative content analysis of participants' textual descriptions of the agents' strategies, shown in Fig~\ref{fig:text_score}, indicates that saliency maps helped participants to correctly describe how the agents use those objects.
The descriptions of the agents' strategies written by participants in condition \conditionHS{} received a higher score than the ones by participants in condition \conditionH{}  (\conditionH{}: mean=1.50, 95\% CI=[0.97, 2.0], \conditionHS{}: mean=2.13, 95\% CI=[1.55, 2.71], Mann-Whitney test U=400, $p=0.088$, \RankBiserialCorrelation{}=0.195).

\paragraph{(H3 + H4) The effect of the summary generation method was greater than that of adding saliency maps} We hypothesized that the summary generation method will affect the performance of participants more than the addition of saliency maps in the \tasktwo{} (H3), and that the saliency maps will have a greater effect than the summary method in the \taskone{} (H4). 
The study results support H3: we found that participants shown HIGHLIGHTS-DIV summaries significantly outperformed participants shown random summaries in the \tasktwo{}, while adding saliency maps only improved performance for the random summaries, and to a lesser extent. 

For selecting the most important objects for the agent's strategy in the \taskone{}, the addition of saliency maps did not improve performance, while HIGHLIGHTS-DIV summaries did improve performance compared to the random summaries. 
Therefore we reject H4, even though the results shown in Fig.~\ref{fig:text_score} indicate that saliency maps improved the textual descriptions of the agent's strategy written by participants in \conditionHS{} compared to \conditionH{}.

\begin{figure}
    \centering
    \footnotesize
   \begin{minipage}{0.48\linewidth}
    \centering
    \includegraphics[width=\linewidth]{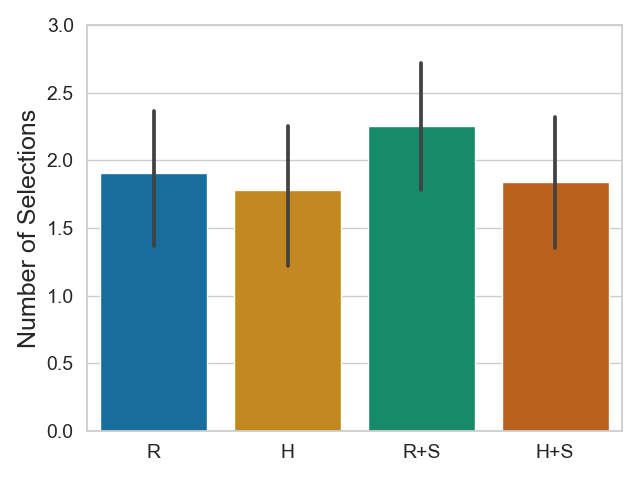}
    (a) Selections of Pacman during the object selection per condition.
    \end{minipage}
     \begin{minipage}{0.48\linewidth}
    \centering
    \includegraphics[width=\linewidth]{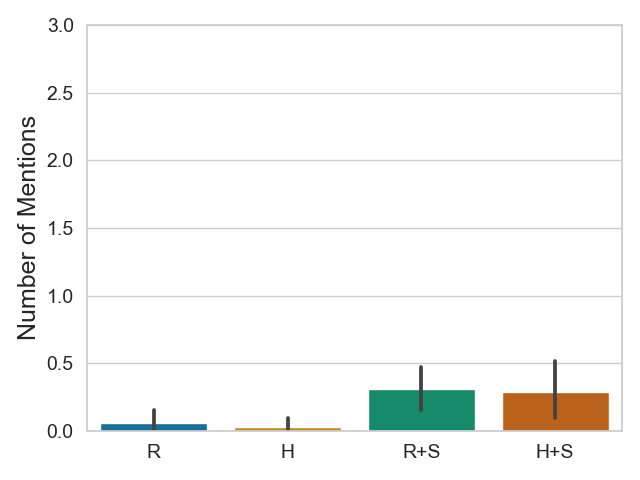}
    (b) Mentions of Pacman's vicinity in the descriptions of the agents' strategies per condition. 
    \end{minipage}
    \caption{The average number of times that participants correctly selected Pacman during the object selection (a), or referred to its vicinity in their textual descriptions (b) of the agents' strategies (sum over all three agents). 
    The results indicate that saliency maps help the participants to identify what information the agents use. 
    }
    \label{fig:retro_pacman}
\end{figure}{}

In line with  Hypothesis H4.1, Fig.~\ref{fig:retro_pacman} indicates that the improvement of the descriptions of the agents' strategies mainly stems from participants in the saliency groups \conditionRS{} and \conditionHS{} identifying that the agent mostly payed attention to the vicinity of Pacman.
This effect was not as strong in the object selection question, since it did not capture the participants' reasoning.

\begin{figure}
    \centering
    \footnotesize
    \begin{minipage}{0.48\linewidth}
    \centering
    \includegraphics[width=\linewidth]{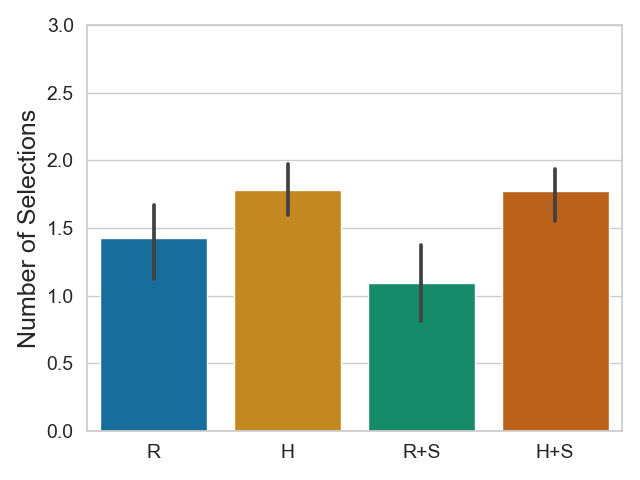}
    (a) Selections of the agent's specific goals in the object selection, per condition.
    \end{minipage}
        \begin{minipage}{0.48\linewidth}
    \centering
    \includegraphics[width=\linewidth]{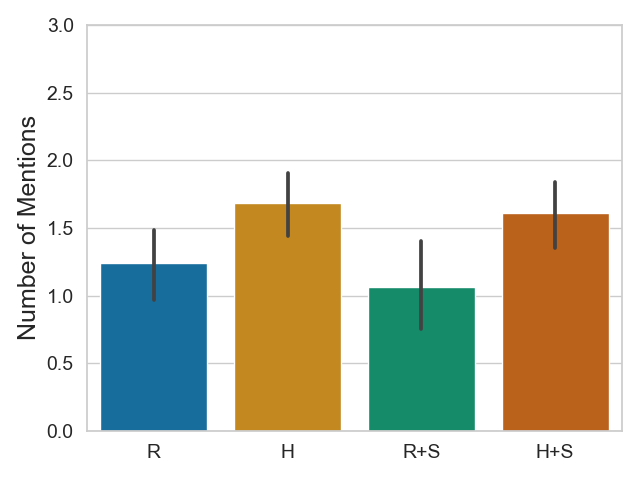}
    (b) Mentions of the agent's specific goals in the strategy descriptions, per condition. 
    \end{minipage}
    \caption{The number of times that participants identified the agent's specific goal in  the object selection (a) and  strategy description (b) components of the \taskone{}. The results are in line with  Hypothesis H4.2 that strategy summarization helps to identify the agents' goals.}
    \label{fig:retro_select_goal}
\end{figure}

Sub-Hypothesis H4.2 stated that strategy summarization would help participants identify the goals of the agents.
The results shown in Fig.~\ref{fig:retro_select_goal} support this Hypothesis, since  participants in the HIGHLIGHTS-DIV conditions \conditionH{} and \conditionHS{} identified the correct goals of the agent more often.

\paragraph{Participants' Justifications}
Across all groups, most participants mainly based their justifications on the agents' gameplay (Fig.~\ref{fig:gameplay_justifications}). 
In the saliency conditions, most participants did not mention the saliency maps in their justifications.
On average, less than one out of 3 justifications in \conditionHS{} and in \conditionRS{} 
referred to the green highlighting during the \taskone{} and during the \tasktwo{} even fewer participants mentioned them (see Fig.~\ref{fig:heatmap_justifications} for more details).

Another interesting point we found in participants' justifications during the \taskone{} is that participants in \conditionH{} gave more unjustified explanations than any other condition (\conditionH{}: mean=0.66, compared to the second highest condition \conditionRS{}: mean=0.38 ). 
This is just an observation and did not repeat in the \tasktwo{} but it might be interesting to investigate further in the future. 
The values for all conditions can be seen in Fig.~\ref{fig:unjustified_justifications}.

\paragraph{Participants' confidence and viewing dynamics} 
In addition to the main metrics used in our study, we further measured participants' confidence (and in particular whether they were more confident when they answered correctly), and their viewing dynamics of the summaries (time and number of pauses). However,  
apart from a slight positive effect for the participants in condition \conditionH{}, there were no interesting differences in the three aforementioned variables (see Fig. \ref{fig:confidences} to \ref{fig:pauses} and ~\ref{appendix:results} for additional details).

%% file: discussion.tex
\section{Discussion \& Future Work}
\label{sec:discussion}

With the increasing use of RL agents in high-stakes domains, there is a growing need in developing and understanding methods for describing the behavior of these agents to their human users.
In this paper, we explored the combination of global information describing agent behavior, in the form of strategy summaries, with local information in the form of saliency maps.
To this end, we augmented HIGHLIGHTS-DIV~\cite{amir18highlights} summaries,  which select important and diverse states (adapted to DQN agents), with saliency maps generated using the LRP-argmax algorithm~\cite{huber2019enhancing}. 

We implemented the combined approach in the Atari Pacman environment, and evaluated the separate and joint benefits of showing users global and local information about the agent.
We used two types of tasks: a
\taskone{} about the agent's strategy and a \tasktwo{}.

\paragraph{Strategy summarization}
The results of this study reinforce our prior findings ~\cite{amir18highlights} showing that summaries generated by HIGHLIGHTS-DIV lead to significantly improved performance of participants in the \tasktwo{} compared to random summaries, and show that this result generalizes to RL agents based on neural networks.
Furthermore, they show that HIGHLIGHTS-DIV summaries were more useful for analyzing agent strategies and were preferred by participants.
Overall, in our study, the choice of states that are shown to participants was more important than the inclusion of local explanations in the form of saliency maps.

\paragraph{Limitations of saliency maps}

With respect to the addition of saliency maps, we found mixed results. 
In contrast to previous studies about saliency maps for image classification tasks, which found weak positive effects for saliency maps \cite{alqaraawi2020evaluating,selvaraju2016grad-cam}, there were no significant differences between the saliency and non-saliency conditions in our study.
When examining participants' answer justifications, we observed that most participants did not mention utilizing the saliency maps, which may provide a partial explanation to their lack of contribution to participants' performance.
Especially in the \tasktwo{}, participants seldom mentioned the saliency maps even though there was a marginally significant difference between performance of participants in condition \conditionR{} and in condition \conditionRS{}. 
Participants' comments also reflect their dissatisfaction with saliency maps, e.g., ``I do not believe that the green highlighting was useful or relevant" and ``The green highlights didn't seem to help much''. 
This suggests that saliency maps in their current form may not be accessible enough to the average user.

Based on the comments from the participants and in depth feedback we received in pilot studies,  we note some possible accessibility barriers.
First, when saliency maps are shown as part of a video, it may be difficult for users to keep track of the agent's attention, compared to displays of static saliency maps, as done in previous user studies \cite{selvaraju2016grad-cam,anderson2019mere-mortals,alqaraawi2020evaluating}.
For instance, one participant reported that ``[i]t wasn't so easy to see the green area, it needed to be bigger or more prominent to be of more use.''
We tried to take measures against this by using a selective saliency map generation algorithm (LRP-argmax) and interpolating between selected saliency maps to reduce the amount of information, as well as allowing participants to pause the video at any time.
However, this does not seem to be enough.

Second, participants were not accustomed to interpreting saliency maps, which can be non-intuitive to non-experts.
One participant even commented that 
``[he/she] feel[s] as though this came with somewhat of a learning curve''.
In our pilot studies we noticed that people who were familiar with reinforcement learning or deep learning could more easily interpret saliency maps than those who were not. 
For example, some participants said that they thought the agent was good when its attention was spread to different areas because they inferred it considered more information, while in fact the agent was attending to different regions because it did not yet learn what the important information is.
Similarly, one study participant commented: ``...I don't know if I would prefer an AI that `looked' around more at the board, or focused more in a small area to accomplish a task''.
It is possible that prior studies which used saliency maps for interpreting image classification~\cite{alqaraawi2020evaluating,selvaraju2016grad-cam} did not encounter this problem due to the more intuitive nature of the task. 
Interpreting a visual highlighting for image classification only requires identifying objects that contributed to the classification, while in RL there is an added layer of complexity as interpretation also requires making inferences regarding how the highlighted regions affect the agent's long-term sequential decision-making policy. 

Finally, while the sanity checks reported in Section \ref{sec:sanity_checks} showed that our saliency maps do analyze what the network learned, they were also found to be indifferent to specific actions. 
Since prior studies have shown that users find class discriminatory explanations more useful for understanding agents' decisions~\cite{goudet2018functioanlcausal,LopezPaz2017causalsignals,byrne2019humanreasoning}, the lack of discrimination between certain actions can be detrimental to the usefulness of saliency maps.

\paragraph{Potential of saliency maps}
Regarding the potential of saliency maps, we made encouraging observations.
Even though saliency maps did not significantly increase  participants' scores in the simple object selection part of the \taskone{}, they did result in improved scores in the textual strategy description.
The difference between our HIGHLIGHTS-DIV conditions \conditionHS{} and \conditionH{} is similar to the one observed by Anderson et al.  \cite{anderson2019mere-mortals} (p=0.086 compared to our p=0.088), who also evaluated participants' mental models for RL agents utilizing a strategy description task.
The poor result of our random condition \conditionRS{} can be explained by the fact that Anderson et al. implicitly chose meaningful states, which we only did with our global explanation method in the HIGHLIGHTS-DIV conditions.

A possible reason for the difference between the object selection and the strategy description sub-tasks is the higher complexity of strategy description. It requires participants to not only identify the correct objects but also to describe how they are used.
Under this assumption, the increased performance of participants in condition \conditionHS{} suggests that saliency maps were useful for putting the objects in the correct context.
For example,  participants' textual descriptions showed that, while the non-saliency groups know that Pacman is important (most likely based on the fact that it is important for them as players), they did not identify it as a central source of information for the agent.

Second, we observed in the \tasktwo{} that saliency maps alone improved participants' ability to place appropriate trust into different agents when comparing conditions \conditionR{} and \conditionRS{}. 
There, performance was comparable to the performance of participants in the HIGHLIGHTS-DIV conditions, \conditionH{} and \conditionHS{}. 
This indicates that there is valuable information for this kind of task within saliency maps.
The lacking improvement of condition \conditionHS{} compared to \conditionH{} might be explained by the accessibility issues of saliency maps mentioned earlier.
When presented with strategy summaries,  participants may have had less reason to rely on the non-intuitive saliency maps.

\paragraph{Combination of local and global explanations}
It is important to note that the  positive effects of saliency maps in the \taskone{} were only visible in the HIGHLIGHTS-DIV conditions \conditionH{} and \conditionHS{}, reinforcing our claim that the choice of states is crucial for explaining RL agents.
Therefore, even if the limitations of saliency maps mentioned above are addressed, the potential benefits might only be visible and likely reinforced by a combination with strategy summarization techniques.  We note that studies that evaluate local explanations 
typically implicitly make a global decision about which states to present local explanations for \cite{anderson2019mere-mortals,madumal2019explainable}.
Our results suggest that this implicit choice may have a substantial impact on participants' understanding of agent behavior.

In the \taskone{}, we observed that local explanations in the form of saliency maps were useful for identifying what objects the agent attends to (see Fig.~\ref{fig:retro_pacman}), while strategy summaries were more useful for identifying the agent's goals (see Fig.~\ref{fig:retro_select_goal}). 
This was reflected by participants' utterances such as: ``The agent seemed to be paying attention to the area directly in front of it and partly to the areas directly to each side.'' and ``Pacman wanted those ghosts! His goal was to move as fast as he could towards them.'' and suggests that the two approaches are indeed complementary.
The local saliency maps contribute to users' understanding of the agents \emph{attention}, as they reflect the information the agent attends to, while strategy summaries contribute to users' understanding of the agent's \emph{intentions}, as they reflect how the agent acts. 

Taken together, our results suggest that there is potential for a combined explanation framework in the future, if the accessibility issues of saliency maps are addressed.

\paragraph{Study limitations}
Our study has several limitations. First, we used a single domain in our user study.
However, other recent work has used strategy summaries similar in spirit to HIGHLIGHTS-DIV in another domain~\cite{sequeira2019interestingness} and several works have used saliency maps in other domains (e.g., several Atari games including Pong and Space invaders were used by Greydanus et al.~\cite{greydanus2018}).

Second, while our combined explanation approach is easily adaptable to other global explanation methods which choose an informative subset of states, and local methods that highlight relevant information in those states, our study only explored one combination of a particular global explanation method and a particular local explanation method. 
We chose the HIGHLIGHTS-DIV summary method since strategy summary approaches that are based on policy reconstruction require making various assumptions about people's computational models, and that these models differ depending on context~\cite{lage2019exploring}.
We chose saliency maps as a local method both because it is visual and thus can be integrated with a visual summary, and also because other methods typically require additional models or assumptions (e.g., causal explanations~\cite{madumal2019explainable} require a causal graph of the domain). 
The specific choice of the LRP-argmax algorithm was motivated by its selectivity, which reduces the amount of information that participants have to process. 
The accessibility problems of saliency maps we identified were mainly related to the presentation of the information. 
This indicates that simply highlighting how relevant parts of the input are for the prediction of an agent is insufficient even when based on other saliency map algorithms.

\paragraph{Future work} There are several directions we intend to explore in future work. 
First, as discussed earlier, there is a need to make saliency maps more understandable to users.
To this end, we plan to augment saliency maps with textual explanations that help users interpret the information correctly, similar to how Rabold et al.~\cite{rabold2019enriching} did with LIME explanations.
Hereby, we aim to train a machine learning model on descriptions written by domain experts confronted with the combination of HIGHLIGHTS-DIV and saliency maps presented in this work.
Furthermore, we plan to build up on our previous work \cite{weitz2019doyou} and explore the presentation of those textual explanations through virtual agents.

Second, we plan to explore interaction approaches that involve the user in the process, e.g., by only showing local information when the user asks for it as we did in the context of cooperative annotation \cite{baur2020explainable}.
This could reduce cognitive load while increasing the user's attention  to the local information when it is needed.

Finally, to verify that our results generalize beyond simulated environments, we would like to conduct user studies in real-world domains such as healthcare. 
Explainability is crucial in AI systems deployed in the medical field (e.g., pain classification~\cite{weitz2019deep}) since possible errors could lead to dire consequences.  
RL methods face additional challenges and requirements in the healthcare domain where random exploration of the state space is not possible and evaluation is challenging~\cite{gottesman2019guidelines,gottesman2018evaluating}, making explanation methods even more important.
In recent work, we have begun exploring the use of strategy summaries in healthcare using an HIV simulator~\cite{lage2019exploring}, and intend to further explore this direction.

%% file: conclusion.tex
\section{Conclusion}
\label{sec:conclusion}
This work is a first step toward the development of combined explanation methods for reinforcement learning (RL) agents that provide users with both global information regarding the agent's strategy, as well as local information regarding its decision-making in specific world-states.
To this end, we present a joint global and local explanation method, building on our prior work on strategy summaries (HIGHLIGHTS-DIV) and on generating saliency maps for deep RL agents (LRP-argmax). 
This method is easily adaptable to other global and local algorithms.

To evaluate this combined global and local explanation method, as well as the contribution of each explanation type, we conducted a user study.
Hereby, we examined participants’ mental models through a \taskone{} and used an \tasktwo{} to investigate
whether their trust was appropriate given agents' capabilities.

Regarding the usefulness of \emph{global strategy summaries}, our results show that HIGHLIGHTS-DIV summaries (1) help to establish appropriate trust in agents based on neural networks (extending prior results about classic RL agents \cite{amir18highlights}) and (2) improve participants' mental models of those agents. 

The evaluation of \emph{local explanations} in the form of LRP saliency maps reveals strengths as well as weaknesses. 
On the one hand, our analysis shows that reinforcement learning comes with additional usability challenges not present in previously evaluated image classification tasks. 
First, presenting saliency maps on videos instead of static images \cite{anderson2019mere-mortals,alqaraawi2020evaluating} overwhelms users with a lot of information in a short amount of time and increases the risk of overlooking crucial information.  
Second, compared to more intuitive image classification tasks \cite{alqaraawi2020evaluating,selvaraju2016grad-cam}, the average users lacks experience to correctly infer how the highlighted regions affect the agent's long-term sequential decision-making.

On the other hand, the results indicate that saliency maps have the potential to (1) extend users' mental models beyond strategy summaries by providing insight into what information the agent used 
and (2) improve users' ability to choose the better agent even with random summaries.

Taken together, the results support a combination of local and global explanations, since participants in the combined explanation condition received the highest scores during our survey.
However, our evaluation suggests that simply highlighting pixels that are relevant for the agent's decision is insufficient for RL agents and that more work is needed to increase the accessibility of saliency maps.

\paragraph{Acknowledgements} 
This work was partially funded by a J.P. Morgan AI Faculty Research Award and the Deutsche Forschungsgemeinschaft (DFG) under project DEEP(Grant Number 392401413).

We thank Otto Grothe for his help with analyzing the participants' textual responses.

%% file: appendix.tex
\appendix

\section{Participants Demographics}
\label{appendix:demographics}
\begin{figure}[ht]
    \centering
    \includegraphics[width=\linewidth]{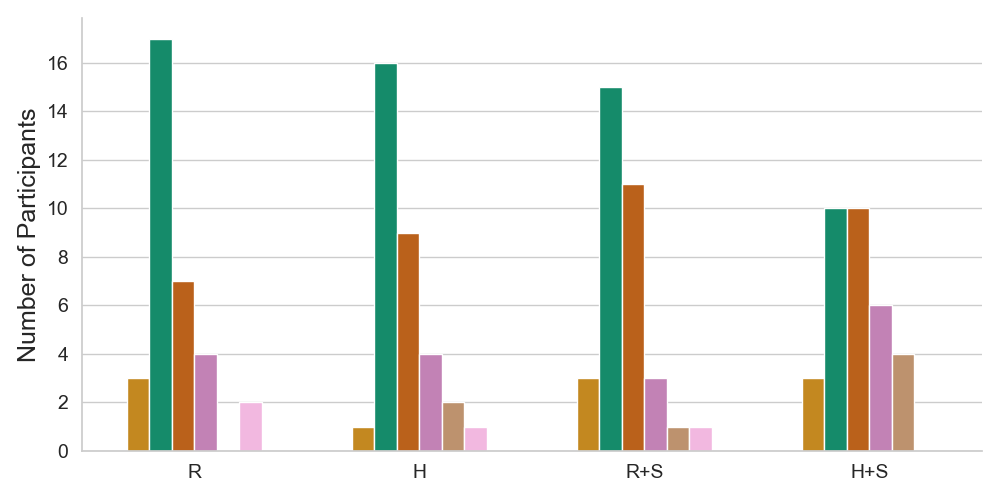}
    \caption{The number of participants in each age group per condition. The bars show from left to right: ``18-24'', ``25-34'',``35-44'', `` 45-54'', ``55-64'' and ``65 or older''.
    The categories ``17 or younger'' and ``do not want to specify'' were never selected.}
    \label{fig:ages}
\end{figure}

\begin{figure}[ht]
    \centering
    \begin{minipage}{0.45\linewidth}
    \centering
    \includegraphics[width=\linewidth]{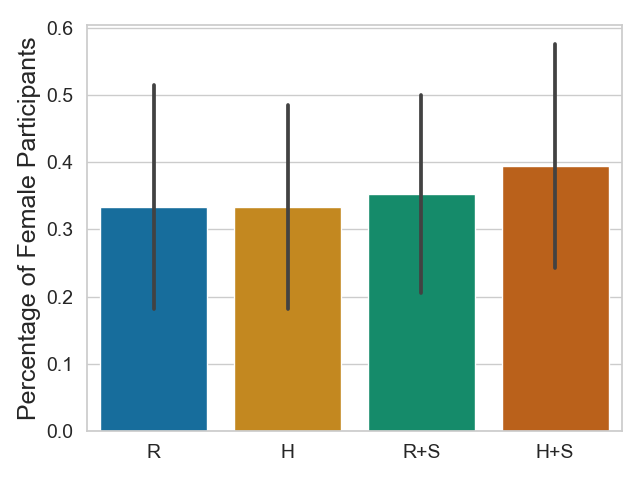}
    \caption{Number of female participants per condition.}
    \label{fig:femals}
    \end{minipage}
        \begin{minipage}{0.45\linewidth}
    \centering
    \includegraphics[width=\linewidth]{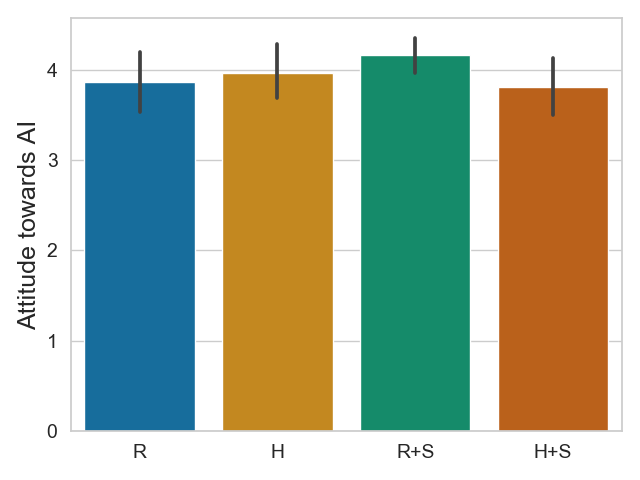}
    \caption{The average attitude towards AI, rated on a 5 point Likert scale. }
    \label{fig:attitude_AI}
    \end{minipage}
\end{figure}

In this section, we provide moe details regarding participants' demographics.
As Fig. \ref{fig:ages} shows, most participants were between 18 and 34 years old.
There were no major differences in gender distribution between the four conditions (Fig.~\ref{fig:femals}). 

{}

\begin{figure}[ht]
    \centering
     \includegraphics[width=0.8\linewidth]{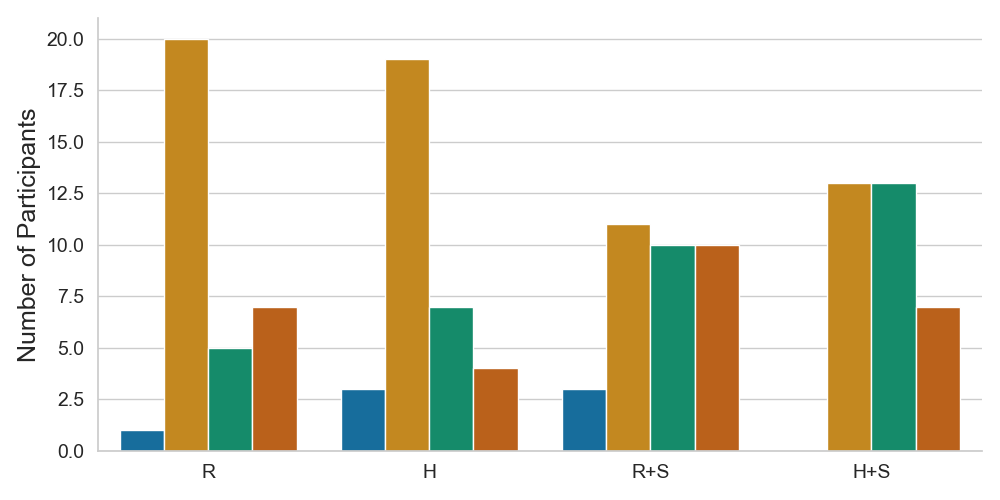}
    \caption{The Pacman experience across all conditions where the bars depict when the participants played Pacman the last time. From left to right the bars represent:  ``never'', ``more than 5 years ago'', ``less then 5 years ago'' and ``less than 1 year ago``. 
    }
    \label{fig:expierience_Pamcan}
\end{figure}

We verified that participants in different conditions did not differ much in their AI experience and views and in their Pacman experience.
To this end, we asked them when they played Pacman for the last time (1=``never'', 2=``more than 5 years ago'', 3=``less then 5 years ago'', 4=``less than 1 year ago'').
Across all four conditions the median group was 2:``I played Pacman more than 5 years ago''.
A comparison is shown in figure \ref{fig:expierience_Pamcan}.
    
For the AI experience we adapted a description of AI from Zhang et al.~\cite{zhang2019artificial} and Russel \cite{russell2016artificial} to ``The following questions ask about Artificial Intelligence (AI). Colloquially, the term `artificial intelligence' is often used to describe machines (or computers) that mimic `cognitive' functions that humans associate with the human mind, such as `learning' and `problem solving'. AI agents are already able to perform some complex tasks better than the median human (today). Examples for such intelligent agents are search engines, chatbots, chessbots and voice assistants.'' 

\begin{figure}[ht]
    \centering
    \small
    \begin{minipage}{0.48\linewidth}
    \centering
    \includegraphics[width=\linewidth]{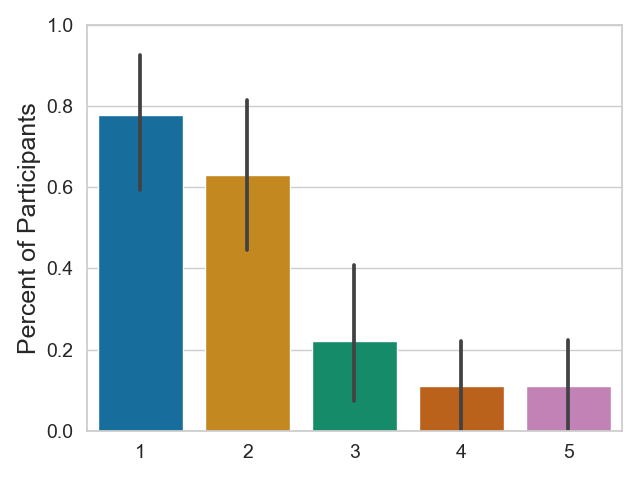}
    \conditionR
    \end{minipage}
        \begin{minipage}{0.48\linewidth}
    \centering
    \includegraphics[width=\linewidth]{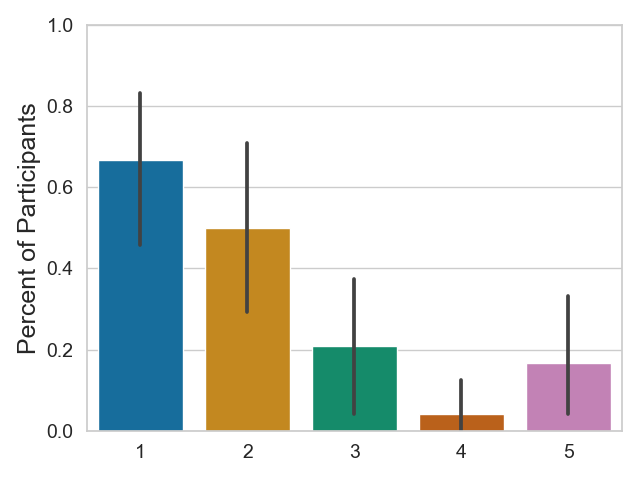}
    \conditionH
    \end{minipage}
    
    \begin{minipage}{0.48\linewidth}
    \centering
    \includegraphics[width=\linewidth]{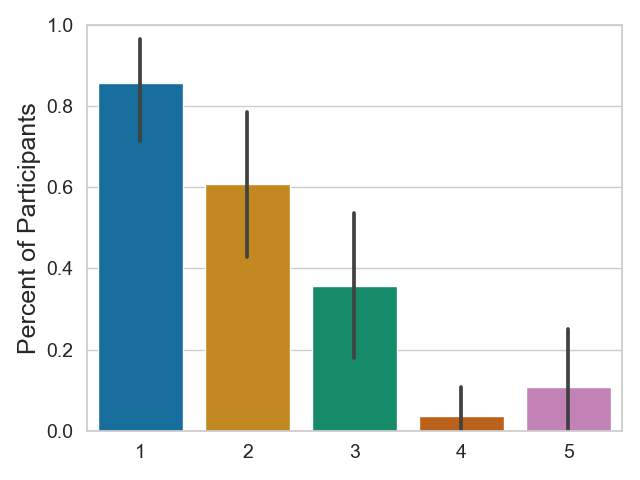}
    \conditionRS
    \end{minipage}
        \begin{minipage}{0.48\linewidth}
    \centering
    \includegraphics[width=\linewidth]{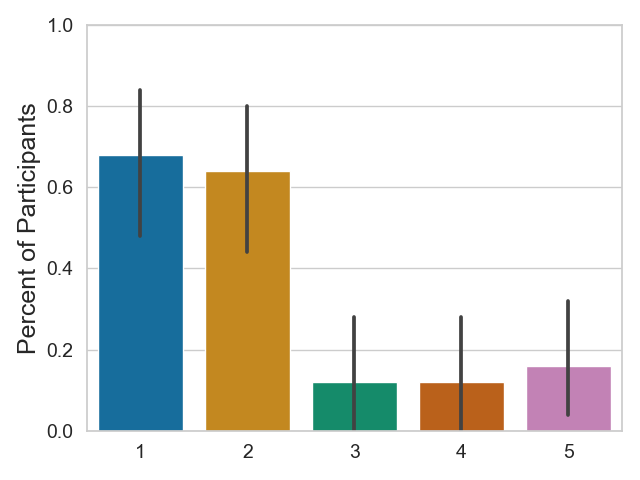}
    \conditionHS
    \end{minipage}
    \caption{Distribution of the chosen AI experience items for each condition. The x-axis depicts the items described above.}
    \label{fig:experience_XAI}
\end{figure}

After that, every participant who stated to have AI experience (104 across all conditions) had to select one or more of the following items:
\begin{itemize}
    \item 1: I know AI from the media.
    \item 2: I use AI technology in my private life.
    \item 3: I use AI technology in my work.
    \item 4: I took at least one AI related course.
    \item 5: I do research on AI related topics.
    \item Other:
\end{itemize}
The last free form option was used exactly once and read ``work on MTurk''.
The distribution of the other items for each condition is shown in Fig. \ref{fig:experience_XAI}.

To measure the participants' attitude towards AI we adapted a question from Zhang et al~\cite{zhang2019artificial} and asked them to rate their answer to the question ``Suppose that AI agents would achieve high-level performance in more areas one day. How positive or negative do you expect the overall impact of such AI agents to be on humanity in the long run?'' on scale from 5 point Likert scale from ``Extremely negative'' to ``Extremely positive''.
The results are shown in Fig.~\ref{fig:attitude_AI}.

{}

\clearpage

\section{Supplementary Results}
\label{appendix:results}
In this section, we present additional information about the results of the study that goes beyond the main hypotheses we explored and described in the paper.

\begin{figure}[ht]
    \centering
    \small
    \begin{minipage}{0.48\linewidth}
    \centering
    \includegraphics[width=\linewidth]{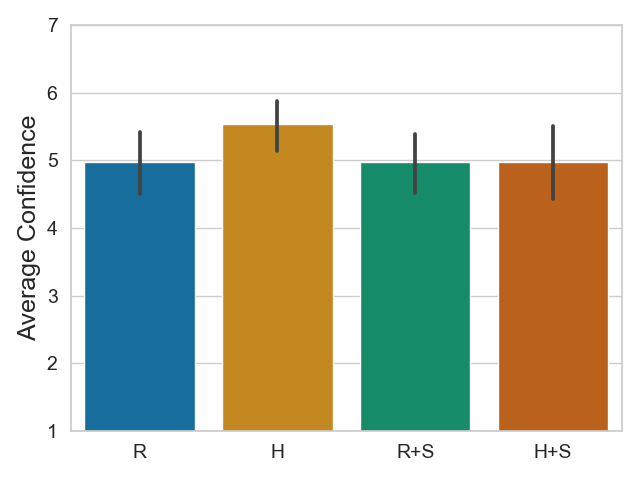}
    (a) \taskone
    \end{minipage}
        \begin{minipage}{0.48\linewidth}
    \centering
    \includegraphics[width=\linewidth]{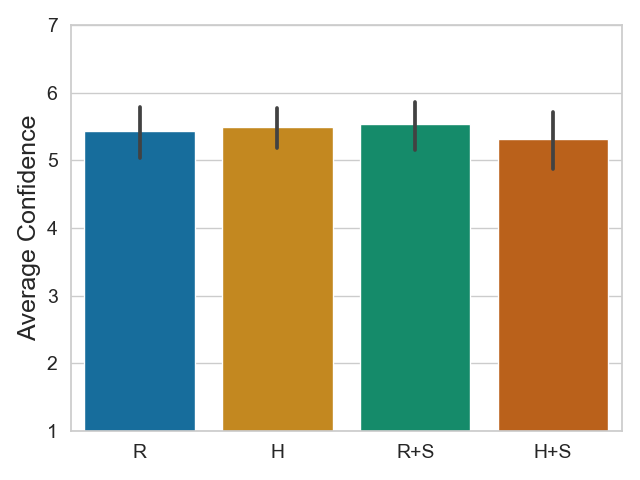}
    (b) \tasktwo
    \end{minipage}
    \caption{The average confidence that participants in each condition had in their answers during each task.}
    \label{fig:confidences}
\end{figure}

\paragraph{Confidence, time and pauses}
To investigate whether participants were confident in their decisions 
, they had to rate the confidence in each of their selections (item selection in the \taskone{} and agent selection in the \tasktwo{}) on a 7 point Likert scale.
The results across each task are shown in Fig. \ref{fig:confidences}.

\begin{figure}[ht]
    \centering
    \small
    \begin{minipage}{0.48\linewidth}
    \centering
    \includegraphics[width=\linewidth]{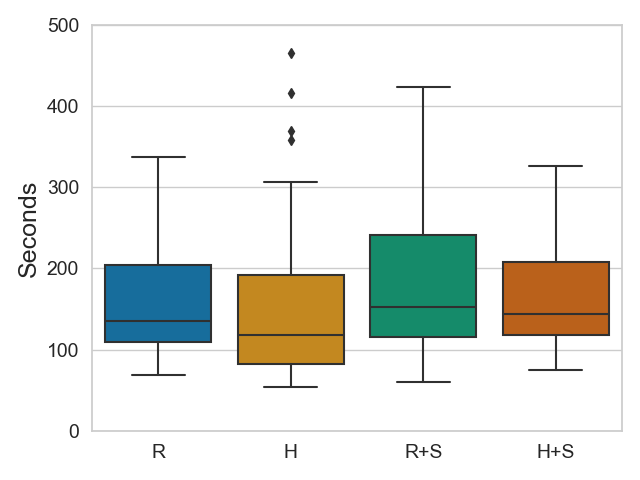}
    (a) \taskone
    \end{minipage}
        \begin{minipage}{0.48\linewidth}
    \centering
    \includegraphics[width=\linewidth]{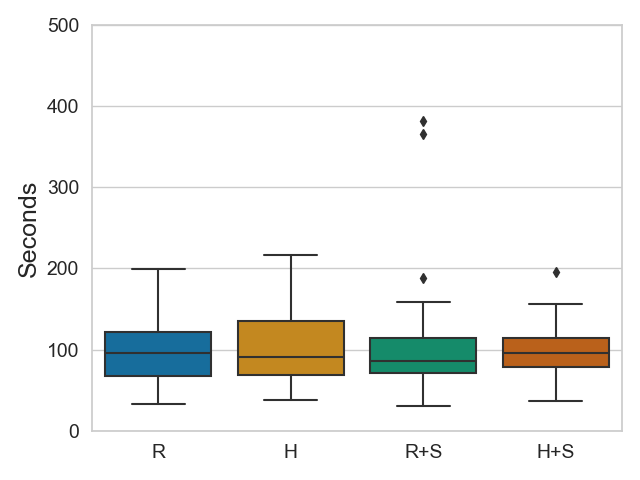}
    (b) \tasktwo
    \end{minipage}
    \caption{The average time taken by participants in each condition per agent analysis (a) and comparison of agent pairs (b).}
    \label{fig:times}
\end{figure}

\begin{figure}[ht]
    \centering
    \small
    \begin{minipage}{0.48\linewidth}
    \centering
    \includegraphics[width=\linewidth]{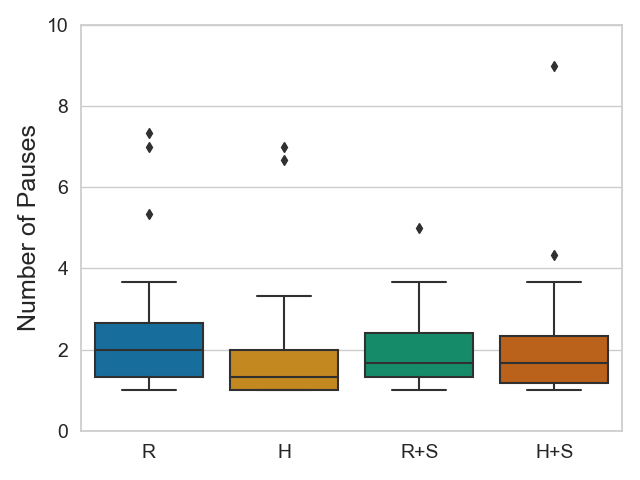}
    (a) \taskone
    \end{minipage}
        \begin{minipage}{0.48\linewidth}
    \centering
    \includegraphics[width=\linewidth]{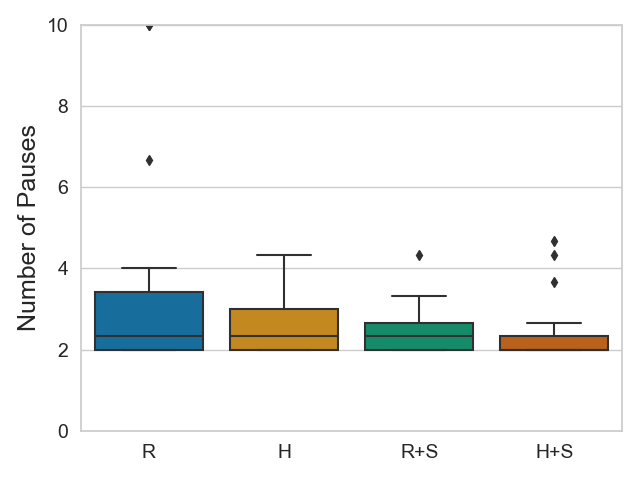}
    (b) \tasktwo
    \end{minipage}
    \caption{The average number of times that participants in each condition paused the videos during each agent analysis (a) and comparison of agent pairs (b).}
    \label{fig:pauses}
\end{figure}

To evaluate whether participants were especially diligent or effective during the tasks, we measured the time that each participant stayed on each of page of the survey and calculated the average time per task (each task consists of three pages).
Furthermore, we kept track of each time a video was paused, as described in section \ref{sec:analysis}.
The average completion times of participants and the average number of pauses are shown in Fig. \ref{fig:times} and \ref{fig:pauses}, respectively (shown in boxplots due to the presence of several outliers that strongly affect the mean values). 

{}

Fig. \ref{fig:confidences} (a) shows that participants in condition \conditionH were slightly more confident on average in their analysis of  the agents.
This is also reflected by the lesser amount of time per analysis (Fig. \ref{fig:times} (a)) and pauses (Fig. \ref{fig:pauses} (a)). 
Apart from this, there are no obvious differences between the average confidence, time and pause values for each task (Fig. \ref{fig:confidences} to \ref{fig:pauses}).

\paragraph{Participants' justifications}
\label{appendix:justification}

\begin{figure}[ht]
    \centering
        \small
    \begin{minipage}{0.48\linewidth}
    \centering
    \includegraphics[width=\linewidth]{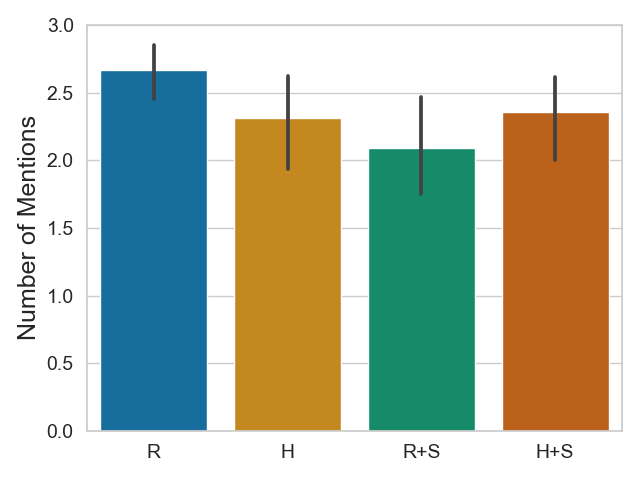}
    \taskone
    \end{minipage}
        \begin{minipage}{0.48\linewidth}
    \centering
    \includegraphics[width=\linewidth]{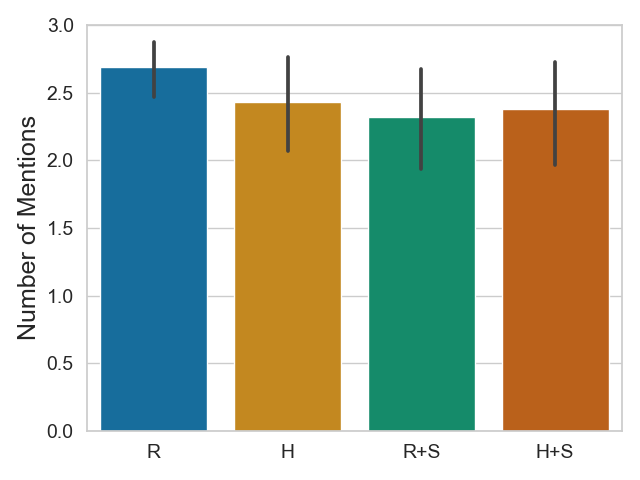}
    \tasktwo
    \end{minipage}
    \caption{Comparison of how often the participants referenced the agents' \textbf{gameplay} in their justifications for their answers.}
    \label{fig:gameplay_justifications}
\end{figure}

\begin{figure}[ht]
    \centering
    \small
    \begin{minipage}{0.48\linewidth}
    \centering
    \includegraphics[width=\linewidth]{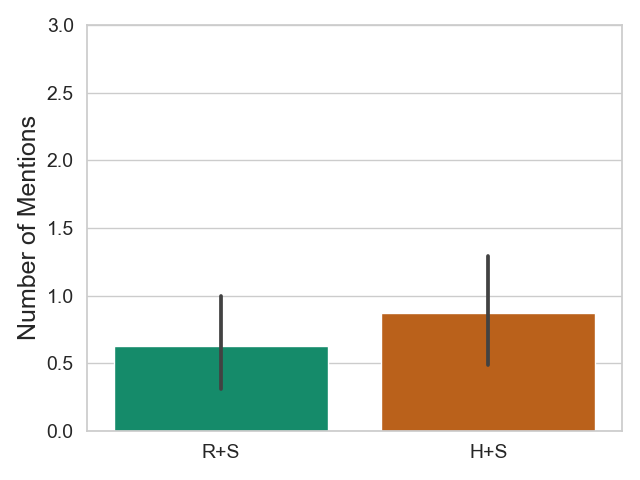}
    (a) \taskone
    \end{minipage}
        \begin{minipage}{0.48\linewidth}
    \centering
    \includegraphics[width=\linewidth]{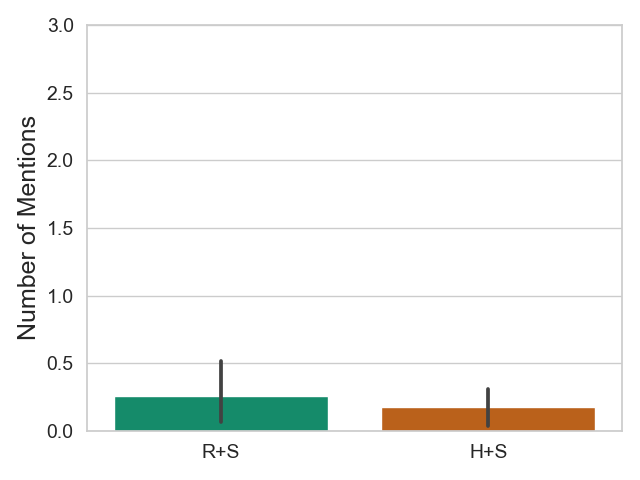}
    (b) \tasktwo
    \end{minipage}
    \caption{
    Comparison of how often the participants referenced the green highlighting of the LRP-argmax \textbf{saliency maps} in their justifications for their answers.
    }
    \label{fig:heatmap_justifications}
\end{figure}

\begin{figure}[ht]
    \centering
    \small
    \begin{minipage}{0.48\linewidth}
    \centering
    \includegraphics[width=\linewidth]{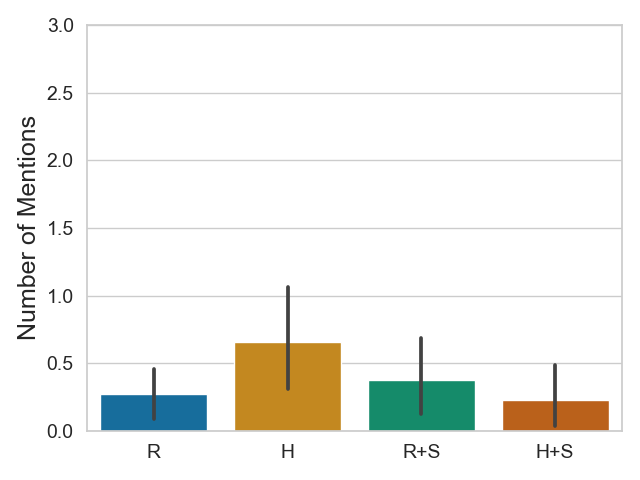}
    (a) \taskone
    \end{minipage}
        \begin{minipage}{0.48\linewidth}
    \centering
    \includegraphics[width=\linewidth]{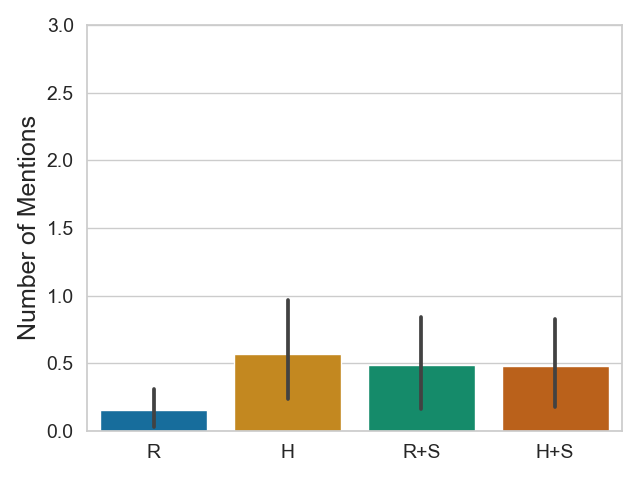}
    (b) \tasktwo
    \end{minipage}
    \caption{Comparison of how often the participants justifications contained \textbf{unjustified} arguments.}
    \label{fig:unjustified_justifications}
\end{figure}

As described in section \ref{sec:analysis}, an independent coder identified different concepts inside the participants' justifications. Figure~\ref{fig:heatmap_justifications} shows the average number of mentions of \emph{gameplay} and of \emph{saliency maps} in the different tasks, across the different conditions.
As discussed in section \ref{sec:results}, most participants mainly based their justifications on the agents' gameplay (Fig.~\ref{fig:gameplay_justifications}) and, in the saliency conditions, participants seldom mention the saliency maps in their justifications (see Fig.~\ref{fig:heatmap_justifications}).
Finally, Fig.~\ref{fig:unjustified_justifications} shows that participants in condition \conditionH{} gave more unjustified explanations in the \taskone{}. 
However, this observation did not repeat in the \tasktwo{}.

{}

{}

\section{Evaluation of the Retrospection Task}
\label{appendix:scoring_functions}

As described in section \ref{sec:analysis}, we evaluated  participants' scores in the object selection part of the retrospection task with a simple scoring function based on predefined answers by two of the authors involved in the training of the agents.
Hereby, we assign a score of $1$ to each object that is connected to the agents' specific goal and their source of information (Pacman's position for all agents),$-1$ for each object that was not related to the agents' reward function and $-0.5$ to objects that were related to the reward but on which the agent did not focus.
The specific scores are shown in table \ref{tb:scoring_object_selection}.

\begin{figure}[ht]
    
\begin{tabular}{|c|C|C|C|}
\hline
 selected object  & \agentPower{} & \agentReg{} & \agentGhost{} \\
\hline
``Pacman''&  1 & 1 & 1 \\
\hline
 ``normal pill''&  -1 & -0.5 & -0.5 \\
 \hline
 ``power pill''& 1 & -0.5 & -0.5 \\
 \hline
 ``normal ghost''& -1 & -0.5 & 1 \\
 \hline
 ``blue ghost''& -1 & 1 & 1\\
 \hline
 ``cherry'' & -1 & -0.5 & -0.5\\
 \hline
\end{tabular}{}

    \caption{Caption}
    \label{tb:scoring_object_selection}
\end{figure}{}   

For the free form answers to the question ``Please describe the strategy of the AI agent'' an independent coder identified various not mutually exclusive concepts contained in the participants answers. 
We aggregated these concepts into the following 16 groups, where the coder used 'G' for ghosts, 'PP' for power pills and 'NP' for normal pills:

\begin{enumerate}
    \item \emph{eating power pills}: ``eating PP'', ``eating as many PP as possible'', ``eat PP when ghosts are near'', ``eat PP when ghosts are near'', ``prioritizing PP'', ``prioritizing PP to eat ghosts'', ``prioritizing PP , but not eat ghosts'', ``eat PP to get points''
    \item \emph{ignore power pills}: ``do not care about PP''
    \item \emph{eat normal pills}: ``eat NP to get points'', ``eating NP'', ``eating as many NP as possible'', ``prioritizing NP'', ``clearing the stage''
    \item \emph{ignore normal pills}: ``do not care about NP'', ``focus on areas wihtout [sic] NP'' 
    
    \item \emph{avoid ghosts}: ``avoiding G'', ``avoiding G strongly'', ``wait for G to go away'', ``outmanoveuring G'', ``hiding from G'', ``mislead ghosts'', ``avoids being eaten / caught'', ``avoiding to lose / staying alive'', ``stays away from danger''   
    
    \item \emph{move towards ghosts}: ``being close to G'', ``trying to eat G NON blue'', ``(easily) caught by G'', ``easily caught by G''
    
    \item \emph{ignore ghosts}: ``do not care about G''

    \item \emph{making ghosts blue}: ``making G blue''
    
    \item \emph{eat blue ghosts}: ``being close to blue G'', ``eating as many G as possible'', ``eat blue G to get points'', ``chasing/going for G'', ``eating the blue G'', ``eating to jail many G''(jailing since the ghosts move back to jail after being eaten),``prioritizing PP to eat ghosts''
    
    \item \emph{avoid blue ghosts}: ``avoiding blue G''
    
    \item \emph{ignore blue ghosts}: ``do not care about blue G'', ``prioritizing PP , but not eat ghosts''
    
    \item \emph{eat cherry}: ``prioritizing cherry'', ``eat cherry to get points'', ``going for cherry'', ``eating cherry''
    
    \item \emph{ignore cherry}: ``do not care about cherry''
    
    \item \emph{random movement}: ``moving randomly'', ``move all over map'', ``switching directions /back\&forth'', ``not moving / being stuck'', ``sticking to walls / outside'', ``confused'', ``without strategy /random'', ``not planning ahead'', ``switching directions''
    
    \item \emph{focus on Pacman}: ``focus on PM'', ``focus on whats in front of/around PM'', ``stuck to itself''
    
    \item \emph{staying in corners}: ``staying in corners''
    
\end{enumerate}

These groups are used to define a simple scoring function.
Depending on the agent, each group could either be positive, neutral or negative.
Positive groups contain concepts that are in line with the predefined descriptions of the agents' strategies by two of the authors involved in the training.
Neutral groups consist of correct observations, which are byproducts of the agent's strategy, and negative concepts go against the agent's strategy.
Each positive group contained in an answer increased the participant's score by $1$ and each negative group decreased the score by $-1$.
Here, we define a group to be ``contained in an answer'' if at least one concept of this group was included in the answer.
Neutral groups did not affect the score.

\agentPower{}:
\begin{itemize}
    \item \emph{positive:} ``eat power pill'',``ignore normal pill'',``ignore ghosts'',``ignore blue ghost'',``ignore cherry'',``focus on Pacman'', ``staying in corners''
    \item \emph{neutral:} ``eat normal pill'',``making ghosts blue''
\end{itemize}

\agentReg{}:
\begin{itemize}
    \item \emph{positive}: ``ignore cherry'',``focus on Pacman'',``making ghosts blue'',``eat blue ghost''
    \item \emph{neutral}: ``eat normal pill'', ``eat power pill'', ``ignore ghosts''
\end{itemize}{}

\agentGhost{}:
\begin{itemize}
    \item \emph{positive}: ``avoid ghost'',``focus on Pacman'',``making ghosts blue'',``eat blue ghost'',``ignore cherry''
    \item \emph{neutral}:``eat normal pill'', ``eat power pill''
\end{itemize}{}

\clearpage

\section{Questionnaire}
\label{appendix:questionnaire}
In this section, we provide the complete questionnaire used in the study. 
On the first page the participants were asked to provide personal information:

\includegraphics[width=0.9\linewidth]{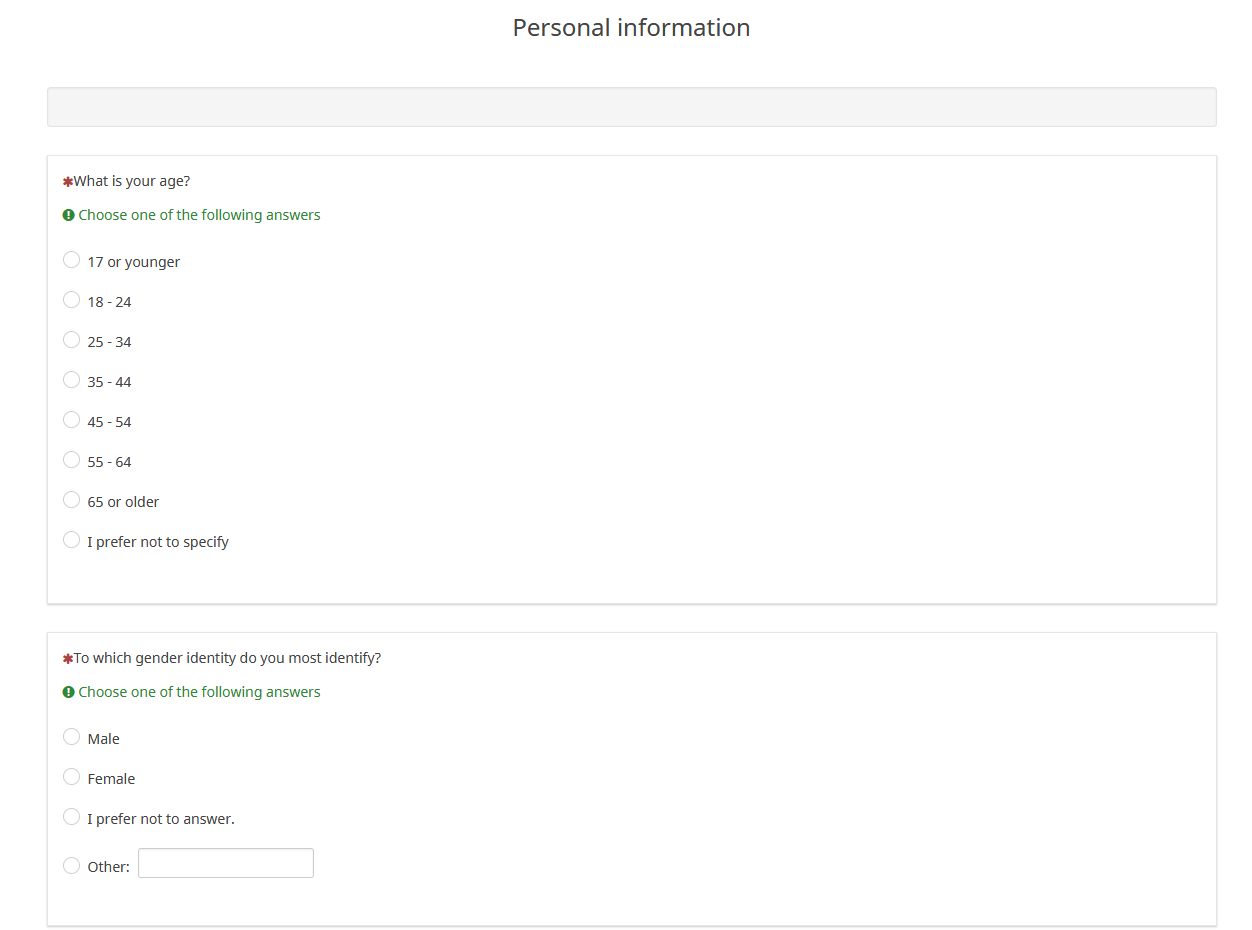}

\includegraphics[width=0.9\linewidth]{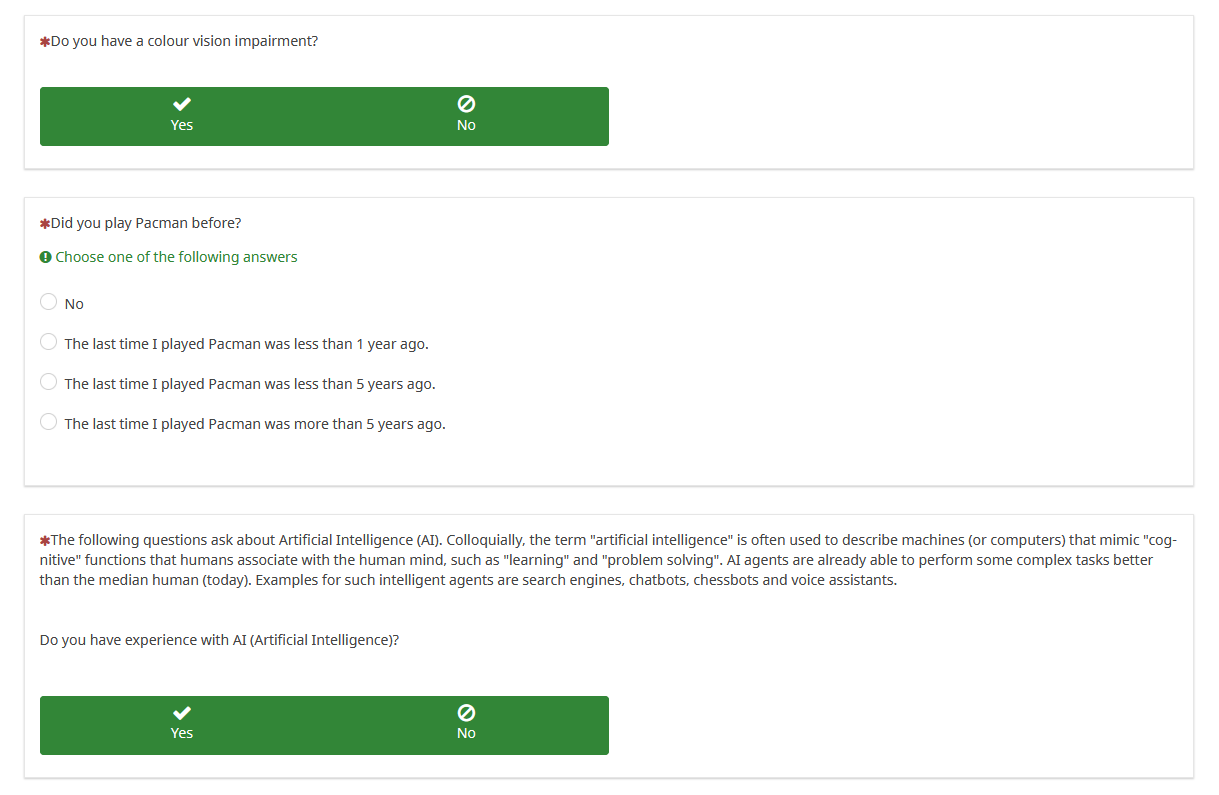}

\includegraphics[width=0.9\linewidth]{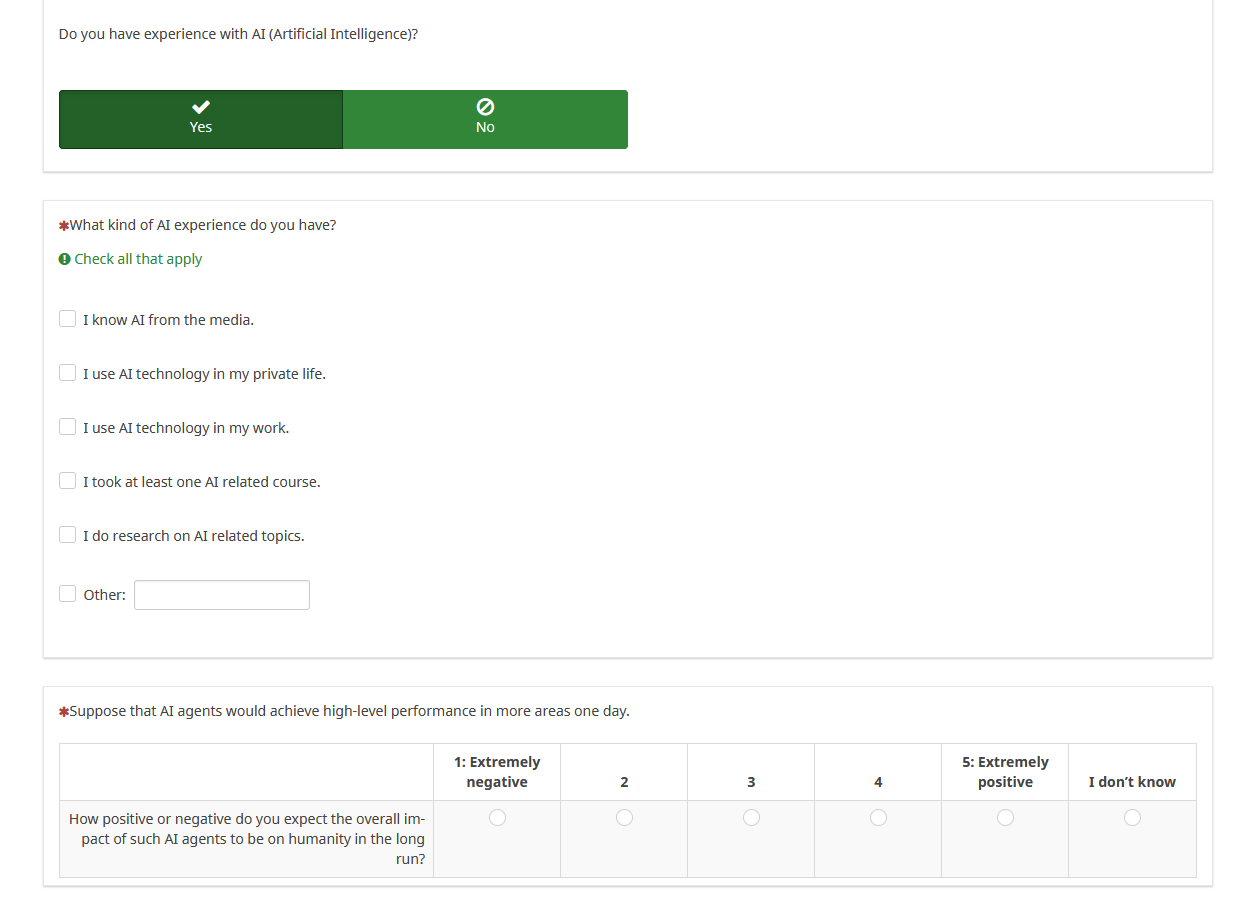}

\clearpage
Information about Pacman:

\includegraphics[width=\linewidth]{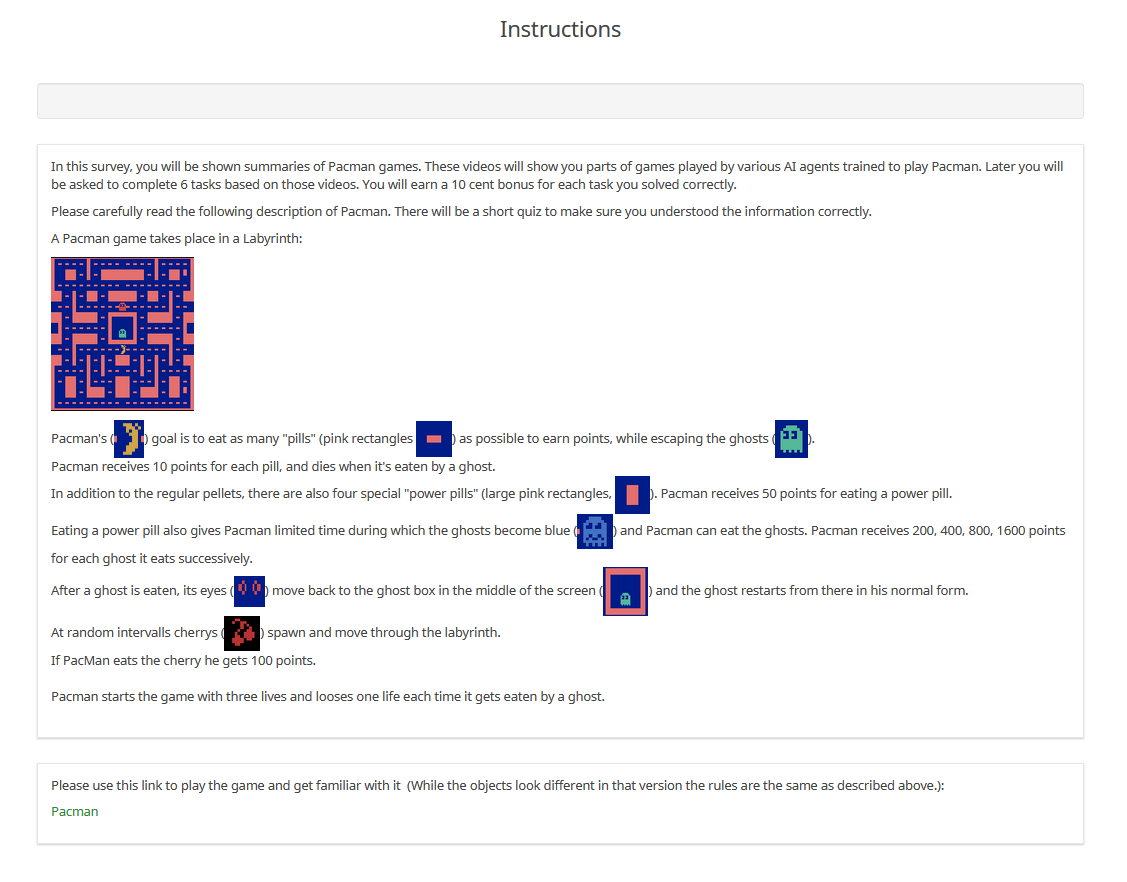}

\clearpage
This quiz tests whether the participants understood the information about Pacman.
Participants were sent back to the previous page if they got an answer wrong.

\includegraphics[width=\linewidth]{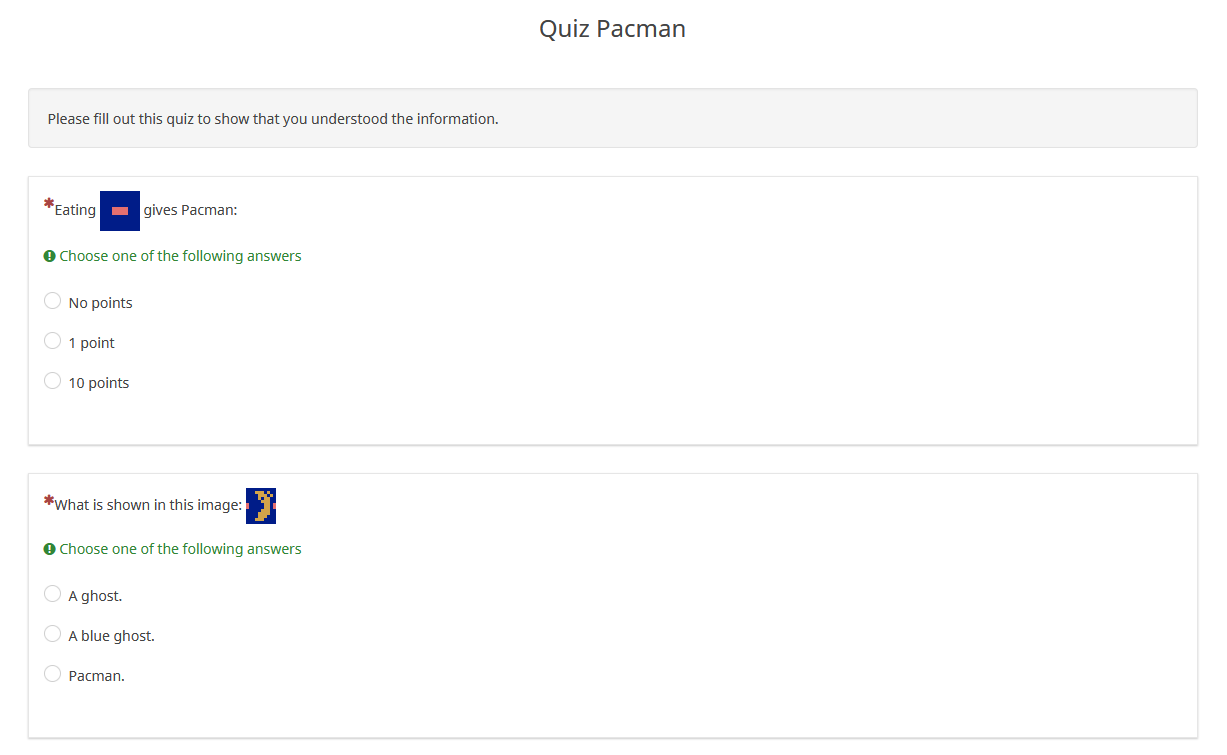}
\includegraphics[width=\linewidth]{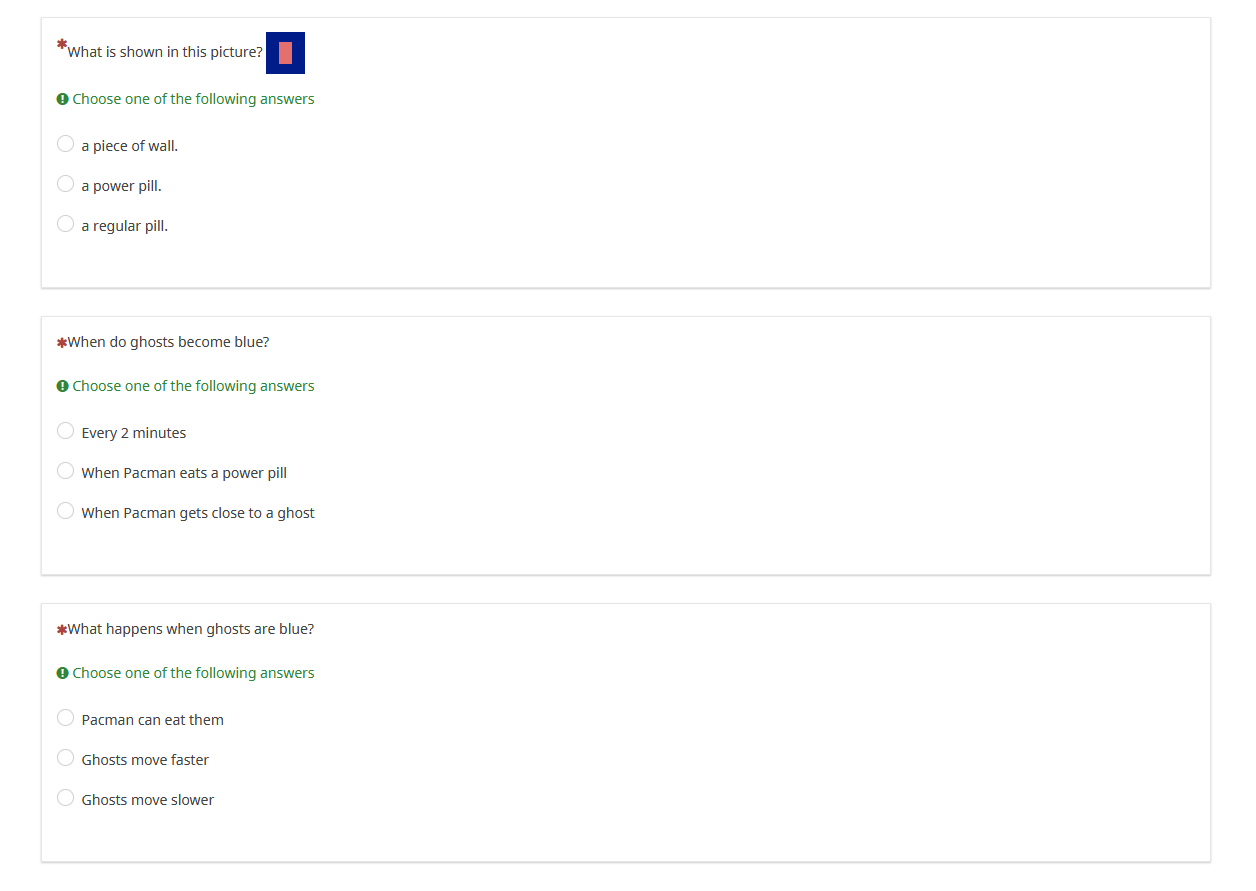}

\clearpage
Additional information about the provided explainable AI methods.
The information about saliency maps was only displayed if the participant was in one of the saliency conditions.

\includegraphics[width=\linewidth]{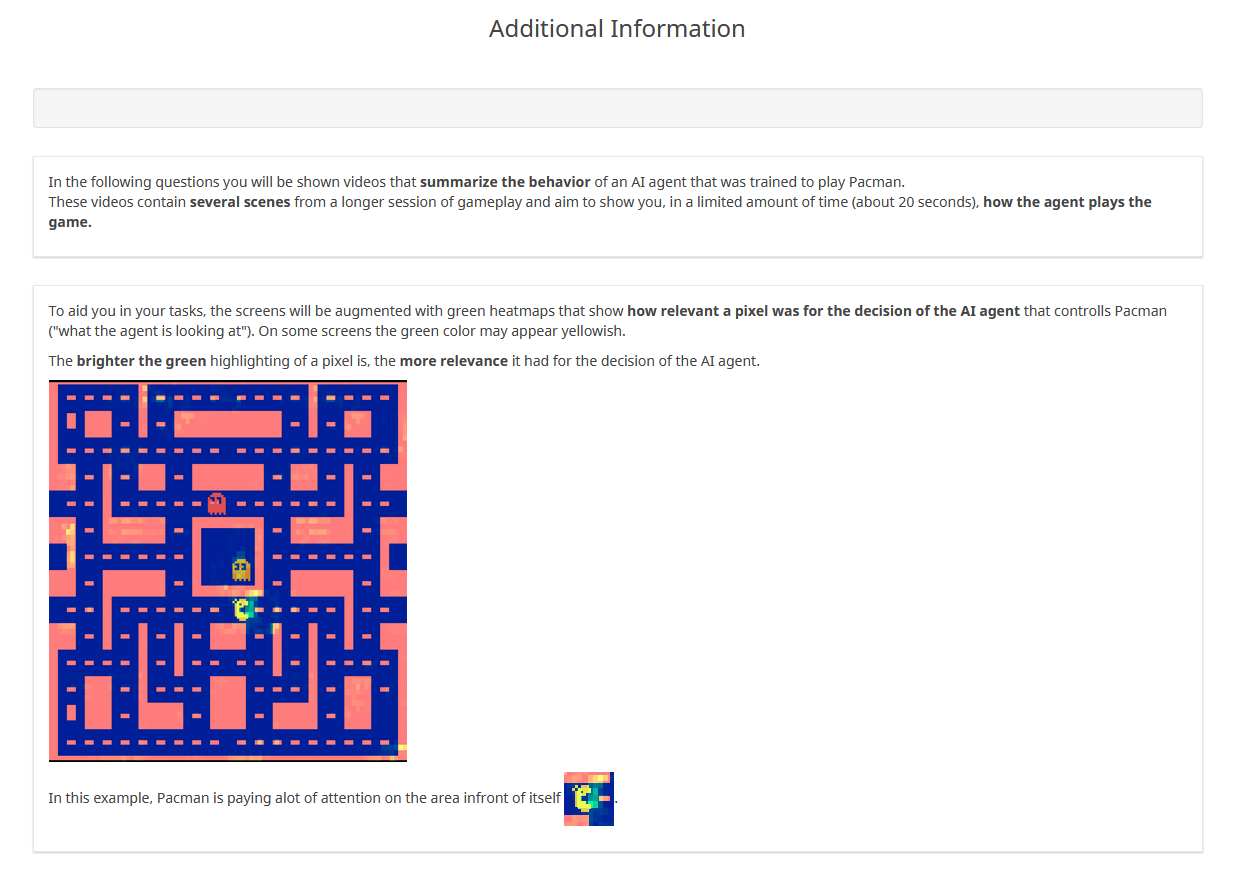}

\clearpage
This quiz tests whether the participants understood the information about the provided explainable AI methods.
Participants were sent back to the previous page if they got an answer wrong.

\includegraphics[width=\linewidth]{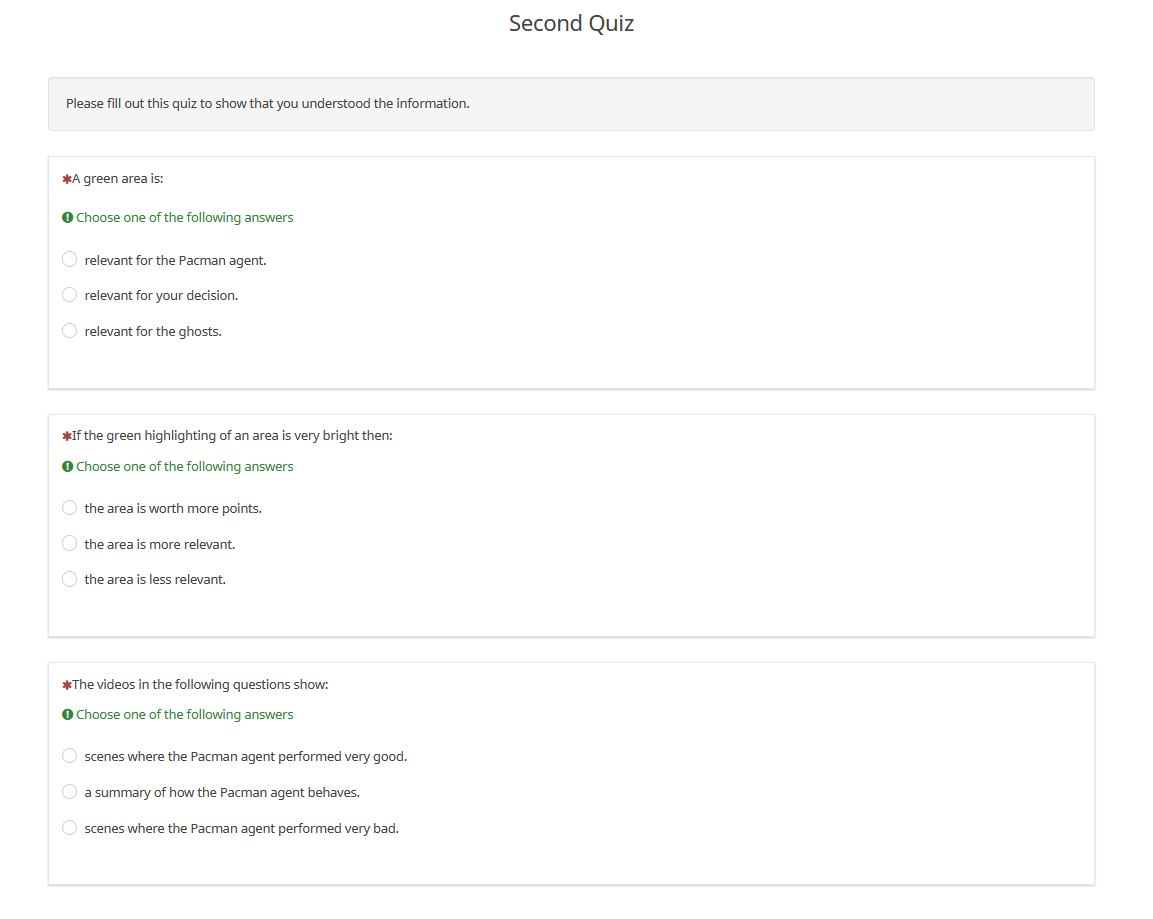}

\clearpage
This is the \taskone{} that was repeated for each of the three agents in a randomized order:

\includegraphics[width=\linewidth]{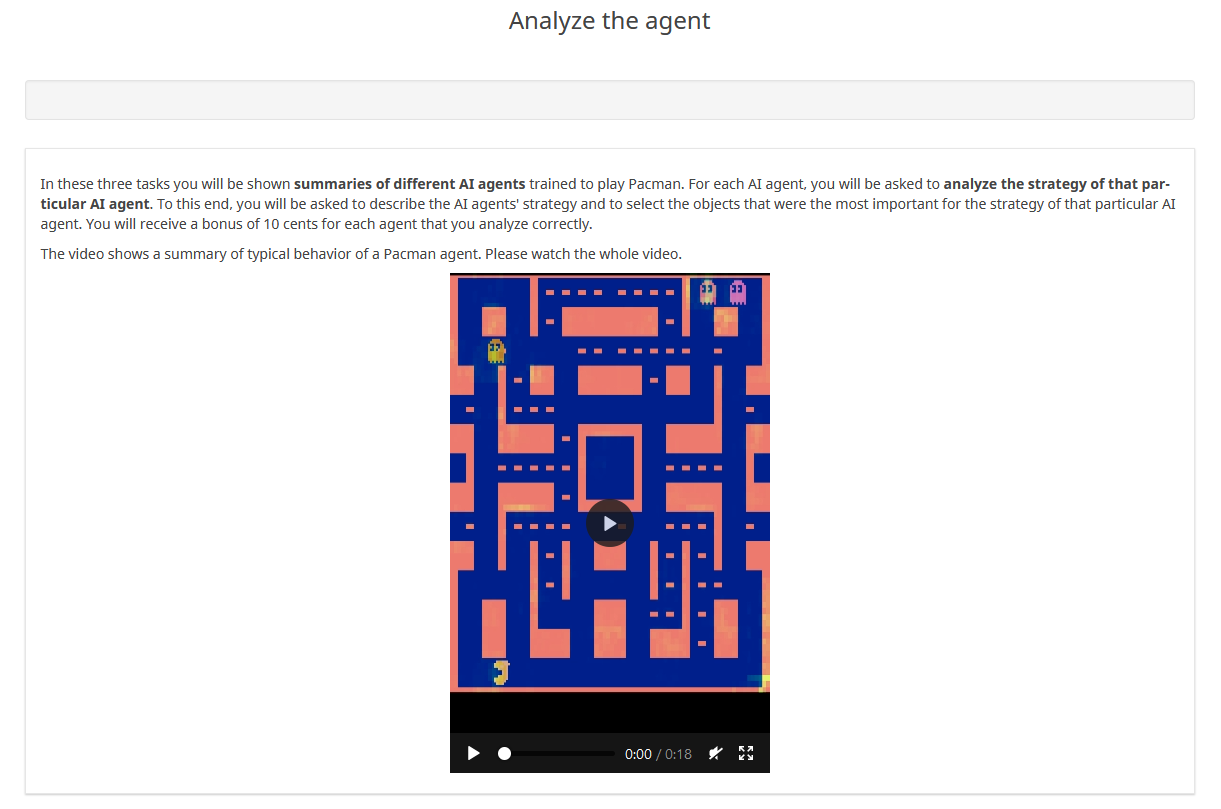}
\includegraphics[width=\linewidth]{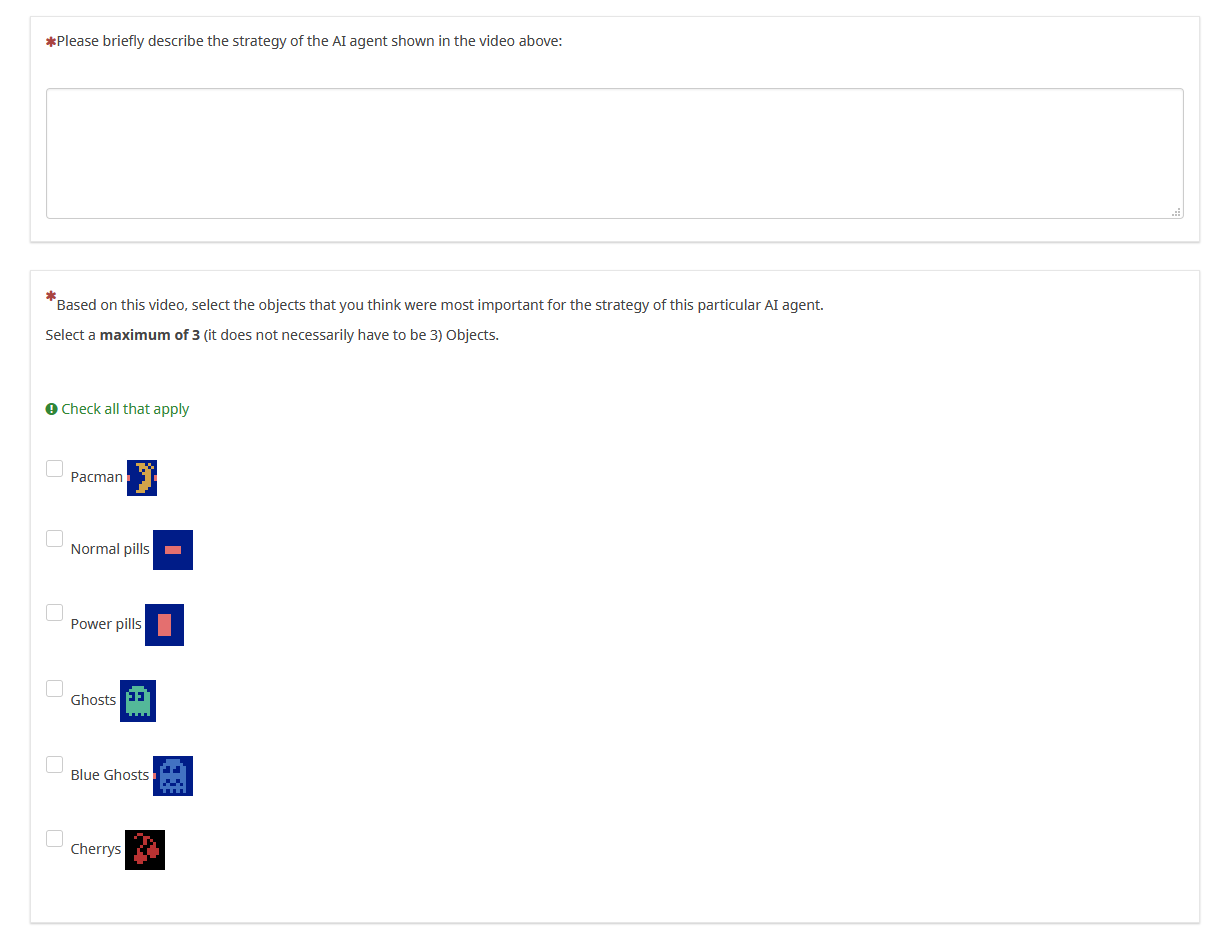}
\includegraphics[width=\linewidth]{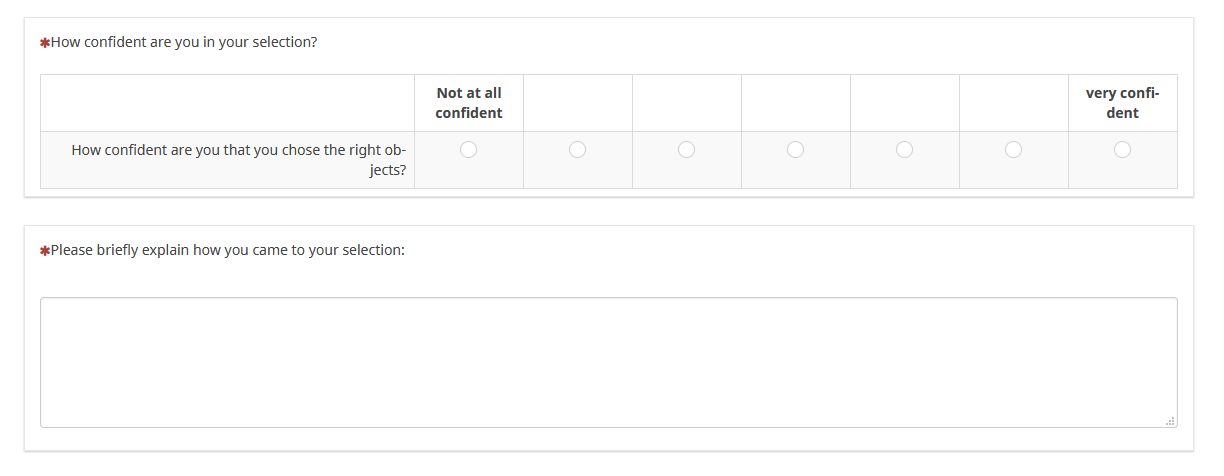}

\clearpage
After all three agents, the participants were asked for their satisfaction:

\includegraphics[width=\linewidth]{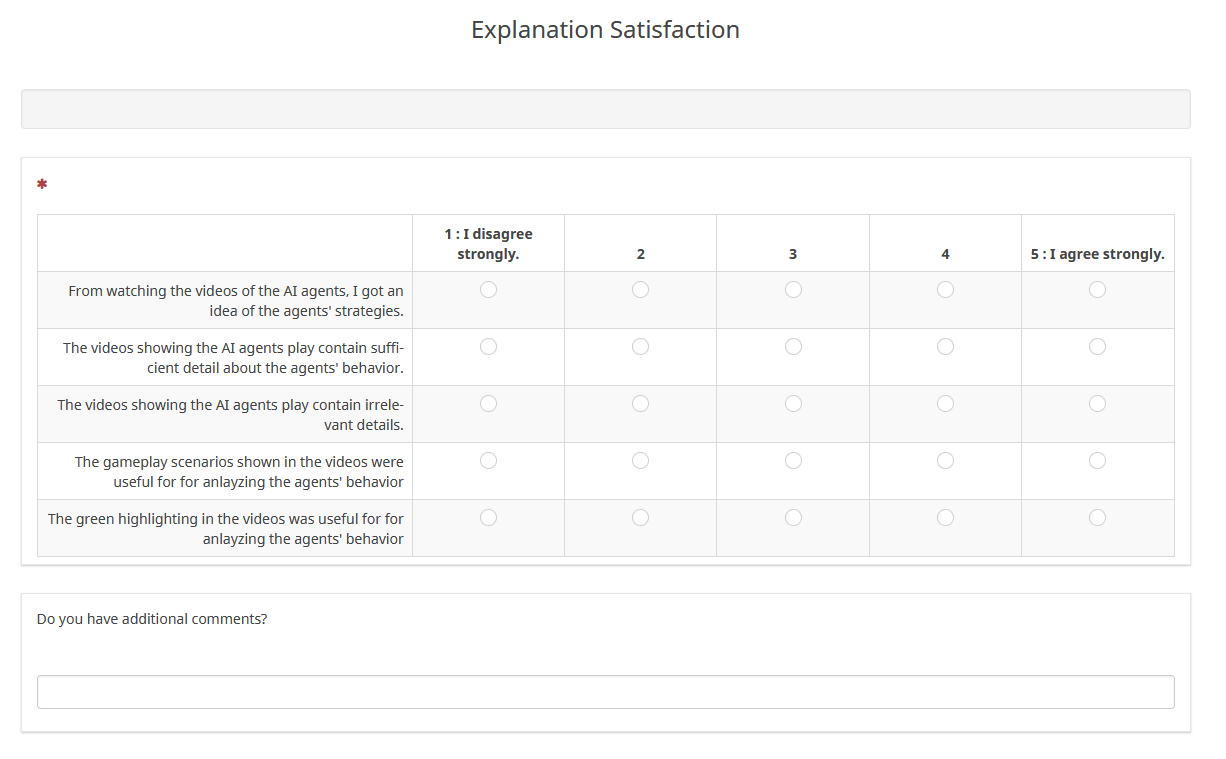}

\clearpage
This is the \tasktwo{} that was repeated for each combination of the three agents in a randomized order: 

\includegraphics[width=\linewidth]{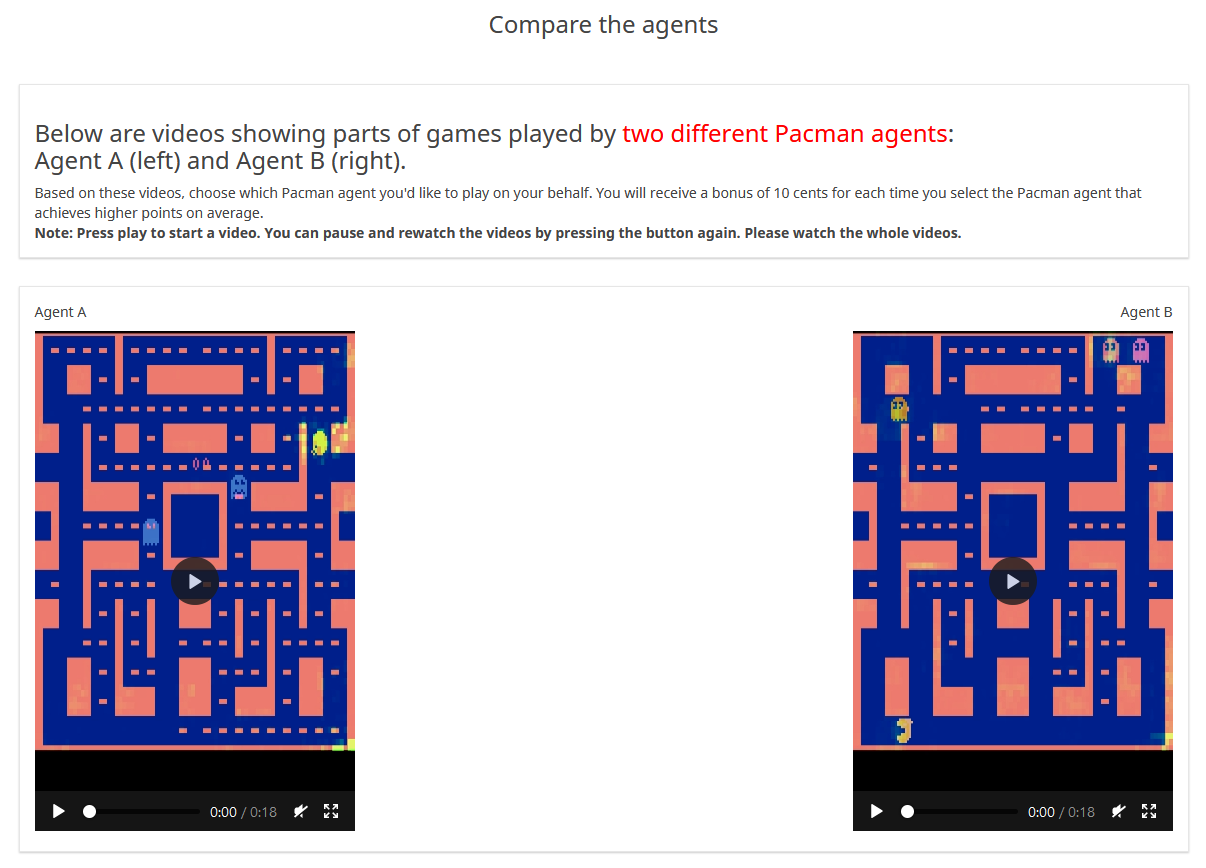}
\includegraphics[width=\linewidth]{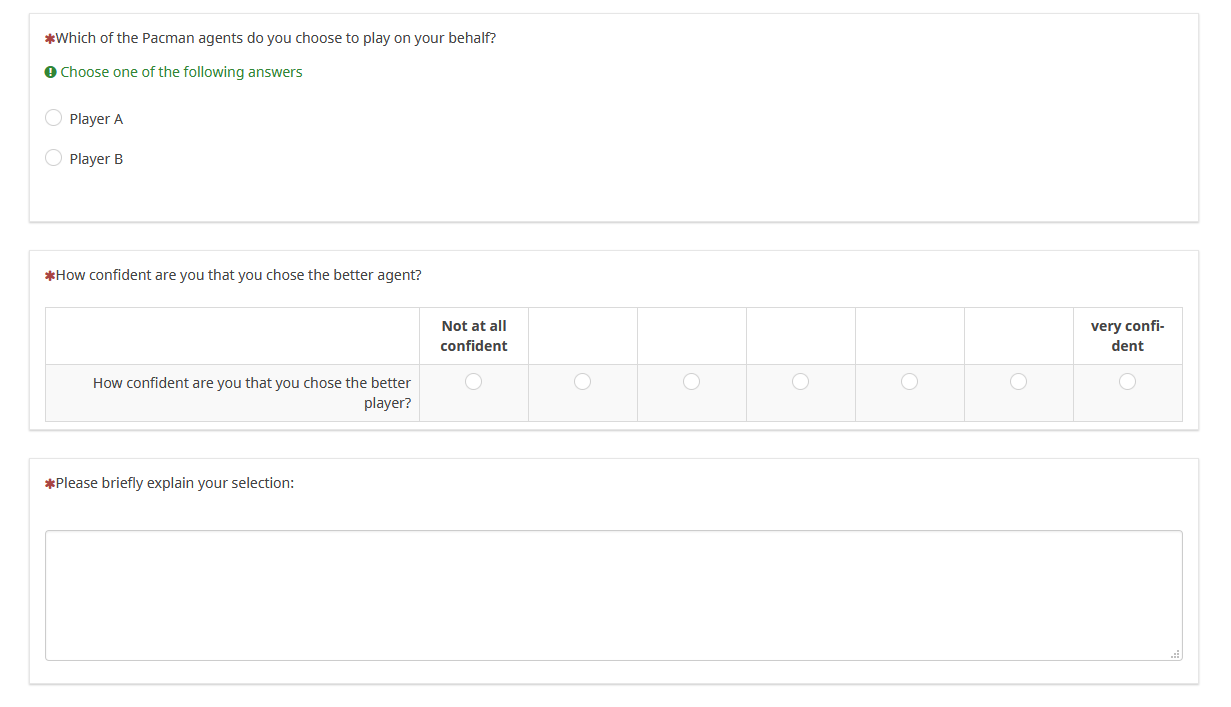}

\clearpage
After all three comparisons, the participants were asked for their satisfaction again:

 \includegraphics[width=\linewidth]{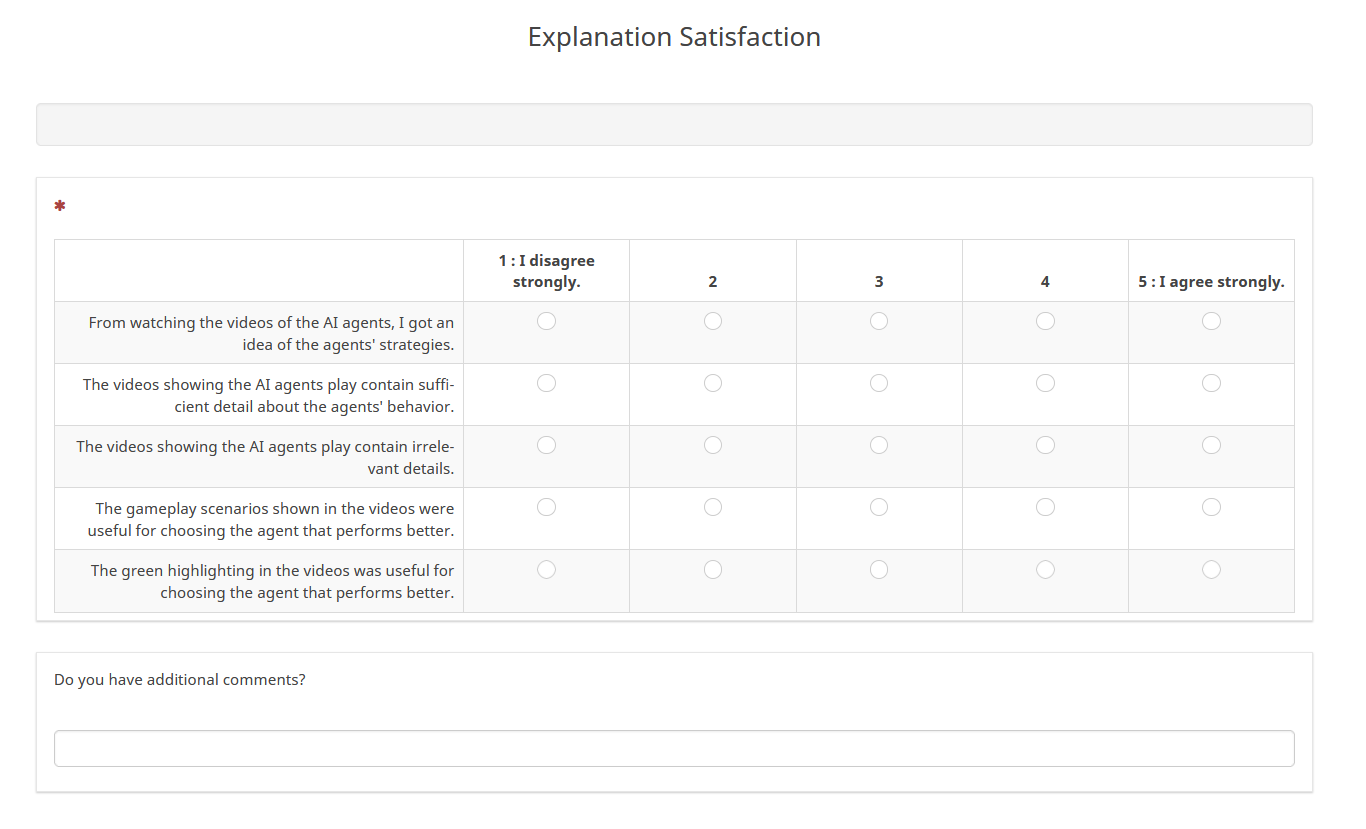}